\documentclass{article}

 \usepackage[preprint]{neurips_2026}

\usepackage[utf8]{inputenc} 
\usepackage[T1]{fontenc}    
\usepackage{hyperref}       
\usepackage{url}            
\usepackage{algorithm}      
\usepackage{algorithmic}    
\usepackage{booktabs}       
\usepackage{amsfonts}       
\usepackage{nicefrac}       
\usepackage{microtype}      

\usepackage{amsmath,amsfonts,bm}

\def\eqref#1{equation~\ref{#1}}

\def\1{\bm{1}}

\DeclareMathAlphabet{\mathsfit}{\encodingdefault}{\sfdefault}{m}{sl}
\SetMathAlphabet{\mathsfit}{bold}{\encodingdefault}{\sfdefault}{bx}{n}

\DeclareMathOperator*{\argmin}{arg\,min}

\usepackage{amsmath}
\usepackage{amssymb}
\usepackage{mathtools}
\usepackage{amsthm}
\usepackage{graphicx}
\usepackage{subcaption}
\usepackage{multirow, multicol}
\usepackage[table]{xcolor}
\usepackage{xcolor}         

\title{Diffusion-Inspired Reconfiguration of Transformers for Uncertainty Calibration}

\author{%
  Manh Cuong Dao\thanks{Equal contribution.} \\
  National University of Singapore \\
  \And
  Quang Hung Pham\footnotemark[1] \\
  Hanoi University of Science and Technology \\
  \And
  Phi Le Nguyen \\
  Hanoi University of Science and Technology \\
  \And
  Thao Nguyen Truong \\
  National Institute of Advanced Industrial Science and Technology \\
  \And
  Bryan Kian Hsiang Low \\
  National University of Singapore \\
  \And
  Trong Nghia Hoang\thanks{Corresponding author.} \\
  Washington State University \\
}

\newcommand{\ourmethod}{\textsc{DIRECTOR}}
\newcommand{\qhung}[1]{#1}

\begin{document}

\maketitle

\begin{abstract}
Uncertainty calibration in pre-trained transformers is critical for their reliable deployment in risk-sensitive applications.~Yet, most existing pre-trained transformers do not have a principled mechanism for uncertainty propagation through their feature transformation stack.~In this work, we propose a diffusion-inspired reconfiguration of transformers in which each feature transformation block is modeled as a probabilistic mapping.~Composing these probabilistic mappings reveals a probability path that mimics the structure of a diffusion process, transporting data mass from the input distribution to the pre-trained feature distribution.~This probability path can then be recompiled on a diffusion process with a unified transition model to enable principled propagation of representation uncertainty throughout the pre-trained model’s architecture while maintaining its original predictive performance.~Empirical results across a variety of vision and language benchmarks demonstrate that our method achieves superior calibration and predictive accuracy compared to existing uncertainty-aware transformers.\vspace{-3mm}
\end{abstract}

\section{Introduction}
\label{sec:intro}

The transformer architecture~\citep{vaswani2017attention} has become a universal backbone in most large-scale pre-trained or foundation models spanning numerous domains.~These include language \citep{devlin2019bert,radford2019language,brown2020language,achiam2023gpt,touvron2023llama}, vision~\citep{dosovitskiy2020image, touvron2021training, liu2021swin}, speech~\citep{ baevski2020wav2vec,hsu2021hubert,radford2023robust}, and even more complex domains with multi-modal data (e.g., text-image)~\citep{radford2021learning, liu2023visual,driess2023palm, team2023gemini}.

\noindent {\bf Challenge.}~Despite their prevalence, existing transformer-based models lack a principled mechanism to assess prediction uncertainty.~This often leads to incorrect predictions being assigned high confidence~\citep{guo2017calibration, mukhoti2020calibrating} which raises safety concerns in high-stake applications~\citep{moon2020confidence,zhu2023openmix}. Uncertainty calibration (UC), or confidence calibration, addresses this issue by ensuring that a model’s predicted confidence faithfully reflects its empirical accuracy, thereby making confidence scores reliable for downstream decision-making.~For example, effective UC techniques can help determine when to defer to human experts in scenarios where the model exhibits high representation and/or prediction uncertainty, particularly in risk-sensitive applications~\citep{tran2022plex, rudner2022tractable, rudner2023function}.~While UC has been extensively studied in conventional low-complexity deep neural networks, existing techniques mainly focus on imposing probabilistic priors on network weights and approximating their posteriors via either variational inference or posterior sampling.~This quickly becomes both inaccurate and prohibitively expensive when the model complexity increases.

\noindent {\bf Emerging Paradigm.}~To sidestep the challenge of computing posteriors over models with exceedingly large complexities, there are emerging approaches that aim to reparameterize the attention outputs as (sparse) Gaussian process predictions~\citep{liu2020simple, chen2023calibrating, bui2025revisiting, chen2024self} and recast the pre-trained transformer as a probabilistic chain mapping from the data distribution to a feature distribution.~This enables principled, uncertainty-aware sampling of feature representations by simulating the chain rather than inferring them via computing the prohibitively expensive model posterior, thereby motivating a more scalable paradigm for UC in large models.


\begin{figure*}[t]
    \centering
    \includegraphics[width=\linewidth]{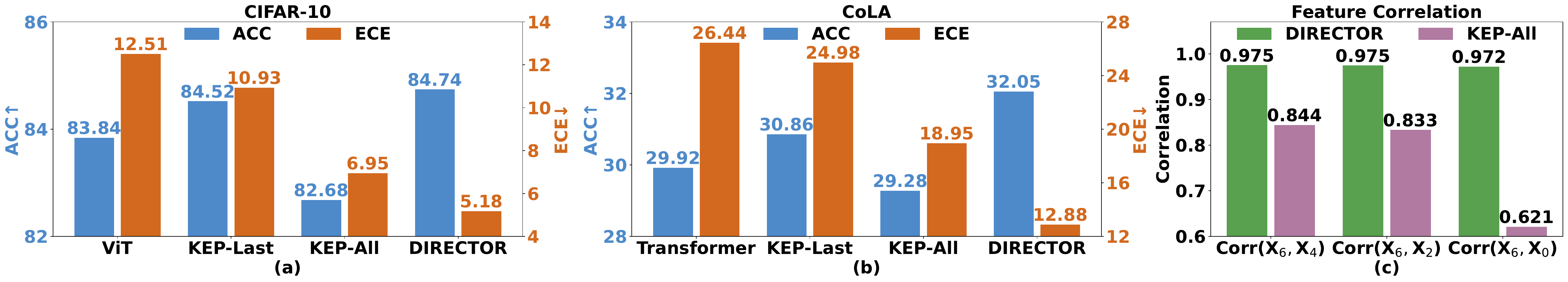}
    \caption{Comparison of accuracy (ACC$\uparrow$) and uncertainty calibration (ECE$\downarrow$) across pretrained models (ViT, Transformer), GP-reparameterized method KEP~\citep{chen2024self} applied to either the last attention block (KEP-last) or all attention blocks (KEP-All), and our method (\ourmethod) on (a) CIFAR-10 and (b) CoLA dataset. Panel (c) compares the correlation between features at the first layer ($\mathbf{X}_6$) and and those at deeper layers ($\mathbf{X}_4$, $\mathbf{X}_2$, and $\mathbf{X}_0$ at the last layer) for \ourmethod{} and KEP-All on CIFAR-10 dataset.} 
    \label{fig:acc_ece_corr}
    \vspace{-10pt}
\end{figure*}

\noindent {\bf Research Gap.} Intuitively, separate reparameterization fails to account for the correlations among feature transformations at different attention blocks that were established during pre-training. This results in an unfavorable exchange between uncertainty calibration and performance. Previous approaches that adopt separate re-parameterization often improve uncertainty calibration at the cost of significantly reduced performance (see Fig. \ref{fig:acc_ece_corr}a, \ref{fig:acc_ece_corr}b), which does not align with the expectation that improved uncertainty calibration should, in general, lead to improved performance. From the above intuition, this is not surprising since accurate uncertainty propagation means being able to properly account for the high correlation across feature transformation steps. This is however not upheld in previous approaches according to our investigation. For instance, our results in Fig. \ref{fig:acc_ece_corr}c show that the correlation between the feature at the first KEP-SVGP’s layer and intermediate representations rapidly decreases as it is propagated further towards the solution head. In contrast, our proposed method DIRECTOR manages to correctly preserve this high correlation. As a result, Fig. \ref{fig:acc_ece_corr} shows that DIRECTOR indeed improves uncertainty calibration while also improving performance as expected.



\noindent {\bf Solution Vision.}~To address this gap, we propose distilling the sequence of reparameterized attention blocks into a unified diffusion model.~Rather than treating each block as an independent reparameterization, we model the entire sequence as a continuous stochastic process over the feature embedding space.~In this view, the observed transformations of a pre-trained model are interpreted as samples from a diffusion process governed by a single spatiotemporal transition kernel that maps the data distribution to the final representation.~This unified view allows us to capture cross-block correlations established during pre-training while providing a principled mechanism for propagating uncertainty.


\noindent {\bf Technical Contributions.}~To substantiate this vision, we develop a diffusion-based framework for unified uncertainty propagation across transformer blocks with the following technical contributions:

\textbf{1.}~We reinterpret the step-wise feature transformations of a pre-trained transformer as transition samples from a probabilistic path that maps the data distribution to the feature distribution.~This perspective generalizes the transformer into a diffusion model parameterized by a unified transition kernel, which can be learned from these observed transitions.~Such reconfiguration supports local uncertainty calibration at individual attention blocks while ensuring an accurate flow of uncertainty propagation across the entire network (see Section~\ref{subsec:cut}).

\textbf{2.}~We design a training algorithm that distills the observed sequence of feature transformations into a unified spatiotemporal transition kernel of a diffusion process.~The learned kernel captures the inherent correlations among feature transformations across attention blocks established during pre-training, providing a tractable and principled procedure for uncertainty calibration in large pre-trained models.~This allows us to establish a generative paradigm for UC, where uncertainty-aware representation samples are drawn directly from the learned diffusion process rather than inferred via intractable model posteriors (see Section~\ref{subsec:diffomer}).

\textbf{3.}~We conduct extensive experiments on vision and language benchmarks to evaluate calibration quality, robustness, and out-of-distribution (OOD) detection.~The results show that our approach consistently improves uncertainty calibration while preserving predictive performance over existing state-of-the-art pre-trained transformer models.~Remarkably, it achieves these gains with fewer parameters than the original model, leading to improved memory efficiency.~These results demonstrate the feasibility of post-hoc embedding probabilistic reasoning into the internal structure of large pre-trained models for uncertainty calibration without sacrificing performance.~This opens a new direction for enhancing their reliability in safety-critical settings (see Section~\ref{sec:experiments}).

\section{Diffusion-Inspired Reconfiguration of Transformers}
\label{sec:method}
Recent efforts to incorporate uncertainty into transformer-based models have uncovered a connection between multi-head self-attention (MHSA) and Gaussian processes (GPs) which shows that the deterministic output of a MHSA block corresponds to the posterior mean of a GP conditioned on its input~\citep{chen2023calibrating,bui2025revisiting,chen2024self} (see Appendix~\ref{app:sgpa-kepsvgp}).~Although this insight offers a principled approach for uncertainty calibration at individual attention blocks, propagating it across MHSA blocks remain challenging.~The difficulty arises because GP-reparameterized MHSA is interleaved with point-estimated components such as feed-forward networks (MLP) and layer normalization (LN).~This disrupts the flow of uncertainty propagation since the interleaved sequence does not align with a well-defined stochastic path with a proper probabilistic transition model, particularly along the point-estimated segments of the pre-trained model.


\noindent To enable principled uncertainty propagation in pre-trained transformers, we instead propose a structural reconfiguration that reorganizes the model into a well-defined probabilistic path.~In particular, we repartition the architecture such that each transformation block ends with an MHSA block (see Fig.~\ref{fig:reconfiguration}), whose output can be interpreted as a Gaussian distribution over intermediate features.~This restructuring instead views the aforementioned point-estimated network segments as additional parameterization of a GP-reparameterized MHSA rather than observations of a latent probabilistic transition (see Section~\ref{subsec:cut}).~The resulting sequence of neuralized Gaussian transitions thus becomes well-aligned with the reverse-time stochastic process of a diffusion model.

These neuralized Gaussian transitions can then be viewed as discrete observations at different time steps of the diffusion's reverse-time process.~We can thus learn this process via distilling these observed transition across different timesteps into a unified spatiotemporal transition kernel.~This can be achieved via adopting variational inference as inspired by score-based diffusion methods~\citep{sohl2015deep,ho2020denoising}.~This reveals a novel reconfiguration of pre-trained transformers into uncertainty-aware diffusion processes that interestingly enables UC via learning generative models translating between raw data and predictive features (see Section~\ref{subsec:diffomer}).

\subsection{Reconfiguring Pre-Trained Transformer as Probability Path}
\label{subsec:cut}

\begin{figure*}
    \centering
    \includegraphics[width=\linewidth]{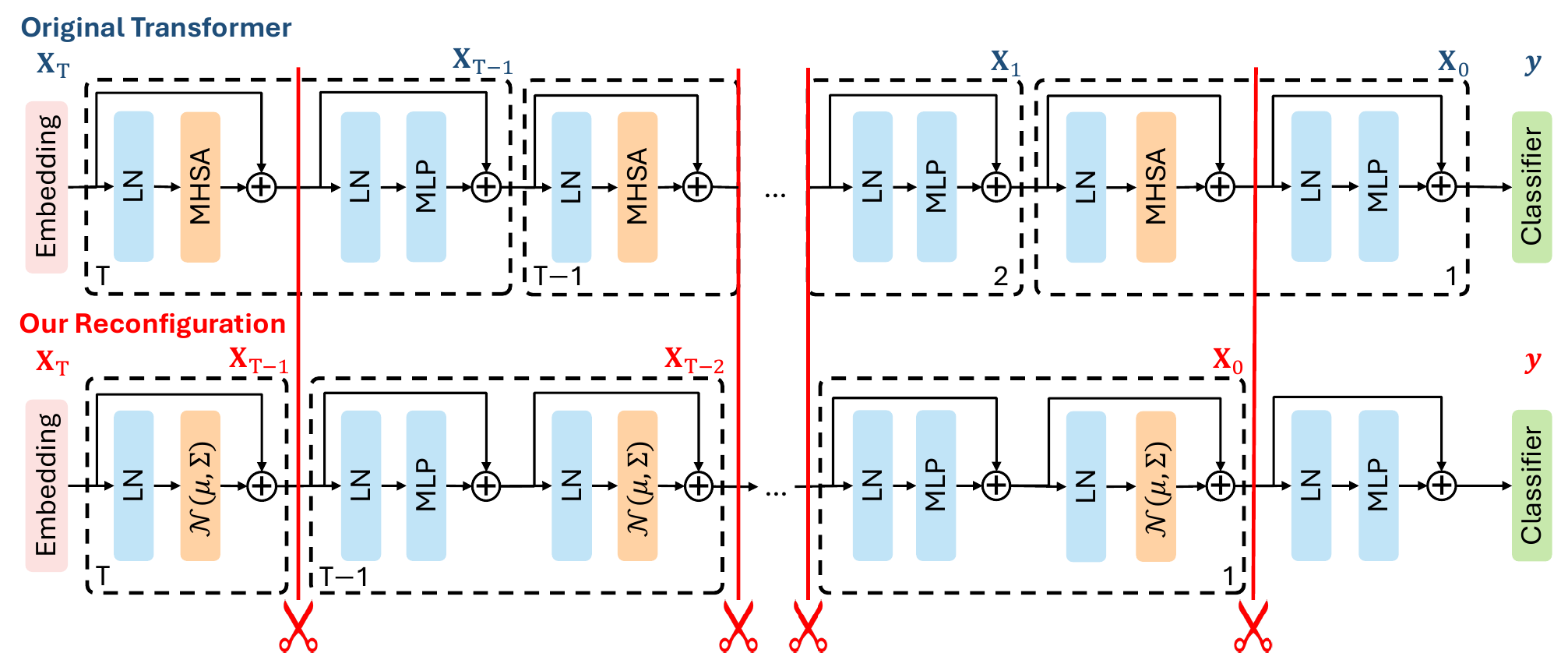}
    \caption{Restructuring a pre-trained transformer such that each block outputs a Gaussian distribution over its intermediate features, effectively aligning its architecture with a probabilistic path.}
    \label{fig:reconfiguration}
    \vspace{-10pt}
\end{figure*}

Following prior work on uncertainty-aware transformers~\citep{chen2023calibrating,bui2025revisiting,chen2024self}, the output of a kernelized attention head can be interpreted as the predictive mean at the input queries of a Gaussian process (GP) posterior conditioned on the key-value pairs.~This means kernelizing attention transforms the original MHSA mechanism into a GP-based variant that naturally incorporates calibrated uncertainty.~Each attention head thus admits a reparameterized Gaussian process (GP) structure which induces a neuralized Gaussian transition,
\begin{eqnarray}
\boldsymbol{F}^{(h)}_t \mid \boldsymbol{U}_t &\sim& \mathbb{N}\left(\bar{m}_t^{(h)}(\boldsymbol{U}_t),\  \bar{\sigma}_t^{(h)}(\boldsymbol{U}_t)\right)  \ ,
\end{eqnarray}
where $\boldsymbol{U}_t$ is the input of the $t$-th MHSA block whereas $\bar{m}_t^{(h)}(\boldsymbol{U}_t), \bar{\sigma}_t^{(h)}(\boldsymbol{U}_t)$ are neuralized mean and covariance functions for its $h$-th attention head under the reparameterization design (Appendix~\ref{app:sgpa-kepsvgp}).

The individual output representation $\boldsymbol{F}_t^{(h)}$ of each attention head $h$ of the $t$-th MHSA block can then be aggregated via a linear combination $\boldsymbol{O}_t$ which preserves the (neuralized) Gaussian structure:
\begin{eqnarray}
\boldsymbol{R}_{t} &=&\boldsymbol{O}_t\left[\boldsymbol{F}^{(1)}_t, \boldsymbol{F}^{(2)}_t, \ldots, \boldsymbol{F}^{(n)}_t\right] \quad \Rightarrow\quad  \boldsymbol{R}_{t} \mid \boldsymbol{U}_t \ \ \sim\ \  \mathbb{N}\Big(\bar{m}_t(\boldsymbol{U}_t), \bar{\sigma}_t(\boldsymbol{U}_t)\Big) \ ,  
\end{eqnarray}
where $\bar{m}_t(\boldsymbol{U}_t) \triangleq [\boldsymbol{O}_t \bar{m}_t^{(1)}(\boldsymbol{U}_t), \ldots, \boldsymbol{O}_t \bar{m}_t^{(n)}(\boldsymbol{U}_t)]$ and $\bar{\sigma}_t(\boldsymbol{U}_t) \triangleq \mathrm{blkdiag}[\boldsymbol{O}_t\bar{\sigma}_t^{(h)}(\boldsymbol{U}_t)\boldsymbol{O}_t^\top]$. 





This reparameterization thus reconfigures a pre-trained (point-estimate) MHSA block into probabilistic transition function with a uncertainty structure which can be optimized as in previous methods~\citep{chen2023calibrating,bui2025revisiting,chen2024self}.~This provides a principled handle for uncertainty calibration.~One can assess the local representation uncertainty via the (learned) predictive variance or generate uncertainty-aware representation samples to propagate downstream.~This propagation is however disrupted in existing approaches as mentioned previously due to the interleaving of MHSA with point-estimated components such as MLPs and layer normalization (LN).~To elaborate, the transformation from one intermediate representation $\boldsymbol{X}_{t-1}$ to the next $\boldsymbol{X}_t$ interleaves the MHSA mechanism with the MLP and LN mechanisms:
\begin{eqnarray}
\hspace{-24mm}\boldsymbol{X}_{t-1} &=& \mathrm{MLP}(\mathrm{LN}(\boldsymbol{Z}_t)) + \boldsymbol{Z}_t \quad \text{where}\quad \boldsymbol{Z}_t  \ \ = \ \ \mathrm{MHSA}\left(\mathrm{LN}(\boldsymbol{X}_t)\right) \ \ +\ \ \boldsymbol{X}_t \ ,   
\end{eqnarray}
where we number transformer block in reverse such that the first transformer block is indexed with $t = T$ and the last is indexed with $t = 1$, as illustrated in the upper part of Fig.~\ref{fig:reconfiguration}

\noindent To propagate uncertainty under this partition, the point-estimated network segment $\mathrm{MLP}(\mathrm{LN}(\boldsymbol{Z}_t))$ can be viewed as an observed function sampled from some function prior.~However, unlike the MHSA which can be viewed as a sampled function from a learnable Gaussian process (GP) prior as established in prior works, it remains unclear how to parameterize and learn a function prior for $\mathrm{MLP}(\mathrm{LN}(\boldsymbol{Z}_t))$ without the risk of prior misspecification.~Otherwise, treating it as a deterministic transition collapses the uncertainty structure and consequently disrupts uncertainty propagation.

\noindent To sidestep this technical challenge, we propose to instead view $\mathrm{MLP}(\mathrm{LN}(\boldsymbol{Z}_t))$ as an additional parameterization of the neuralized Gaussian transition induced by the GP-reparameterized MHSA.~This can be achieved via a rearrangement of transformer's computation blocks as detailed below:
\begin{eqnarray}
\label{eq:rearrangement}
\hspace{-6mm}\boldsymbol{X}_{t-1} &=& \mathrm{MHSA}(\mathrm{LN}(\boldsymbol{Z}_t)) \ \ +\ \ \boldsymbol{Z}_t \ \  \text{where}\ \  \boldsymbol{Z}_t  \ \ = \ \ \begin{cases} 
    \mathrm{MLP}\left(\mathrm{LN}(\boldsymbol{X}_t)\right) + \boldsymbol{X}_t, & \text{if} \  t \neq T \\
    \boldsymbol{X}_T, & \text{otherwise} 
\end{cases} .    
\end{eqnarray}
This reconfiguration guarantees that each re-partitioned computation block terminates with an MHSA module (see the lower part of Fig.~\ref{fig:reconfiguration}).~The deterministic transition $\mathrm{MLP}(\mathrm{LN}(\boldsymbol{Z}_t))$ now become additional parameters of the MHSA which can be reparameterized into a neuralized Gaussian transition.~Note that the skip connection does not break Gaussianity but only shifts the mean.~Consequently, this construction induces a stochastic process $\{\mathbf{X}_t\}_{t=0}^T$ with Gaussian transitions:
\begin{eqnarray}
p\Big(\boldsymbol{X}_{t-1} \mid \boldsymbol{X}_t\Big) = \mathbb{N}\Big(\boldsymbol{X}_{t-1} \mid  m_t(\boldsymbol{X}_t),\ \sigma_t(\boldsymbol{X}_t)\Big), \label{eq:reconfig}
\end{eqnarray}
where $m_t(\boldsymbol{X}_t) = \bar{m}_t(\mathrm{LN}(\boldsymbol{Z}_t)) + \boldsymbol{Z}_t$ and $\sigma_t(\boldsymbol{X}_t) = \bar{\sigma}_t(\mathrm{LN}(\boldsymbol{Z}_t))$ with $\boldsymbol{Z}_t$ is defined in Eq.~\ref{eq:rearrangement}.~These separately parameterized Gaussian transitions across timesteps can be distilled into a unified spatiotemporal Gaussian transition that defines the reverse-time process of a diffusion model as discussed in Section~\ref{subsec:diffomer}.~This unified parameterization enables seamless uncertainty propagation while explicitly encoding transition correlations across steps as desired.


\textbf{Remark.}~We note that the above reconfiguration does not alter the pre-trained computation, but changes how the point-estimated segment $\mathrm{MLP}(\mathrm{LN}(\boldsymbol{Z}_t))$ is interpreted. Instead of being treated as an observed function drawn from an unknown prior, it is parameterized as part of a Gaussian transition, revealing a learnable representation that is more amenable to uncertainty propagation.~For brevity, we also abuse the notation $\boldsymbol{X}/\boldsymbol{F}$ to denote $\mathrm{vec}(\boldsymbol{X})/\mathrm{vec}(\boldsymbol{F})$.

\subsection{Distilling Transformer-based Probability Path on Diffusion Model} 
\label{subsec:diffomer}
Under our proposed reconfiguration in Section~\ref{subsec:cut}, the transformer induces a stochastic path $\{\mathbf{X}_t\}_{t=0}^T$ with Gaussian transitions (Eq.~\ref{eq:reconfig}), which closely resembles a reverse diffusion process.~This defines a distribution over intermediate features conditioned on the original input embedding $\boldsymbol{X}_T$:
\begin{eqnarray}
\hspace{-14.5mm}p\big(\mathbf{X}_{T-1}, \ldots, \mathbf{X}_0 \mid \boldsymbol{X}_T\big) \hspace{-1mm}&=&\hspace{-1mm} \prod_{t=1}^T p\big(\mathbf{X}_{t-1}\mid \mathbf{X}_t\big) \ = \ \prod_{t=1}^T \mathbb{N}\Big(\boldsymbol{X}_{t-1} \mid  m_t(\boldsymbol{X}_t),\ \sigma_t(\boldsymbol{X}_t)\Big)\ ,
\end{eqnarray}
where the transition probability $p(\mathbf{X}_{t-1} \mid \mathbf{X}_t)$ is modeled independently at each timestep. Calibrating uncertainty in this block-wise, decoupled structure is difficult as it does not capture transition correlation and hence does not generalize across timesteps.

\noindent To address this limitation, we require a more parsimonious representation that characterizes the entire sequence of transition models in a unified manner.~This can be achieved by re-compiling it into a reverse-time diffusion process with a unified spatiotemporal transition model:
\begin{eqnarray}
\hspace{-12mm}q_\theta\big(\mathbf{X}_{T-1}, \ldots, \mathbf{X}_0 \mid \boldsymbol{X}_T\big) \hspace{-1mm}&=&\hspace{-1mm} \prod_{t=1}^T q_\theta\big(\mathbf{X}_{t-1}\mid \mathbf{X}_t\big) \ = \ \prod_{t=1}^T \mathbb{N}\Big(\boldsymbol{X}_{t-1} \mid  m_\theta(\boldsymbol{X}_t),\ \sigma_\theta(\boldsymbol{X}_t)\Big)\ ,
\end{eqnarray}
In particular, the entire sequence of neuralized Gaussian transitions derived from the previously described GP-reparameterized of pre-trained transformer can be absorbed into the reverse-time diffusion with a unified spatiotemporal transition via minimizing the following negative log-likelihood, analogous to score matching in diffusion models~\citep{sohl2015deep,ho2020denoising}:
\begin{eqnarray}
L(\theta) &=& \mathbb{E}_{p(\boldsymbol{X}_0 \mid \boldsymbol{X}_T)}\Big[-\log q_\theta\Big(\boldsymbol{X}_0 \mid \boldsymbol{X}_T\Big)\Big]  \ .
\end{eqnarray}
This negative log-likelihood (NLL) loss admits the following upper-bound via variational inference:
\begin{eqnarray}
\hspace{-13mm}L(\theta) &\leq& \mathbb{H}\Big(p(\mathbf{X}_0 \mid \mathbf{X}_T)\Big) \ +\  \sum_{t=1}^T \mathbb{E}_{p(\mathbf{X}_t \mid \mathbf{X}_T)} \Big[ D_{\mathrm{KL}}\Big(p(\mathbf{X}_{t-1} | \mathbf{X}_t) \ \| \ q_\theta(\mathbf{X}_{t-1} \mid \mathbf{X}_t)\Big) \Big] \ , \label{eq:bound}
\end{eqnarray}
with proof deferred to Appendix~\ref{app:vlb_loss}. Since the entropy term $\mathbb{H}\big(p(\mathbf{X}_0 \mid \mathbf{X}_T)\big)$ is independent of $\theta$, optimizing the bound in Eq.~\ref{eq:bound} reduces to minimizing the Kullback-Leibler (KL) divergence:
\begin{eqnarray}
L_1(\theta) &=&  \underset{t\ \sim\  \mathbb{U}(1,T)} {\mathbb{E}} \ \ \underset{\boldsymbol{X}_t\sim p(\mathbf{X}_t \mid \mathbf{X}_T)}{\mathbb{E}}\Big[D_{\mathrm{KL}}\Big(p(\boldsymbol{X}_{t-1}\mid \boldsymbol{X}_t) \ \| \ q_\theta(\boldsymbol{X}_{t-1}\mid \boldsymbol{X}_t)\Big)\Big] \ ,
\label{eq:original-loss}
\end{eqnarray}
where $t \sim \mathbb{U}(1, T)$ and $\boldsymbol{X}_t$ is sampled via sampling data $\boldsymbol{X}$ and simulating the corresponding output of the $(T - t)$-th block of the pre-trained transformer.
This loss aligns the probability path with a diffusion-style transition kernel while enabling generalization across timesteps. To ensure that the learned uncertainty propagation process maps from data to feature distributions which are informative for downstream prediction, we regularize it with an additional performance loss:
\begin{eqnarray}
L_2(\theta) = \underset{(\boldsymbol{X}, \boldsymbol{y}) \sim \boldsymbol{D} \ \ }{\mathbb{E}}\underset{\boldsymbol{X}_0 \sim q_\theta(\boldsymbol{X}_0 \mid \boldsymbol{X}_T)}{\mathbb{E}}\Big[\mathrm{loss}\Big(\boldsymbol{X}_0,\boldsymbol{y}\Big)\Big]  \ ,
\end{eqnarray}
where $\boldsymbol{X}$ is sampled from the training dataset $\boldsymbol{D}$ and is embedded with $\boldsymbol{X}_T = \mathrm{embed}(\boldsymbol{X})$. $\boldsymbol{X}_0$ is then sampled via iteratively simulating the current estimate of the probability path $q_\theta(\boldsymbol{X}_{t-1}\mid \boldsymbol{X}_t)$.~The parameterization of the unified spatiotemporal transition model can then be obtained via:
\begin{eqnarray}
\label{eq:loss-l1+l2}
\theta &=& \argmin_{\theta}\ \ \Big\{ L_1(\theta) \ \ +\ \  L_2(\theta) \Big\} \ ,
\end{eqnarray}
which combines the uncertainty-aware (reconfiguration) loss with the performance-aware loss.~For implementation details of the above algorithm, please refer to Appendix~\ref{app:loss-derivation}.

\section{Experiments}
\label{sec:experiments}
This section evaluates the efficacy of our proposed method, \ourmethod: \underline{\textbf{D}}iffusion-\underline{\textbf{I}}nspired \underline{\textbf{REC}}onfiguration of \underline{\textbf{T}}ransf\underline{\textbf{OR}}mers for Uncertainty Calibration, by reconfiguring existing uncertainty-aware transformers into diffusion-based models and comparing their uncertainty calibration and predictive performance against those of the original versions.~We describe our experiment settings in Section~\ref{sec:exp_setting} and report detailed empirical results in Section~\ref{sec:result1}.

\subsection{Experiment Settings}
\label{sec:exp_setting}

\textbf{Datasets.}~We evaluate~\ourmethod{} using datasets in computer vision (CV) and natural language processing (NLP). In CV, we use the CIFAR-10 and CIFAR-100 datasets~\citep{krizhevsky2009learning}.~Each dataset contains 45,000 training, 5000 validation, and 10,000 test images.~In NLP, we use the IMDB dataset~\citep{maas2011learning}, with 20,000 training, 5,000 validation, and 25,000 test samples; and the CoLA dataset~\citep{warstadt2019neural}, with 6,355 training, 907 validation, and 1,816 test samples.

\textbf{Baselines}.~We compare \ourmethod{} with various uncertainty-aware baselines: Temperature Scaling (TS)~\citep{guo2017calibration}, Monte Carlo Dropout (MCD)~\citep{gal2016dropout}, Stochastic Variational Deep Kernel Learning (SV-DKL)~\citep{wilson2016stochastic}, Kronecker-Factored Last-Layer Laplace Approximation (KFLLA)~\citep{kristiadi2020being}, Sparse Gaussian Process Attention (SGPA)~\citep{chen2023calibrating}, and Kernel-Eigen Pair Sparse Variational Gaussian Processes Attention (KEP-SVGP or KEP for brevity)~\citep{chen2024self}.

\textbf{Pre-Trained Models.}~We conduct our uncertainty-aware reconfiguration experiments on two pre-trained architectures which include (i) vanilla transformer-based model and (ii) uncertainty-aware transformer KEP~\citep{chen2024selfattention}.~For each architecture, we pre-train a 7-layer vision transformer (ViT~\citep{dosovitskiy2020image}) for experiments on CIFAR-10 and CIFAR-100, and a 5-layer transformer~\citep{vaswani2017attention} for CoLA and IMDB. We also evaluate variants of uncertainty-aware transformer KEP~\citep{chen2024selfattention}, where the first $n-k$ attention blocks use standard MHSA and the last $k$ blocks use GP-reparameterized attention; we denote this variant as KEP-$k$/$n$. 

\textbf{Unified Transition Model.}~We implement the unified spatiotemporal transition model in our diffusion-based reparameterization using a single-block DiT~\citep{Peebles2022DiT}, with embedding dimensions matched to the pre-trained models (see Appendix~\ref{subsubsec:utm-training} for full hyperparameters). The model is designed to be smaller than the original backbone for improved memory efficiency, containing only 2.7M parameters versus 6.24M in ViT for vision tasks, and 2.59M versus 3.38M in the text transformer for NLP tasks. Results for alternative transition model architectures are reported in Appendix~\ref{subsubsec:varying-utm}.

\begin{table*}[t]
    \caption{Performance comparison between pre-trained transformers and their diffusion-inspired reconfiguration using~\ourmethod{} on in-distribution classification tasks. KEP-$k$/$n$ denotes a pre-trained transformer using GP-reparameterized architecture (KEP~\citep{chen2024self}) for the last $k$ attention blocks and standard MHSA for the remaining blocks.~Better results are shown in {\bf bold}.}
    \label{tab::in_dist_kep_diff}
    \begin{center}
    \resizebox{0.95\textwidth}{!}{
    \begin{tabular}{|c|l|lllllll|}
        \toprule
        \textbf{Dataset} & \textbf{Method} & \textbf{ACC/MCC} $\uparrow$ & \textbf{AURC} $\downarrow$ & \textbf{AUROC} $\uparrow$ & \textbf{FPR95} $\downarrow$ & \textbf{ECE} $\downarrow$ & \textbf{NLL} $\downarrow$ & \textbf{Brier} $\downarrow$ 
        \\ \midrule
        \multirow{6}{*}{\rotatebox{90}{\parbox{2cm}{\centering CIFAR-10}}} 
        & ViT       
        & 83.84$\pm$0.09 & 4.05$\pm$0.11 & 86.42$\pm$0.37 & 67.13$\pm$1.98 & 12.51$\pm$0.20 & 10.91$\pm$0.39 & 28.03$\pm$0.15 \\
        & \ourmethod         
        & \textbf{85.67}$\pm$0.88 & \textbf{3.10}$\pm$0.45 & \textbf{87.90}$\pm$1.04 & \textbf{61.92}$\pm$1.85 & \textbf{9.67}$\pm$0.62 & \textbf{6.94}$\pm$0.52 & \textbf{23.54}$\pm$1.46 \\
        \cmidrule(lr){2-9}
        & KEP-1/7       
        & 84.52$\pm$0.25 & 3.52$\pm$0.11 & 87.52$\pm$0.27 & 65.14$\pm$1.27 & 10.93$\pm$0.26 & 8.21$\pm$0.15 & 25.80$\pm$0.40 \\
        & \ourmethod              
        & \textbf{86.51}$\pm$0.57 & \textbf{2.78}$\pm$0.16 & \textbf{88.28}$\pm$0.16 & \textbf{61.85}$\pm$1.93 & \textbf{8.61}$\pm$0.49 & \textbf{6.11}$\pm$0.27 & \textbf{21.90}$\pm$0.82 \\
        \cmidrule(lr){2-9} 
        & KEP-7/7       
        & 82.68$\pm$0.12 & 4.46$\pm$0.05 & 85.71$\pm$0.34 & 66.90$\pm$1.78 & 6.95$\pm$0.36 & 5.89$\pm$0.11 & 25.90$\pm$0.23 \\
        & \ourmethod
        & \textbf{84.74}$\pm$0.40 & \textbf{{3.47}}$\pm$0.18 & \textbf{{87.34}}$\pm$0.33 & \textbf{{62.88}}$\pm$1.20 & \textbf{5.18}$\pm$0.33 & \textbf{4.84}$\pm$0.11 & \textbf{{22.48}}$\pm$0.53 \\
        \midrule
        \multirow{6}{*}{\rotatebox{90}{\parbox{2cm}{\centering CIFAR-100}}}
        & ViT       
        & 52.94$\pm$0.63 & 22.88$\pm$0.64 & 81.07$\pm$0.56 & 75.96$\pm$2.39 & 30.73$\pm$0.61 & 32.71$\pm$0.91 & 74.72$\pm$1.20 \\
        & \ourmethod              
        & \textbf{57.79}$\pm$0.57 & \textbf{18.58}$\pm$0.37 & \textbf{82.60}$\pm$0.11 & \textbf{71.50}$\pm$1.42 & \textbf{22.31}$\pm$0.49 & \textbf{22.16}$\pm$0.36 & \textbf{62.57}$\pm$0.85 \\
        \cmidrule(lr){2-9}
        & KEP-1/7      
        & 55.74$\pm$0.77 & 20.20$\pm$0.64 & 82.10$\pm$0.20 & 73.82$\pm$0.92 & 27.07$\pm$0.71 & 27.45$\pm$0.62 & 68.54$\pm$1.18 \\
        & \ourmethod              
        & \textbf{58.78}$\pm$1.52 & \textbf{17.63}$\pm$1.08 & \textbf{{82.99}}$\pm$0.41 & \textbf{{71.40}}$\pm$1.24 & \textbf{21.17}$\pm$1.03 & \textbf{21.15}$\pm$0.93 & \textbf{60.96}$\pm$1.73 \\
        \cmidrule(lr){2-9}
        & KEP-7/7      
        & 57.06$\pm$0.56 & 19.38$\pm$0.60 & 82.02$\pm$0.39 & 72.78$\pm$0.69 & \textbf{21.31}$\pm$3.85 & 22.41$\pm$2.29 & 63.07$\pm$2.11 \\
        & \ourmethod              
        & \textbf{{60.85}}$\pm$2.98 & \textbf{{16.35}}$\pm$2.34 & \textbf{82.91}$\pm$0.67 & \textbf{71.46}$\pm$1.43 & 21.43$\pm$2.09 & \textbf{{20.55}}$\pm$2.42 & \textbf{{58.91}}$\pm$4.63 \\
        \midrule
        \multirow{6}{*}{\rotatebox{90}{\parbox{2cm}{\centering IMDB}}} 
        & Transformer     
        & 85.59$\pm$0.50 & 4.73$\pm$0.32 & 80.75$\pm$0.55 & 75.45$\pm$1.10 & 6.96$\pm$2.05 & 3.95$\pm$0.44 & 22.28$\pm$1.26 \\
        & \ourmethod              
        & \textbf{86.07}$\pm$0.61 & \textbf{4.57}$\pm$0.39 & \textbf{80.84}$\pm$0.67 & \textbf{74.24}$\pm$0.87 & \textbf{5.40}$\pm$1.90 & \textbf{3.60}$\pm$0.34 & \textbf{21.08}$\pm$1.32 \\
        \cmidrule(lr){2-9}
        & KEP-1/5    
        & 85.76$\pm$0.71 & 4.54$\pm$0.42 & 81.02$\pm$0.70 & 74.87$\pm$0.87 & 5.51$\pm$2.94 & 3.79$\pm$0.51 & 21.62$\pm$1.42 \\
        & \ourmethod              
        & \textbf{{87.13}}$\pm$0.19 & \textbf{{4.07}}$\pm$0.30 & \textbf{{81.55}}$\pm$0.62 & \textbf{{73.35}}$\pm$0.14 & \textbf{3.16}$\pm$2.54 & \textbf{{3.24}}$\pm$0.31 & \textbf{{19.31}}$\pm$0.97 \\
        \cmidrule(lr){2-9}
        & KEP-5/5     
        & 84.57$\pm$0.81 & 5.48$\pm$0.60 & 79.32$\pm$1.25 & 77.03$\pm$1.48 & 7.83$\pm$3.28 & 5.17$\pm$2.13 & 24.23$\pm$2.56 \\
        & \ourmethod              
        & \textbf{85.74}$\pm$0.34 & \textbf{4.58}$\pm$0.23 & \textbf{80.95}$\pm$0.42 & \textbf{75.08}$\pm$1.15 & \textbf{{2.33}}$\pm$1.54 & \textbf{3.36}$\pm$0.12 & \textbf{20.70}$\pm$0.65 \\
        \midrule
        \multirow{6}{*}{\rotatebox{90}{\parbox{2cm}{\centering CoLA}}}
        & Transformer       
        & 29.92$\pm$1.17 & 20.80$\pm$1.21 & 64.22$\pm$1.46 & 90.01$\pm$2.84 & 26.44$\pm$1.90 & 19.66$\pm$4.18 & 55.09$\pm$2.68 \\
        & \ourmethod              
        & \textbf{31.85}$\pm$2.46 & \textbf{19.74}$\pm$1.62 & \textbf{64.52}$\pm$1.71 & \textbf{89.52}$\pm$3.61 & \textbf{23.94}$\pm$0.49 & \textbf{14.29}$\pm$3.01 & \textbf{50.82}$\pm$1.19 \\
        \cmidrule(lr){2-9}
        & KEP-1/5      
        & 30.86$\pm$2.03 & 19.86$\pm$2.04 & {65.18}$\pm$1.83 & \textbf{{89.10}}$\pm$2.60 & 24.98$\pm$2.08 & 16.28$\pm$4.35 & 52.86$\pm$2.88 \\
        & \ourmethod              
        & \textbf{31.84}$\pm$1.88 & \textbf{19.42}$\pm$1.83 & \textbf{{65.54}}$\pm$1.78 & 90.37$\pm$0.89 & \textbf{13.77}$\pm$6.58 & \textbf{8.26}$\pm$3.75 & \textbf{43.54}$\pm$3.65 \\
        \cmidrule(lr){2-9} 
        & KEP-5/5     
        & 29.28$\pm$1.21 & 20.82$\pm$1.98 & 64.47$\pm$0.90 & 89.65$\pm$1.04 & 18.95$\pm$5.66 & 11.83$\pm$7.96 & 48.65$\pm$5.84 \\
        & \ourmethod              
        & \textbf{{32.05}}$\pm$1.56 & \textbf{{18.69}}$\pm$1.39 & \textbf{64.71}$\pm$1.37 & \textbf{89.53}$\pm$1.67 & \textbf{{12.88}}$\pm$5.74 & \textbf{7.68}$\pm$1.70 & \textbf{{42.20}}$\pm$2.50 \\
        \bottomrule
    \end{tabular}}
    \vspace{-10pt}
    \end{center}
\end{table*}

\textbf{Evaluation Metrics.} For in-distribution classification, we evaluate predictive performance using accuracy (ACC) for CIFAR-10, CIFAR-100, and IMDB, and Matthew’s Correlation Coefficient (MCC) for CoLA. Calibration is assessed with Negative Log-Likelihood (NLL $\times$ 10), Expected Calibration Error (ECE~\%), and Brier Score (\%). Failure prediction is measured using Area Under the Risk-Coverage Curve (AURC~\%), Area Under the Receiver Operating Characteristic Curve (AUROC~\%), and False Positive Rate at 95\% True Positive Rate (FPR95~\%). For out-of-distribution (OOD) robustness, the same metrics are applied to the CIFAR-10-C and CoLA-OOD datasets. OOD detection performance is quantified using AUROC~\% and Area Under the Precision-Recall Curve (AUPR~\%). All metrics are reported as mean $\pm$ standard error over five runs.~All experiments are conducted on 8 NVIDIA L40 GPUs.

\subsection{Results and Discussion}
\label{sec:result1}

\subsubsection{In-Distribution Classification}

\textbf{Comparison with Pre-Trained Models.} \ourmethod{} demonstrates superior performance compared to pre-trained models across nearly all tasks and metrics (Table \ref{tab::in_dist_kep_diff}).~On CIFAR-10, \ourmethod{} outperforms KEP-7/7 on every reported metric, with higher accuracy (84.74\% versus 82.68\%), lower ECE (5.18 versus 6.95), lower NLL (4.84 versus 5.89), and a lower Brier score (22.48 versus 25.90), while also leading in the failure-prediction metrics.~On CIFAR-100, although \ourmethod{} does not outperform KEP-7/7 in ECE (21.43 versus 21.31), the difference is marginal ($< 0.2\%$), while \ourmethod{} achieves significantly higher accuracy (60.85\% versus 57.06\%) and a lower Brier score. Additionally, significance testing, reported in Appendix~\ref{subsubsec:signigicance-testing}, indicates that most improvements are statistically significant ($p$-value $< 0.05$).~These results demonstrate that \ourmethod{} not only provides better-calibrated uncertainty estimates but also enhances predictive accuracy, underscoring the effectiveness of our propagation-based model for UC.


\textbf{Comparison with Existing Uncertainty-Aware Baselines.} Beyond vanilla pre-trained models, \ourmethod{} also surpasses other uncertainty-aware approaches in both predictive performance and uncertainty calibration across multiple tasks (Table~\ref{tab::in_dist_baselines_diff}). For a fair comparison, \ourmethod{} and KEP are configured with optimal settings and evaluated against standard baselines.~Overall, \ourmethod{} achieves the \textcolor{blue}{\textbf{highest}} performance in 20/28 settings (across $4$ datasets and $7$ UC/performance metrics) and the \textcolor{brown}{\textbf{second-highest}} in 6/28 settings, establishing a new state-of-the-art in UC.

\begin{table*}[t]
    \caption{Comparison of average performance achieved by the diffusion-based reconfigured KEP~\citep{chen2024self} produced by~\ourmethod{} and other uncertainty-aware baselines on in-distribution classification tasks.~\textbf{\textcolor{blue}{Blue}} marks the best result across all baselines for a dataset while \textbf{\textcolor{brown}{brown}} denotes the second-best. \textbf{$\uparrow$} indicates that higher values are better, while \textbf{$\downarrow$} indicates that lower values are better. }
    \label{tab::in_dist_baselines_diff}
    \begin{center}
    \resizebox{0.95\textwidth}{!}{
    \begin{tabular}{|c|l|lllllll|}
        \toprule
        \textbf{Dataset} & \textbf{Method} & \textbf{ACC/MCC} $\uparrow$ & \textbf{AURC} $\downarrow$ & \textbf{AUROC} $\uparrow$ & \textbf{FPR95} $\downarrow$ & \textbf{ECE} $\downarrow$ & \textbf{NLL} $\downarrow$ & \textbf{Brier} $\downarrow$ 
        \\ \midrule
        \multirow{8}{*}{\rotatebox{90}{\parbox{2cm}{\centering CIFAR-10}}} 
        & TS 
        & 83.84$\pm$0.09 & 3.88$\pm$0.10 & 86.82$\pm$0.37 & 65.99$\pm$1.94 & 9.22$\pm$0.36 & 6.58$\pm$0.16 & 25.50$\pm$0.13 \\
        & MCD 
        & 84.06$\pm$0.23 & 8.65$\pm$0.03 & 86.51$\pm$0.32 & 66.15$\pm$0.60 & 9.47$\pm$0.16 & 8.36$\pm$0.32 & 25.45$\pm$0.29 \\
        & KFLLA 
        & 83.84$\pm$0.10 & 3.91$\pm$0.11 & 86.71$\pm$0.45 & 65.44$\pm$1.58 & 8.18$\pm$0.80 & \textcolor{brown}{\textbf{6.09}}$\pm$0.40 & \textcolor{brown}{\textbf{24.98}}$\pm$0.59 \\
        & SV-DKL 
        & 83.23$\pm$0.17 & 4.39$\pm$0.18 & 85.94$\pm$0.36 & 66.96$\pm$1.30 & 11.64$\pm$0.81 & 9.85$\pm$1.09 & 27.97$\pm$0.77 \\
        & SGPA 
        & 75.59$\pm$3.63 & 8.41$\pm$2.37 & 82.65$\pm$1.71 & 71.78$\pm$2.73 & \textcolor{blue}{\textbf{1.92}}$\pm$0.55 & 7.11$\pm$0.95 & 33.98$\pm$4.57 \\
        & ViT       
        & 83.84$\pm$0.09 & 4.05$\pm$0.11 & 86.42$\pm$0.37 & 67.13$\pm$1.98 & 12.51$\pm$0.20 & 10.91$\pm$0.39 & 28.03$\pm$0.15 \\
        & KEP
        & \textcolor{brown}{\textbf{84.52}}$\pm$0.25 & \textcolor{brown}{\textbf{3.52}}$\pm$0.11 & \textcolor{blue}{\textbf{87.52}}$\pm$0.27 & \textcolor{brown}{\textbf{65.14}}$\pm$1.27 & 10.93$\pm$0.26 & 8.21$\pm$0.15 & 25.80$\pm$0.40 \\
        \cmidrule(lr){2-9}
        & \ourmethod              
        & \textcolor{blue}{{\textbf{84.74}}}$\pm$0.40 & \textcolor{blue}{{\textbf{3.47}}}$\pm$0.18 & \textcolor{brown}{\textbf{87.34}}$\pm$0.33 & \textcolor{blue}{{\textbf{62.88}}}$\pm$1.20 & \textcolor{brown}{\textbf{5.18}}$\pm$0.33 & \textcolor{blue}{\textbf{4.84}}$\pm$0.11 & \textcolor{blue}{{\textbf{22.48}}}$\pm$0.53 \\
        \midrule
        \multirow{8}{*}{\rotatebox{90}{\parbox{2cm}{\centering CIFAR-100}}}
        & TS 
        & 52.94$\pm$0.63 & 22.34$\pm$0.61 & \textcolor{brown}{\textbf{82.29}}$\pm$0.48 & 71.65$\pm$1.98 & \textcolor{brown}{\textbf{17.06}}$\pm$0.42 & 21.57$\pm$0.52 & 64.77$\pm$0.95 \\
        & MCD 
        & 53.49$\pm$0.62 & 22.24$\pm$0.56 & 81.60$\pm$0.19 & 73.02$\pm$0.51 & 25.93$\pm$0.37 & 29.24$\pm$0.73 & 70.02$\pm$0.92 \\
        & KFLLA 
        & 52.27$\pm$0.86 & 23.96$\pm$0.78 & 81.30$\pm$0.48 & \textcolor{blue}{\textbf{71.42}}$\pm$1.92 & 18.52$\pm$5.40 & 20.89$\pm$0.57 & 66.51$\pm$1.66 \\
        & SV-DKL
        & 51.03$\pm$0.60 & 24.38$\pm$0.43 & 81.32$\pm$0.50 & 73.99$\pm$1.40 & 25.46$\pm$0.72 & 28.93$\pm$0.66 & 71.90$\pm$0.74 \\
        & SGPA 
        & 52.77$\pm$0.52 & 22.84$\pm$0.52 & 81.65$\pm$0.36 & 72.02$\pm$1.74 & \textcolor{blue}{\textbf{10.33}}$\pm$2.25 & \textcolor{blue}{\textbf{19.10}}$\pm$0.57 & \textcolor{brown}{\textbf{62.08}}$\pm$1.04 \\
        & ViT       
        & 52.94$\pm$0.63 & 22.88$\pm$0.64 & 81.07$\pm$0.56 & 75.96$\pm$2.39 & 30.73$\pm$0.61 & 32.71$\pm$0.91 & 74.72$\pm$1.20 \\
        & KEP 
        & \textcolor{brown}{\textbf{57.06}}$\pm$0.56 & \textcolor{brown}{\textbf{19.38}}$\pm$0.60 & 82.02$\pm$0.39 & 72.78$\pm$0.69 & {21.31}$\pm$3.85 & 22.41$\pm$2.29 & 63.07$\pm$2.11 \\
        \cmidrule(lr){2-9}
        & \ourmethod              
        & \textcolor{blue}{{\textbf{60.85}}}$\pm$2.98 & \textcolor{blue}{{\textbf{16.35}}}$\pm$2.34 & \textcolor{blue}{{\textbf{82.91}}}$\pm$0.67 & \textcolor{brown}{\textbf{71.46}}$\pm$1.43 & 21.43$\pm$2.09 & \textcolor{brown}{{\textbf{20.55}}}$\pm$2.42 & \textcolor{blue}{{\textbf{58.91}}}$\pm$4.63 \\
        \midrule
        \multirow{8}{*}{\rotatebox{90}{\parbox{2cm}{\centering IMDB}}} 
        & TS
        & 85.59$\pm$0.50 & 4.73$\pm$0.32 & 80.75$\pm$0.55 & 75.45$\pm$1.10 & \textcolor{blue}{\textbf{2.91}}$\pm$1.51 & \textcolor{brown}{\textbf{3.41}}$\pm$0.13 & 21.04$\pm$0.74 \\
        & MCD
        & \textcolor{brown}{\textbf{85.96}}$\pm$0.42 & \textcolor{brown}{\textbf{4.40}}$\pm$0.24 & \textcolor{brown}{\textbf{81.40}}$\pm$0.55 & \textcolor{brown}{\textbf{74.79}}$\pm$0.88 & 4.18$\pm$2.03 & 3.47$\pm$0.23 & \textcolor{brown}{\textbf{20.72}}$\pm$0.82 \\
        & KFLLA
        & 85.59$\pm$0.50 & 4.71$\pm$0.30 & 80.82$\pm$0.48 & 75.45$\pm$1.11 & 5.84$\pm$2.21 & 6.93$\pm$0.00 & 21.86$\pm$1.19 \\
        & SV-DKL
        & 85.69$\pm$0.66 & 5.58$\pm$0.79 & 78.54$\pm$2.20 & 75.32$\pm$0.84 & 8.52$\pm$1.57 & 4.49$\pm$0.65 & 23.10$\pm$1.66 \\
        & SGPA
        & 85.39$\pm$0.36 & 4.96$\pm$0.49 & 80.04$\pm$1.14 & 76.44$\pm$0.96 & 6.04$\pm$1.71 & 3.96$\pm$0.50 & 22.19$\pm$1.06 \\
        & Transformer     
        & 85.59$\pm$0.50 & 4.73$\pm$0.32 & 80.75$\pm$0.55 & 75.45$\pm$1.10 & 6.96$\pm$2.05 & 3.95$\pm$0.44 & 22.28$\pm$1.26 \\
        & KEP
        & 85.76$\pm$0.71 & 4.54$\pm$0.42 & 81.02$\pm$0.70 & 74.87$\pm$0.87 & 5.51$\pm$2.94 & 3.79$\pm$0.51 & 21.62$\pm$1.42 \\
        \cmidrule(lr){2-9}
        & \ourmethod              
        & \textcolor{blue}{{\textbf{87.13}}}$\pm$0.19 & \textcolor{blue}{{\textbf{4.07}}}$\pm$0.30 & \textcolor{blue}{{\textbf{81.55}}}$\pm$0.62 & \textcolor{blue}{{\textbf{73.35}}}$\pm$0.14 & \textcolor{brown}{\textbf{3.16}}$\pm$2.54 & \textcolor{blue}{{\textbf{3.24}}}$\pm$0.31 & \textcolor{blue}{{\textbf{19.31}}}$\pm$0.97 \\
        \midrule
        \multirow{8}{*}{\rotatebox{90}{\parbox{2cm}{\centering CoLA}}}
        & TS 
        & 29.92$\pm$1.17 & 20.84$\pm$1.23 & 64.31$\pm$1.44 & 89.93$\pm$2.95 & \textcolor{brown}{\textbf{23.22}}$\pm$2.99 & \textcolor{brown}{\textbf{11.04}}$\pm$1.91 & \textcolor{brown}{\textbf{51.70}}$\pm$3.42 \\
        & MCD 
        & 30.04$\pm$1.02 & 20.66$\pm$1.11 & 64.53$\pm$1.00 & 89.51$\pm$1.35 & 24.96$\pm$1.79 & 17.83$\pm$3.61 & 53.52$\pm$2.59 \\
        & KFLLA 
        & 29.89$\pm$1.14 & 20.82$\pm$1.26 & 64.22$\pm$1.46 & 89.87$\pm$3.24 & 24.36$\pm$2.25 & 12.16$\pm$1.49 & 52.80$\pm$2.85 \\
        & SV-DKL 
        & 30.07$\pm$1.41 & 22.76$\pm$2.28 & 61.98$\pm$3.09 & \textcolor{blue}{\textbf{89.00}}$\pm$2.55 & 25.71$\pm$1.60 & 17.96$\pm$3.26 & 54.40$\pm$2.13 \\
        & SGPA 
        & \textcolor{brown}{\textbf{31.53}}$\pm$2.05 & 20.44$\pm$2.60 & 64.34$\pm$1.95 & 90.79$\pm$0.87 & 26.22$\pm$1.51 & 28.65$\pm$7.23 & 54.08$\pm$2.44 \\
        & Transformer       
        & 29.92$\pm$1.17 & 20.80$\pm$1.21 & 64.22$\pm$1.46 & 90.01$\pm$2.84 & 26.44$\pm$1.90 & 19.66$\pm$4.18 & 55.09$\pm$2.68 \\
        & KEP   
        & 30.86$\pm$2.03 & \textcolor{brown}{\textbf{19.86}}$\pm$2.04 & \textcolor{blue}{\textbf{65.18}}$\pm$1.83 & {\textcolor{brown}{\textbf{89.10}}}$\pm$2.60 & 24.98$\pm$2.08 & 16.28$\pm$4.35 & 52.86$\pm$2.88 \\
        \cmidrule(lr){2-9}
        & \ourmethod              
        & \textcolor{blue}{{\textbf{32.05}}}$\pm$1.56 & \textcolor{blue}{{\textbf{18.69}}}$\pm$1.39 & \textcolor{brown}{\textbf{64.71}}$\pm$1.37 & {89.53}$\pm$1.67 & \textcolor{blue}{{\textbf{12.88}}}$\pm$5.74 & \textcolor{blue}{{\textbf{7.68}}}$\pm$1.70 & \textcolor{blue}{{\textbf{42.20}}}$\pm$2.50 \\
        \bottomrule
    \end{tabular}}
    \vspace{-10pt}
    \end{center}
\end{table*}

\begin{table}[t]
\centering
\begin{minipage}{0.485\linewidth}
\centering
\caption{Comparison on CIFAR10-C.}\vspace{2mm}
\label{tab::dis_shift}
\setlength\tabcolsep{3pt} 
\resizebox{\columnwidth}{!}{
\begin{tabular}{|l|llll|}
    \toprule
    \textbf{Method} 
    & \textbf{ACC} $\uparrow$ 
    & \textbf{ECE} $\downarrow$ 
    & \textbf{NLL} $\downarrow$ 
    & \textbf{Brier} $\downarrow$ 
    \\ \midrule
    ViT             
    & \textbf{69.67}$\pm$0.34 & 24.30$\pm$0.31 & 23.59$\pm$1.00 & 53.07$\pm$0.59 
    \\
    \ourmethod   
    & 68.89$\pm$1.44 & \textbf{22.32}$\pm$1.09 & \textbf{17.71}$\pm$1.02 & \textbf{51.77}$\pm$2.39
    \\ \midrule
    KEP-1/7
    & \textbf{69.87}$\pm$0.45 & 22.12$\pm$0.47 & 18.54$\pm$0.63 & 50.65$\pm$0.90
    \\
    \ourmethod 
    & 69.29$\pm$0.66 & \textbf{20.98}$\pm$0.65 & \textbf{16.23}$\pm$0.60 & \textbf{50.07}$\pm$1.20
    \\ \midrule
    KEP-7/7       
    & 59.57$\pm$0.30 & \textbf{21.78}$\pm$0.59 & \textbf{17.17}$\pm$0.39 & 60.67$\pm$0.72
    \\
    \ourmethod  
    & \textbf{68.12}$\pm$0.26 & 22.19$\pm$0.19 & 17.70$\pm$0.17 & \textbf{52.27}$\pm$0.35
    \\ \bottomrule
\end{tabular}}
\end{minipage}
\hfill
\begin{minipage}{0.485\linewidth}
\centering
\caption{~Comparison on CoLA OOD.}\vspace{2mm}
\label{tab:dis_shift_cola}
\setlength\tabcolsep{3pt} 
\resizebox{\columnwidth}{!}{
\begin{tabular}{|l|llll|}
    \toprule
    \textbf{Method} 
    & \textbf{MCC} $\uparrow$ 
    & \textbf{ECE} $\downarrow$ 
    & \textbf{NLL} $\downarrow$ 
    & \textbf{Brier} $\downarrow$ 
    \\ \midrule
    Transformer             
    & 18.43$\pm$3.55 & 31.99$\pm$2.70 & 23.82$\pm$5.16 & 65.50$\pm$4.32 
    \\
    \ourmethod   
    & \textbf{23.06}$\pm$4.69 & \textbf{28.72}$\pm$1.87 & \textbf{17.18}$\pm$3.75 & \textbf{59.91}$\pm$2.20
    \\ \midrule
    KEP-1/5
    & 19.44$\pm$1.94 & 30.33$\pm$1.48 & 19.67$\pm$4.82 & 61.97$\pm$2.25
    \\
    \ourmethod 
    & \textbf{22.10}$\pm$5.49 & \textbf{17.50}$\pm$7.52 & \textbf{9.34}$\pm$4.50 & \textbf{49.92}$\pm$5.80
    \\ \midrule
    KEP-5/5       
    & \textbf{21.14}$\pm$3.48 & 23.27$\pm$6.61 & 14.25$\pm$10.66 & 55.12$\pm$7.78
    \\
    \ourmethod  
    & 20.20$\pm$5.46 & \textbf{17.49}$\pm$5.93 & \textbf{8.61}$\pm$1.99 & \textbf{48.55}$\pm$3.67
    \\ \bottomrule
\end{tabular}}
\end{minipage}
\end{table}
\begin{table}[t]
    \caption{Comparison of average OOD detection performance in AUROC (\%) and AUPR (\%) (with reported standard error) achieved by the tested baselines over $5$ independent runs.}
    \label{tab::ood}
    \begin{center}
    \resizebox{0.85\columnwidth}{!}
    {
    \begin{tabular}{|l|ll|ll|ll|}
        \toprule
        \multirow{2}{*}{\textbf{Method}} & \multicolumn{2}{c|}{\textbf{SVHN}} & \multicolumn{2}{c|}{\textbf{CIFAR-100}} & \multicolumn{2}{c|}{\textbf{LSUN}} \\
        \cmidrule(lr){2-3} \cmidrule(lr){4-5} \cmidrule(lr){6-7}
        & \textbf{AUROC} $\uparrow$ & \textbf{AUPR} $\uparrow$ & \textbf{AUROC} $\uparrow$ & \textbf{AUPR} $\uparrow$ & \textbf{AUROC} $\uparrow$ & \textbf{AUPR} $\uparrow$ \\
        \midrule
        MCD & 87.09$\pm$8.53 & 91.46$\pm$4.87 & 76.27$\pm$0.35 & 78.82$\pm$0.41 & 88.41$\pm$2.05 & 91.19$\pm$1.51 \\
        KFLLA & 89.47$\pm$9.07 & 92.92$\pm$5.27 & 77.27$\pm$0.42 & 79.88$\pm$0.42 & 90.77$\pm$2.93 & 92.61$\pm$2.24 \\
        SVDKL & 86.59$\pm$6.86 & 90.78$\pm$4.04 & 75.99$\pm$0.74 & 77.89$\pm$1.23 & 87.81$\pm$2.52 & 90.60$\pm$1.91 \\
        SGPA & 61.57$\pm$5.11 & 74.59$\pm$3.50 & 73.42$\pm$1.87 & 75.93$\pm$1.87 & 67.34$\pm$9.77 & 76.76$\pm$5.93 \\
        \midrule
        ViT 
        & \textbf{87.09}$\pm$8.53 & \textbf{91.46}$\pm$4.87 & 76.27$\pm$0.35 & 78.82$\pm$0.41 & \textbf{88.41}$\pm$2.05 & \textbf{91.19}$\pm$1.51 \\
        \ourmethod 
        & 83.19$\pm$10.94 & 88.84$\pm$6.37 & \textbf{78.57}$\pm$0.94 & \textbf{81.33}$\pm$0.95 & 83.97$\pm$7.19 & 88.70$\pm$4.50 \\
        \midrule
        KEP-1/7
        & 75.28$\pm$19.12 & 81.92$\pm$13.36 & 77.93$\pm$0.39 & 80.85$\pm$0.46 & 85.64$\pm$4.47 & 88.98$\pm$3.57 \\
        \ourmethod 
        & \textbf{90.73}$\pm$4.07 & \textbf{93.21}$\pm$2.78 & \textbf{79.29}$\pm$0.50 & \textbf{82.11}$\pm$0.40 & \textbf{89.38}$\pm$3.22 & \textbf{92.08}$\pm$2.37 \\
        \midrule
        KEP-2/7 & 88.25$\pm$4.67 & 91.56$\pm$3.14 & 77.71$\pm$0.55 & 80.58$\pm$0.53 & 88.35$\pm$3.62 & 91.18$\pm$2.75 \\
        \ourmethod & \textbf{\textcolor{blue}{92.14}}$\pm$5.70 & \textbf{\textcolor{blue}{94.49}}$\pm$3.59 & \textbf{\textcolor{blue}{79.43}}$\pm$0.57 & \textbf{\textcolor{blue}{82.15}}$\pm$0.57 & \textbf{\textcolor{blue}{91.39}}$\pm$2.74 & \textbf{\textcolor{blue}{93.63}}$\pm$1.86 \\
        \midrule
        KEP-7/7
        & 77.16$\pm$1.62 & 84.09$\pm$1.28 & 76.21$\pm$0.41 & 78.82$\pm$0.41 & 77.01$\pm$3.28 & 82.42$\pm$2.73 \\
        \ourmethod 
        & \textbf{79.33}$\pm$20.83 & \textbf{85.92}$\pm$13.88 & \textbf{79.11}$\pm$0.39 & \textbf{81.70}$\pm$0.38 & \textbf{86.84}$\pm$4.07 & \textbf{90.05}$\pm$3.16 \\
        \bottomrule
    \end{tabular}}
    \end{center}
\end{table}

\subsubsection{Distribution Shift Robustness}

We also assess both the uncertainty calibration and predictive performance of \ourmethod{} in scenarios with distribution shifts in both image classification and linguistic acceptability tasks.~For vision, we use the CIFAR-10-C dataset, which includes $15$ corruption types (e.g., noise, blur) at $5$ severity levels~\citep{hendrycks2019robustness}.~For language, we use the CoLA OOD dataset, which assesses novel linguistic structures \citep{warstadt2019neural}.On CIFAR-10-C (see Table~\ref{tab::dis_shift}), \ourmethod{} substantially improves the robustness of KEP-7/7 in predictive accuracy and Brier score, while remaining comparable in ECE and NLL under distribution shift.~It also remains competitive with ViT and KEP-1/7 in predictive performance while improving on calibration metrics.~On CoLA OOD (see Table~\ref{tab:dis_shift_cola}), \ourmethod{} significantly improves calibration metrics without sacrificing MCC, except when compared to KEP-5/5 which has slightly better MCC but is weaker on UC metrics.~These observations consistently demonstrate both the performance robustness and generalization capability of \ourmethod{} under scenarios with distribution shifts.

\subsubsection{Out-of-Distribution Detection}
Uncertainty-aware baselines can also be evaluated in terms of their abilities to distinguish between (i) correctly classified in-distribution samples, (ii) misclassified in-distribution samples, and (iii) out-of-distribution (OOD) samples.~To assess this capability, we report the average performance with standard error of~\ourmethod{} in a number of OOD detection scenarios (see Table~\ref{tab::ood}) using the AUROC/AUPR metrics and two standard methods: (1) Maximum Softmax Probability~\citep{hendrycks2016baseline} and Entropy Maximization~\citep{chan2021entropymaximizationmetaclassification}.~Using CIFAR-10 as the in-distribution dataset, our evaluation on SVHN, CIFAR-100, and LSUN demonstrates that the performance of most pre-trained transformer (using CIFAR-100 data) in most cases can be substantially improved by their corresponding diffusion-based reconfiguration.~Notably, the diffusion-inspired reconfiguration of KEP-2/7 produced by~\ourmethod{} outperforms all baselines as highlighted in \textcolor{blue}{blue}.



Additional results, including large-scale experiments on ViT-B-16 (86M) \qhung{and Qwen-2.5 (494M)} , deep ensembles~\citep{lakshminarayanan2017simple}, \qhung{reconfiguration of the SGPA backbone, post-hoc Temperature Scaling applied to every baseline,} computational and memory overhead, and loss component ablations are provided in Appendix~\ref{app:experiments}.

\section{Related Work}
In safety-critical decision-making applications (e.g., healthcare~\citep{Shamsi2021,Lopez2023InformativePI,Band2022BenchmarkingBD}), models must recognize when their confidence is low to defer decisions to human experts~\citep{Pietraszek2007OnTU,Tran2022PlexTR}. However, existing transformers typically ignore uncertainty due to point-estimate designs throughout their stack of neural transformations~\citep{Papamarkou2024PositionBD}. Prior investigation in Bayesian deep learning (BDL) are often restricted to moderate-sized DL architectures~\citep{Wang2016TowardsBD,mukhoti2018evaluating, kendall2017uncertainties,gustafsson2020evaluating, chien2015bayesian,ritter2021sparse,tran2019bayesian,Fortuin2021BayesianNN,Tran2020AllYN,Rudner2023ContinualLV,Qiu2023ShouldWL}, limiting scalability to large networks~\citep{Papamarkou2024PositionBD}.~To elaborate, we next discuss two main UC paradigms:\vspace{1mm} 

\noindent {\bf UC integration during training.}~These methods treat model parameters as random variables and learn their posterior distributions conditioned on data.~Bayesian neural networks approximate posteriors via MCMC or variational inference~\citep{blundell2015weight,guo2022uncertainty}, while deep ensembles~\citep{lakshminarayanan2017simple} approximate them non-parametrically through diverse initializations. Evidential methods~\citep{pmlr-v51-wilson16,sensoy2018evidential} map features to prior parameters over the likelihood for a closed-form uncertainty estimation via conjugate priors. Sampling-based methods offer higher fidelity by avoiding structural assumptions but incur prohibitive sampling cost for large models~\citep{pmlr-v119-wenzel20a}. In contrast, variational and evidential approaches are more scalable but less accurate due to biased approximations and restrictive parameterizations~\citep{pmlr-v176-wilson22a,chen2015geometric}.~Overall, these approaches approximate or sample from the parameter posterior, which remains highly intractable and often yields unreliable estimates.


\noindent {\bf UC integration post-training.}~These methods recalibrate prediction confidence by augmenting a trained model’s output without altering most its parameters, including data augmentation~\citep{wang2019aleatoric}, Monte Carlo dropout~\citep{gal2016dropout}, and input-gradient norms~\citep{ash2019deep}.~There are also learning-based approaches that adjust output probabilities to better reflect correctness, such as temperature scaling~\citep{guo2017calibration}, replacing the solution head with probabilistic alternatives (e.g., Gaussian processes~\citep{rasmussen2003gaussian}, SNGPs~\citep{liu2020simple,bradshaw2017adversarial}), or Laplace approximation that fits a local Gaussian approximation to the weight posterior around the model’s learned parameters~\citep{he2026survey}.~More recently, conformal prediction~\citep{marx2022modular} offers a black-box calibration method that uses a pre-trained model's softmax scores and test-time data to produce prediction sets with marginal coverage guarantees.

\noindent {\bf Uncertainty calibration versus uncertainty quantification.}~Beyond the uncertainty calibration methods discussed above, uncertainty quantification also plays an important role in safety-critical decision-making applications. While closely related, uncertainty calibration (UC) and uncertainty quantification (UQ) address different aspects of predictive uncertainty. UC focuses on the reliability of a model’s confidence estimates, ensuring that predicted confidence aligns with empirical correctness (e.g., predictions made with confidence $p$ are correct approximately $p$ fraction of the time). In contrast, UQ aims to model and represent uncertainty itself, characterizing predictive variability arising from sources such as data noise (aleatoric uncertainty) and model uncertainty (epistemic uncertainty). As a result, a model may exhibit rich uncertainty estimates but remain poorly calibrated, or be well-calibrated yet provide uninformative uncertainty. This distinction highlights that UC and UQ are complementary but non-interchangeable objectives, and motivates approaches that explicitly target calibration reliability in large-scale transformer models.

\label{sec:result2}



\section{Conclusion}
\label{sec:conclude}
We introduce a diffusion-inspired reconfiguration of pre-trained transformers that enables principled uncertainty propagation across the entire feature transformation stack. Our approach builds on the established connection between multi-head self-attention (MHSA) and Gaussian process (GP) prediction, reparameterizing the feature transformation stack as a sequence of neuralized Gaussian transitions. This sequence is distilled into a diffusion process with a learnable unified spatiotemporal transition model that maps between data and feature distributions, embedding expressive uncertainty-aware structure within the original transformer while preserving predictive performance. We comprehensively evaluate our method across diverse vision and language tasks, consistently demonstrating its effectiveness. These results point to a new direction for embedding probabilistic reasoning into large pre-trained models, enhancing reliability in risk-sensitive applications.

\newpage




\bibliography{main, bibs/Jana-bib,bibs/Nghia-bib-1,bibs/Nghia-bib-2,bibs/Nghia-bib-3,bibs/Nghia-bib-4}

\begin{thebibliography}{10}

\bibitem{Shamsi2021}
Shamsi A., Asgharnezhad H., and Jokandan SS.
\newblock An uncertainty-aware transfer learning-based framework for covid-19 diagnosis.
\newblock {\em {IEEE} {T}ransaction on {N}eural {N}etwork {L}earning {S}ystems}, 2021.

\bibitem{achiam2023gpt}
Josh Achiam, Steven Adler, Sandhini Agarwal, Lama Ahmad, Ilge Akkaya, Florencia~Leoni Aleman, Diogo Almeida, Janko Altenschmidt, Sam Altman, Shyamal Anadkat, et~al.
\newblock Gpt-4 technical report.
\newblock {\em arXiv preprint arXiv:2303.08774}, 2023.

\bibitem{ash2019deep}
Jordan~T Ash et~al.
\newblock Deep batch active learning by diverse, uncertain gradient lower bounds.
\newblock {\em arXiv preprint arXiv:1906.03671}, 2019.

\bibitem{baevski2020wav2vec}
Alexei Baevski, Yuhao Zhou, Abdelrahman Mohamed, and Michael Auli.
\newblock wav2vec 2.0: A framework for self-supervised learning of speech representations.
\newblock {\em Advances in neural information processing systems}, 33:12449--12460, 2020.

\bibitem{Band2022BenchmarkingBD}
Neil Band, Tim G.~J. Rudner, Qixuan Feng, Angelos Filos, Zachary Nado, Michael~W. Dusenberry, Ghassen Jerfel, Dustin Tran, and Yarin Gal.
\newblock Benchmarking bayesian deep learning on diabetic retinopathy detection tasks.
\newblock {\em ArXiv}, abs/2211.12717, 2022.

\bibitem{blundell2015weight}
Charles Blundell, Julien Cornebise, Koray Kavukcuoglu, and Daan Wierstra.
\newblock Weight uncertainty in neural network.
\newblock In {\em International conference on machine learning}, pages 1613--1622. PMLR, 2015.

\bibitem{bradshaw2017adversarial}
John Bradshaw, Alexander G de~G Matthews, and Zoubin Ghahramani.
\newblock Adversarial examples, uncertainty, and transfer testing robustness in gaussian process hybrid deep networks.
\newblock {\em arXiv preprint arXiv:1707.02476}, 2017.

\bibitem{brown2020language}
Tom Brown, Benjamin Mann, Nick Ryder, Melanie Subbiah, Jared~D Kaplan, Prafulla Dhariwal, Arvind Neelakantan, Pranav Shyam, Girish Sastry, Amanda Askell, et~al.
\newblock Language models are few-shot learners.
\newblock {\em Advances in neural information processing systems}, 33:1877--1901, 2020.

\bibitem{bui2025revisiting}
Long~Minh Bui, Tho~Tran Huu, Duy Dinh, Tan~Minh Nguyen, and Trong~Nghia Hoang.
\newblock Revisiting kernel attention with correlated gaussian process representation.
\newblock {\em arXiv preprint arXiv:2502.20525}, 2025.

\bibitem{chan2021entropymaximizationmetaclassification}
Robin Chan, Matthias Rottmann, and Hanno Gottschalk.
\newblock Entropy maximization and meta classification for out-of-distribution detection in semantic segmentation, 2021.

\bibitem{chen2015geometric}
Tian Chen, Jeffrey Streets, and Babak Shahbaba.
\newblock A geometric view of posterior approximation.
\newblock {\em arXiv preprint arXiv:1510.00861}, 2015.

\bibitem{chen2023calibrating}
Wenlong Chen and Yingzhen Li.
\newblock Calibrating transformers via sparse gaussian processes.
\newblock {\em arXiv preprint arXiv:2303.02444}, 2023.

\bibitem{chen2024primal}
Yingyi Chen, Qinghua Tao, Francesco Tonin, and Johan Suykens.
\newblock Primal-attention: Self-attention through asymmetric kernel svd in primal representation.
\newblock {\em Advances in Neural Information Processing Systems}, 36, 2024.

\bibitem{chen2024selfattention}
Yingyi Chen, Qinghua Tao, Francesco Tonin, and Johan Suykens.
\newblock Self-attention through kernel-eigen pair sparse variational gaussian processes.
\newblock In {\em Forty-first International Conference on Machine Learning}, 2024.

\bibitem{chen2024self}
Yingyi Chen, Qinghua Tao, Francesco Tonin, and Johan~AK Suykens.
\newblock Self-attention through kernel-eigen pair sparse variational gaussian processes.
\newblock {\em arXiv preprint arXiv:2402.01476}, 2024.

\bibitem{chien2015bayesian}
Jen-Tzung Chien and Yuan-Chu Ku.
\newblock Bayesian recurrent neural network for language modeling.
\newblock {\em IEEE transactions on neural networks and learning systems}, 27(2):361--374, 2015.

\bibitem{devlin2019bert}
Jacob Devlin, Ming-Wei Chang, Kenton Lee, and Kristina Toutanova.
\newblock Bert: Pre-training of deep bidirectional transformers for language understanding.
\newblock In {\em Proceedings of the 2019 conference of the North American chapter of the association for computational linguistics: human language technologies, volume 1 (long and short papers)}, pages 4171--4186, 2019.

\bibitem{dosovitskiy2020image}
Alexey Dosovitskiy, Lucas Beyer, Alexander Kolesnikov, Dirk Weissenborn, Xiaohua Zhai, Thomas Unterthiner, Mostafa Dehghani, Matthias Minderer, Georg Heigold, Sylvain Gelly, et~al.
\newblock An image is worth 16x16 words: Transformers for image recognition at scale.
\newblock {\em arXiv preprint arXiv:2010.11929}, 2020.

\bibitem{driess2023palm}
Danny Driess, Fei Xia, Mehdi~SM Sajjadi, Corey Lynch, Aakanksha Chowdhery, Ayzaan Wahid, Jonathan Tompson, Quan Vuong, Tianhe Yu, Wenlong Huang, et~al.
\newblock Palm-e: An embodied multimodal language model.
\newblock {\em arXiv preprint arXiv:2303.03378}, 2023.

\bibitem{Fortuin2021BayesianNN}
Vincent Fortuin, Adri{\`a} Garriga-Alonso, Sebastian~W. Ober, Florian Wenzel, Gunnar Ratsch, Richard~E Turner, Mark van~der Wilk, and Laurence Aitchison.
\newblock Bayesian neural network priors revisited.
\newblock In {\em International Conference on Learning Representations}, 2022.

\bibitem{gal2016dropout}
Yarin Gal and Zoubin Ghahramani.
\newblock Dropout as a bayesian approximation: Representing model uncertainty in deep learning.
\newblock In {\em international conference on machine learning}, pages 1050--1059. PMLR, 2016.

\bibitem{guo2017calibration}
Chuan Guo, Geoff Pleiss, Yu~Sun, and Kilian~Q Weinberger.
\newblock On calibration of modern neural networks.
\newblock In {\em International conference on machine learning}, pages 1321--1330. PMLR, 2017.

\bibitem{guo2022uncertainty}
Hongji Guo, Hanjing Wang, and Qiang Ji.
\newblock Uncertainty-guided probabilistic transformer for complex action recognition.
\newblock In {\em Proceedings of the IEEE/CVF Conference on Computer Vision and Pattern Recognition}, pages 1935--1944, 2022.

\bibitem{gustafsson2020evaluating}
Fredrik~K Gustafsson, Martin Danelljan, and Thomas~B Schon.
\newblock Evaluating scalable bayesian deep learning methods for robust computer vision.
\newblock In {\em Proceedings of the IEEE/CVF conference on computer vision and pattern recognition workshops}, pages 318--319, 2020.

\bibitem{he2026survey}
Wenchong He, Zhe Jiang, Tingsong Xiao, Zelin Xu, and Yukun Li.
\newblock A survey on uncertainty quantification methods for deep learning.
\newblock {\em ACM Computing Surveys}, 58(7):1--35, 2026.

\bibitem{hendrycks2019robustness}
Dan Hendrycks and Thomas Dietterich.
\newblock Benchmarking neural network robustness to common corruptions and perturbations.
\newblock {\em Proceedings of the International Conference on Learning Representations}, 2019.

\bibitem{hendrycks2016baseline}
Dan Hendrycks and Kevin Gimpel.
\newblock A baseline for detecting misclassified and out-of-distribution examples in neural networks.
\newblock In {\em International Conference on Learning Representations}, 2017.

\bibitem{hinton2015distillingknowledgeneuralnetwork}
Geoffrey Hinton, Oriol Vinyals, and Jeff Dean.
\newblock Distilling the knowledge in a neural network, 2015.

\bibitem{ho2020denoising}
Jonathan Ho, Ajay Jain, and Pieter Abbeel.
\newblock Denoising diffusion probabilistic models.
\newblock {\em Advances in Neural Information Processing Systems}, 33:6840--6851, 2020.

\bibitem{hsu2021hubert}
Wei-Ning Hsu, Benjamin Bolte, Yao-Hung~Hubert Tsai, Kushal Lakhotia, Ruslan Salakhutdinov, and Abdelrahman Mohamed.
\newblock Hubert: Self-supervised speech representation learning by masked prediction of hidden units.
\newblock {\em IEEE/ACM transactions on audio, speech, and language processing}, 29:3451--3460, 2021.

\bibitem{kendall2017uncertainties}
Alex Kendall and Yarin Gal.
\newblock What uncertainties do we need in bayesian deep learning for computer vision?
\newblock {\em Advances in {N}eural {I}nformation {P}rocessing {S}ystems}, 30, 2017.

\bibitem{kristiadi2020being}
Agustinus Kristiadi, Matthias Hein, and Philipp Hennig.
\newblock Being {B}ayesian, even just a bit, fixes overconfidence in {R}e{LU} networks.
\newblock In {\em International Conference on Machine Learning}, pages 5436--5446, 2020.

\bibitem{krizhevsky2009learning}
Alex Krizhevsky, Geoffrey Hinton, et~al.
\newblock Learning multiple layers of features from tiny images.
\newblock {\em Technical Report}, 2009.

\bibitem{lakshminarayanan2017simple}
Balaji Lakshminarayanan, Alexander Pritzel, and Charles Blundell.
\newblock Simple and scalable predictive uncertainty estimation using deep ensembles.
\newblock {\em Advances in neural information processing systems}, 30, 2017.

\bibitem{liu2023visual}
Haotian Liu, Chunyuan Li, Qingyang Wu, and Yong~Jae Lee.
\newblock Visual instruction tuning.
\newblock {\em Advances in neural information processing systems}, 36:34892--34916, 2023.

\bibitem{liu2020simple}
Jeremiah Liu, Zi~Lin, Shreyas Padhy, Dustin Tran, Tania Bedrax~Weiss, and Balaji Lakshminarayanan.
\newblock Simple and principled uncertainty estimation with deterministic deep learning via distance awareness.
\newblock {\em Advances in neural information processing systems}, 33:7498--7512, 2020.

\bibitem{liu2021swin}
Ze~Liu, Yutong Lin, Yue Cao, Han Hu, Yixuan Wei, Zheng Zhang, Stephen Lin, and Baining Guo.
\newblock Swin transformer: Hierarchical vision transformer using shifted windows.
\newblock In {\em Proceedings of the IEEE/CVF international conference on computer vision}, pages 10012--10022, 2021.

\bibitem{Lopez2023InformativePI}
L.~Julian~Lechuga Lopez, Tim G.~J. Rudner, Farah~E. Shamout, Sanyam Kapoor, Shikai Qiu, Andrew Gordon, Yiqiu Shen, Nan Wu, Aakash Kaku, Jungkyu Park, Taro Makino, Stanislaw Jastrzkeski, Jan Witowski, Duo Wang, Ben Zhang, Dustin Tran, Jeremiah Liu, Michael~W. Dusenberry, Du~Phan, Mark Collier, Jie Ren, Kehang Han, Zi~Wang, Zelda~E. Mariet, Huiyi Hu, Neil Band, Karan Singhal, Zachary Nado, Joost~R. van Amersfoort, Andreas Kirsch, Rodolphe Jenat-ton, Nithum Thain, Honglin Yuan, Kelly Buchanan, Kevin Murphy, D.~Sculley, Yarin Gal, Zoubin Ghahramani, Jasper Snoek, and Balaji Lakshmi-narayanan.
\newblock Informative priors improve the reliability of multimodal clinical data classification.
\newblock {\em ArXiv}, abs/2312.00794, 2023.

\bibitem{maas2011learning}
Andrew Maas, Raymond~E Daly, Peter~T Pham, Dan Huang, Andrew~Y Ng, and Christopher Potts.
\newblock Learning word vectors for sentiment analysis.
\newblock In {\em Annual Meeting of the Association for Computational Linguistics: Human Language Technologies}, pages 142--150, 2011.

\bibitem{marx2022modular}
Charles Marx, Shengjia Zhao, Willie Neiswanger, and Stefano Ermon.
\newblock Modular conformal calibration.
\newblock In {\em Proceedings of the 39th International Conference on Machine Learning (ICML)}, 2022.

\bibitem{moon2020confidence}
Jooyoung Moon, Jihyo Kim, Younghak Shin, and Sangheum Hwang.
\newblock Confidence-aware learning for deep neural networks.
\newblock In {\em international conference on machine learning}, pages 7034--7044. PMLR, 2020.

\bibitem{mukhoti2018evaluating}
Jishnu Mukhoti and Yarin Gal.
\newblock Evaluating bayesian deep learning methods for semantic segmentation.
\newblock {\em arXiv preprint arXiv:1811.12709}, 2018.

\bibitem{mukhoti2020calibrating}
Jishnu Mukhoti, Viveka Kulharia, Amartya Sanyal, Stuart Golodetz, Philip Torr, and Puneet Dokania.
\newblock Calibrating deep neural networks using focal loss.
\newblock {\em Advances in neural information processing systems}, 33:15288--15299, 2020.

\bibitem{Papamarkou2024PositionBD}
Theodore Papamarkou, Maria Skoularidou, Konstantina Palla, Laurence Aitchison, Julyan Arbel, David Dunson, Maurizio Filippone, Vincent Fortuin, Philipp Hennig, Jos'e~Miguel Hern'andez-Lobato, Aliaksandr Hubin, Alexander Immer, Theofanis Karaletsos, Mohammad~Emtiyaz Khan, Agustinus Kristiadi, Yingzhen Li, Stephan Mandt, Christopher Nemeth, Michael~A. Osborne, Tim G.~J. Rudner, David Rugamer, Yee~Whye Teh, Max Welling, Andrew~Gordon Wilson, and Ruqi Zhang.
\newblock Position: {B}ayesian deep learning is needed in the age of large-scale {AI}.
\newblock In {\em International Conference on Machine Learning}, 2024.

\bibitem{Peebles2022DiT}
William Peebles and Saining Xie.
\newblock Scalable diffusion models with transformers.
\newblock {\em arXiv preprint arXiv:2212.09748}, 2022.

\bibitem{Pietraszek2007OnTU}
Tadeusz Pietraszek.
\newblock On the use of {ROC} analysis for the optimization of abstaining classifiers.
\newblock {\em Machine Learning}, 68:137--169, 2007.

\bibitem{Qiu2023ShouldWL}
Shikai Qiu, Tim G.~J. Rudner, Sanyam Kapoor, and Andrew~Gordon Wilson.
\newblock Should we learn most likely functions or parameters?
\newblock In {\em Thirty-seventh Conference on Neural Information Processing Systems}, 2023.

\bibitem{radford2021learning}
Alec Radford, Jong~Wook Kim, Chris Hallacy, Aditya Ramesh, Gabriel Goh, Sandhini Agarwal, Girish Sastry, Amanda Askell, Pamela Mishkin, Jack Clark, et~al.
\newblock Learning transferable visual models from natural language supervision.
\newblock In {\em International conference on machine learning}, pages 8748--8763. PmLR, 2021.

\bibitem{radford2023robust}
Alec Radford, Jong~Wook Kim, Tao Xu, Greg Brockman, Christine McLeavey, and Ilya Sutskever.
\newblock Robust speech recognition via large-scale weak supervision.
\newblock In {\em International conference on machine learning}, pages 28492--28518. PMLR, 2023.

\bibitem{radford2019language}
Alec Radford, Jeffrey Wu, Rewon Child, David Luan, Dario Amodei, Ilya Sutskever, et~al.
\newblock Language models are unsupervised multitask learners.
\newblock {\em OpenAI blog}, 1(8):9, 2019.

\bibitem{rasmussen2003gaussian}
Carl~Edward Rasmussen and Christopher~KI Williams.
\newblock Gaussian processes for machine learning.
\newblock {\em MIT press Cambridge, MA}, 1(3), 2006.

\bibitem{ritter2021sparse}
Hippolyt Ritter, Martin Kukla, Cheng Zhang, and Yingzhen Li.
\newblock Sparse uncertainty representation in deep learning with inducing weights.
\newblock {\em Advances in Neural Information Processing Systems}, 34:6515--6528, 2021.

\bibitem{Rudner2023ContinualLV}
Tim G.~J. Rudner, Freddie Bickford~Smith, Qixuan Feng, Yee~Whye Teh, and Yarin Gal.
\newblock Continual learning via sequential function-space variational inference.
\newblock In Kamalika Chaudhuri, Stefanie Jegelka, Le~Song, Csaba Szepesvari, Gang Niu, and Sivan Sabato, editors, {\em Proceedings of the 39th International Conference on Machine Learning}, volume 162 of {\em Proceedings of Machine Learning Research}, pages 18871--18887. PMLR, 17--23 Jul 2022.

\bibitem{rudner2022tractable}
Tim~GJ Rudner, Zonghao Chen, Yee~Whye Teh, and Yarin Gal.
\newblock Tractable function-space variational inference in bayesian neural networks.
\newblock {\em Advances in Neural Information Processing Systems}, 35:22686--22698, 2022.

\bibitem{rudner2023function}
Tim~GJ Rudner, Sanyam Kapoor, Shikai Qiu, and Andrew~Gordon Wilson.
\newblock Function-space regularization in neural networks: A probabilistic perspective.
\newblock In {\em International Conference on Machine Learning}, pages 29275--29290. PMLR, 2023.

\bibitem{sensoy2018evidential}
Murat Sensoy, Lance Kaplan, and M~R Kandemir.
\newblock Evidential deep learning to quantify classification uncertainty.
\newblock {\em arXiv preprint arXiv:1806.01768}, 2018.

\bibitem{sohl2015deep}
Jascha Sohl-Dickstein, Eric Weiss, Niru Maheswaranathan, and Surya Ganguli.
\newblock Deep unsupervised learning using nonequilibrium thermodynamics.
\newblock In {\em International conference on machine learning}, pages 2256--2265. pmlr, 2015.

\bibitem{team2023gemini}
Gemini Team, Rohan Anil, Sebastian Borgeaud, Jean-Baptiste Alayrac, Jiahui Yu, Radu Soricut, Johan Schalkwyk, Andrew~M Dai, Anja Hauth, Katie Millican, et~al.
\newblock Gemini: a family of highly capable multimodal models.
\newblock {\em arXiv preprint arXiv:2312.11805}, 2023.

\bibitem{touvron2021training}
Hugo Touvron, Matthieu Cord, Matthijs Douze, Francisco Massa, Alexandre Sablayrolles, and Herv{\'e} J{\'e}gou.
\newblock Training data-efficient image transformers \& distillation through attention.
\newblock In {\em International conference on machine learning}, pages 10347--10357. PMLR, 2021.

\bibitem{touvron2023llama}
Hugo Touvron, Thibaut Lavril, Gautier Izacard, Xavier Martinet, Marie-Anne Lachaux, Timoth{\'e}e Lacroix, Baptiste Rozi{\`e}re, Naman Goyal, Eric Hambro, Faisal Azhar, et~al.
\newblock Llama: Open and efficient foundation language models.
\newblock {\em arXiv preprint arXiv:2302.13971}, 2023.

\bibitem{Tran2020AllYN}
Ba-Hien Tran, Simone Rossi, Dimitrios Milios, and Maurizio Filippone.
\newblock All you need is a good functional prior for bayesian deep learning.
\newblock {\em Journal of Machine Learning Research}, 23:74:1--74:56, 2020.

\bibitem{tran2019bayesian}
Dustin Tran, Mike Dusenberry, Mark Van Der~Wilk, and Danijar Hafner.
\newblock Bayesian layers: A module for neural network uncertainty.
\newblock {\em Advances in neural information processing systems}, 32, 2019.

\bibitem{tran2022plex}
Dustin Tran, Jeremiah Liu, Michael~W Dusenberry, Du~Phan, Mark Collier, Jie Ren, Kehang Han, Zi~Wang, Zelda Mariet, Huiyi Hu, et~al.
\newblock Plex: Towards reliability using pretrained large model extensions.
\newblock {\em arXiv preprint arXiv:2207.07411}, 2022.

\bibitem{Tran2022PlexTR}
Dustin Tran, Jeremiah~Zhe Liu, Michael~W. Dusenberry, Du~Phan, Mark Collier, Jie~Jessie Ren, Kehang Han, Z.~Wang, Zelda~E. Mariet, Huiyi Hu, Neil Band, Tim G.~J. Rudner, K.~Singhal, Zachary Nado, Joost~R. van Amersfoort, Andreas Kirsch, Rodolphe Jenatton, Nithum Thain, Honglin Yuan, E.~Kelly Buchanan, Kevin Murphy, D.~Sculley, Yarin Gal, Zoubin Ghahramani, Jasper Snoek, and Balaji Lakshminarayanan.
\newblock Plex: Towards reliability using pretrained large model extensions.
\newblock {\em ArXiv}, abs/2207.07411, 2022.

\bibitem{tsai2019transformer}
Yao-Hung~Hubert Tsai, Shaojie Bai, Makoto Yamada, Louis-Philippe Morency, and Ruslan Salakhutdinov.
\newblock Transformer dissection: a unified understanding of transformer's attention via the lens of kernel.
\newblock {\em arXiv preprint arXiv:1908.11775}, 2019.

\bibitem{vaswani2017attention}
Ashish Vaswani, Noam Shazeer, Niki Parmar, Jakob Uszkoreit, Llion Jones, Aidan~N Gomez, {\L}ukasz Kaiser, and Illia Polosukhin.
\newblock Attention is all you need.
\newblock {\em Advances in neural information processing systems}, 30, 2017.

\bibitem{wang2019aleatoric}
Guotai Wang et~al.
\newblock Aleatoric uncertainty estimation with test-time augmentation for medical image segmentation with convolutional neural networks.
\newblock {\em Neurocomputing}, 338:34--45, 2019.

\bibitem{Wang2016TowardsBD}
Hao Wang and D.~Y. Yeung.
\newblock Towards bayesian deep learning: A survey.
\newblock {\em ArXiv}, abs/1604.01662, 2016.

\bibitem{warstadt2019neural}
Alex Warstadt, Amanpreet Singh, and Samuel~R Bowman.
\newblock Neural network acceptability judgments.
\newblock {\em Transactions of the Association for Computational Linguistics}, 7:625--641, 2019.

\bibitem{pmlr-v119-wenzel20a}
Florian Wenzel, Kevin Roth, Bastiaan Veeling, Jakub Swiatkowski, Linh Tran, Stephan Mandt, Jasper Snoek, Tim Salimans, Rodolphe Jenatton, and Sebastian Nowozin.
\newblock How good is the {B}ayes posterior in deep neural networks really?
\newblock In Hal~Daumé III and Aarti Singh, editors, {\em Proceedings of the 37th International Conference on Machine Learning}, volume 119 of {\em Proceedings of Machine Learning Research}, pages 10248--10259. PMLR, 13--18 Jul 2020.

\bibitem{wilson2016stochastic}
Andrew~G Wilson, Zhiting Hu, Russ~R Salakhutdinov, and Eric~P Xing.
\newblock Stochastic variational deep kernel learning.
\newblock {\em Advances in Neural Information Processing Systems}, 29, 2016.

\bibitem{pmlr-v51-wilson16}
Andrew~Gordon Wilson, Zhiting Hu, Ruslan Salakhutdinov, and Eric~P. Xing.
\newblock Deep kernel learning.
\newblock In Arthur Gretton and Christian~C. Robert, editors, {\em Proceedings of the 19th International Conference on Artificial Intelligence and Statistics}, volume~51 of {\em Proceedings of Machine Learning Research}, pages 370--378, Cadiz, Spain, 09--11 May 2016. PMLR.

\bibitem{pmlr-v176-wilson22a}
Andrew~Gordon Wilson, Pavel Izmailov, Matthew~D Hoffman, Yarin Gal, Yingzhen Li, Melanie~F Pradier, Sharad Vikram, Andrew Foong, Sanae Lotfi, and Sebastian Farquhar.
\newblock Evaluating approximate inference in bayesian deep learning.
\newblock In Douwe Kiela, Marco Ciccone, and Barbara Caputo, editors, {\em Proceedings of the NeurIPS 2021 Competitions and Demonstrations Track}, volume 176 of {\em Proceedings of Machine Learning Research}, pages 113--124. PMLR, 2022.

\bibitem{zhu2023openmix}
Fei Zhu, Zhen Cheng, Xu-Yao Zhang, and Cheng-Lin Liu.
\newblock Openmix: Exploring outlier samples for misclassification detection.
\newblock In {\em Proceedings of the IEEE/CVF Conference on Computer Vision and Pattern Recognition}, pages 12074--12083, 2023.

\end{thebibliography}
\bibliographystyle{plain}

\newpage
\appendix
\section{Appendix / supplemental material}

\subsection{Multi-Head Self-Attention}

\textbf{Self-Attention.} Given an input of the attention layer $\mathbf{U} \in \mathbb{R}^{N \times d}$, where $N$ is the number of data points and $d$ is the embedding dimension, self-attention computes queries, keys, and values via $\mathbf{Q} = \mathbf{U}\mathbf{W}_q$, $\mathbf{K} = \mathbf{U}\mathbf{W}_k$, and $\mathbf{V} = \mathbf{U}\mathbf{W}_v$, with projection matrices $\mathbf{W}_q, \mathbf{W}_k, \mathbf{W}_v \in \mathbb{R}^{d \times d_h}$, where $d_h$ is projected dimension. The output of the self-attention is:
\begin{eqnarray}
\mathbf{F} &=& \mathrm{softmax}\left( \frac{\mathbf{Q} \mathbf{K}^\top}{\sqrt{d_h}} \right) \mathbf{V}=\mathbf{A}_{qk}\mathbf{V},
\label{eq:self-attention}
\end{eqnarray}
where attention matrix $\mathbf{A}_{qk} \in \mathbb{R}^{N \times N}$ encodes the pairwise similarity between the queries and keys.

\textbf{Multi-Head Self-Attention (MHSA).} MHSA employs $n$ parallel attention heads, each independently computing a self-attention output $\mathbf{F}^{(h)}$ with corresponding $\mathbf{W}^{(h)}_q, \mathbf{W}^{(h)}_k, \mathbf{W}^{(h)}_v \in \mathbb{R}^{d \times d_h}$as defined in Eq.~\ref{eq:self-attention}. To maintain computational efficiency, the dimensionality of each head is typically set to $d_h = d / n$. The outputs from all attention heads are concatenated and subsequently projected back to the input dimension, forming MHSA's output:
\begin{equation}
\mathbf{R} = \mathbf{O} \left[ \mathbf{F}^{(1)}, \mathbf{F}^{(2)}, \dots, \mathbf{F}^{(n)} \right] \,
\label{eq:mhsa}
\end{equation}
where $\mathbf{O} \in \mathbb{R}^{d \times (n \cdot d_h)}$ is a projection matrix. 

\subsection{Self-Attention as Gaussian Process Inference}
\label{app:sgpa-kepsvgp}
\textbf{Kernel Attention or K-Attention~\citep{tsai2019transformer}} has extended the attention mechanism by replacing cosine similarity with a general kernel function $\kappa(\cdot, \cdot):\mathbb{R}^d \times \mathbb{R}^d \rightarrow \mathbb{R}$ to compute pairwise similarities. Specifically, the attention matrix $\mathbf{A}_{qk}$ is replaced by a kernel matrix $\mathcal{K}_{qk}$, where each entry is defined as $[\mathcal{K}_{qk}]_{i,j} = \kappa(\mathbf{U}_{i,:}, \mathbf{U}_{j,:})$. Consequently, the output of the kernel attention mechanism is:
\begin{equation}\mathbf{F} = \mathcal{K}_{qk} \mathbf{V},\label{eq:kernel_attention}
\end{equation}

\textbf{Sparse Gaussian Process Attention (SGPA)~\citep{chen2023calibrating}} leverages the Sparse Variational Gaussian Process (SVGP) framework to approximate posterior variances in the attention mechanism. For each dimension $i$ of attention output, the posterior mean and covariance are given by:
\begin{align}
    \mu_i = \mathcal{K}_{qk} \mathbf{V}_{:,i} \ \text{, and} \ \
    \Sigma_i = \mathcal{K}_{qq} + \mathcal{K}_{qk} \left( \mathcal{K}_{kk}^{-1} [\mathcal{S}]_{:,:,i} \mathcal{K}_{kk}^{-1} - \mathcal{K}_{kk}^{-1} \right) \mathcal{K}_{kq}
\end{align}
where $\mathcal{S} \in \mathbb{R}^{N \times N \times d_h}$ is a set of variational covariance parameters, optimized via the SVGP evidence lower bound. The output for each dimension $i$ is then sampled using the reparameterization trick:
\begin{equation}
    \mathbf{F}_{:,i} = \mu_i + \Sigma_i^{1/2} \cdot \epsilon_i, \quad \epsilon_i \sim \mathcal{N}(0, I)
\end{equation}

By stacking the results across all $d_h$ output dimensions, the final output of the attention head is:
\begin{equation}
    \mathbf{F} = \left[ \mu_1 + \Sigma_1^{1/2} \epsilon_1 \, , \,\, \cdots \,\, , \,\, \mu_{d_h} + \Sigma_{d_h}^{1/2} \epsilon_{d_h} \right] \in \mathbb{R}^{N \times d_h}
\end{equation}

\textbf{Kernel-Eigen Pair Sparse Variational Gaussian Processes Attention (KEP-SVGP)~\citep{chen2024self}.} 
The constraint of imposing a symmetric kernel matrix $\mathcal{K}_{qk}$ in \textbf{K-Attention} and \textbf{SGPA} (requiring $\mathbf{W}_q = \mathbf{W}_k$) can restrict the model's representation capacity by eliminating the inherent asymmetry of the original attention matrix $\mathbf{A}_{qk}$. To overcome this restriction, \textbf{KEP-SVGP} employs two Gaussian Processes (GPs), leveraging the symmetry of $\mathcal{K}_{qk}\mathcal{K}_{qk}^\top$ and $\mathcal{K}_{qk}^\top\mathcal{K}_{qk}$. Each attention output dimension is then computed by combining the contributions from the two GPs. 

More specifically, building on the KSVD framework and the Primal-Attention formulation~\citep{chen2024primal}, KEP-SVGP introduces two sets of $s$-dimensional attention outputs to model the left and right eigenspaces, denoted as $F^e_{[i]} := F^e[:,i]$ and $F^r_{[i]} := F^r[:,i] \in \mathbb{R}^N$ for $i=1,\dots,s$, corresponding to the primal features $e(\mathbf{U})$ and $r(\mathbf{U})$ in \citep{chen2024primal}, respectively. To model these outputs, the following SVGP priors are defined based on the induced symmetric kernels $\mathcal{K}_{qk} \mathcal{K}_{qk}^\top$ and $\mathcal{K}_{qk}^\top \mathcal{K}_{qk}$:
\begin{equation}
\begin{pmatrix}
\mathbf{F}^e_{i} \\
\mathbf{u}^e_{i}
\end{pmatrix} \sim \mathcal{GP} \left( \mathbf{0}, \begin{pmatrix}
\mathcal{K}_{qk} \mathcal{K}_{qk}^\top & \mathbf{H}_e \Lambda^2 \\
\Lambda^2 \mathbf{H}_e^\top & \Lambda^2
\end{pmatrix} \right), \quad
\begin{pmatrix}
\mathbf{F}^r_{i} \\
\mathbf{u}^r_{i}
\end{pmatrix} \sim \mathcal{GP} \left( \mathbf{0}, \begin{pmatrix}
\mathcal{K}_{qk}^\top \mathcal{K}_{qk} & \mathbf{H}_r \Lambda^2 \\
\Lambda^2 \mathbf{H}_r^\top & \Lambda^2
\end{pmatrix} \right),
\label{eq:prior-KEP}
\end{equation}
where $\Lambda \in \mathbb{R}^{s \times s}$ is a diagonal matrix of the top-$s$ singular values of $\mathcal{K}_{qk}$, and $\mathbf{H}_e, \mathbf{H}_r \in \mathbb{R}^{N \times s}$ contain the corresponding top-$s$ left and right singular vectors, respectively. Using variational distributions $\mathbf{u}_{[i]}^e, \mathbf{u}_{[i]}^r \sim\mathcal{N}(\mathbf{m}_{\mathbf{u},{[i]}}, S_{\mathbf{uu},{[i]}})$, closed-form posteriors $q(\mathbf{F}_i^c | \mathbf{U}) = \int q(\mathbf{F}_i^c | \mathbf{u}_i) q(\mathbf{u}_i) \, \mathrm{d} \mathbf{u}_i$ ($c \in \{ e, r \}$) are derived as:
\begin{align}
q(\mathbf{F}_i^e | \mathbf{U}) &\sim \mathcal{N} \Big( \underbrace{E_U\Lambda^{-1}\mathbf{m}_{\mathbf{u},[d]}}_{\mathbf{\mu}^e:=\mathbf{m}^e_{[i]}},
    \,
    \underbrace{E_U\Lambda^{-2}S_{\mathbf{uu},[i]} E_U^\top}_{\Sigma^e:=\mathbf{L}^e_{[i]}{\mathbf{L}^e_{[i]}}^\top} \Big), \nonumber \\
q(\mathbf{F}_i^r | \mathbf{U}) &\sim \mathcal{N} \Big(
    \underbrace{R_U\Lambda^{-1}\mathbf{m}_{\mathbf{u},[d]}}_{\mathbf{\mu}^r:=\mathbf{m}^r_{[i]}},
    \,
    \underbrace{R_U\Lambda^{-2}S_{\mathbf{uu},[i]} R_U^\top}_{\Sigma^r:=\mathbf{L}^r_{[i]}{\mathbf{L}^r_{[i]}}^\top}
    \Big),
\label{eq:posterior-KEP}
\end{align}

where $E_U:=[e(\mathbf{U}_i),\ldots,e(\mathbf{U}_N)]^\top\in\mathbb{R}^{N\times s}$ and
$R_U:=[r(\mathbf{U}_i),\ldots,r(\mathbf{U}_N)]^\top \in \mathbb R^{N\times s} $
are the projection matrices w.r.t. right and left singular vectors of KSVD in \citep{chen2024primal}. The variational parameters are $\mathbf{m}_{\mathbf{u}} \in\mathbb{R}^{s\times s}$, $S_{\mathbf{uu}} \in\mathbb{R}^{s\times s \times s}$ with the components $i$ -th defined as
$\mathbf{m}_{\mathbf{u},{[i]}}:=\mathbf{m}_{\mathbf{u}}[:,i]\in\mathbb{R}^s$, and
$S_{\mathbf{uu},[i]}:=S_{\mathbf{uu}}[:,:,i] \in\mathbb{R}^{s\times s}$.

The outputs of the two SVGPs are sampled via the reparameterization trick:
\begin{equation}
\label{eq:two:outputs}
    F_{[i]}^e= \mathbf{m}^e_{[i]} +\mathbf{L}^e_{[i]}\mathbf{\epsilon}, 
    \quad
    F_{[i]}^r=\mathbf{m}^r_{[i]}+\mathbf{L}^r_{[i]}\mathbf{\epsilon}, \text{ with } \mathbf{\epsilon}\sim\mathcal{N}(0,I_N)
\end{equation} 
To fuse the outputs, two schemes are proposed: Addition ($F^{add}_{[i]}:=F_{[i]}^e + F_{[i]}^r \in\mathbb{R}^{N}$) and Concatenation ($F^{cat}_{[i]}:=[F_{[i]}^e;F_{[i]}^r ]\in\mathbb{R}^{2N}$). To align with standard Transformer architecture, the $s$-dimensional attention outputs are linearly projected to the target dimension $d_h$. The final output $\mathbf{F} \in \mathbb{R}^{N \times d_h}$ is computed as:
$\mathbf{F} = \mathbf{F}^{\text{add}}\mathbf{W}^{\text{add}}$ for the addition, $\mathbf{F}= \mathbf{W}^{\text{cat}}_1 \mathbf{F}^{\text{cat}} \mathbf{W}^{\text{cat}}_2$ for the concatenation, where the projection matrices are $\mathbf{W}^{\text{add}}\in\mathbb{R}^{s\times d_h}$,
$\mathbf{W}^{\text{cat}}_1\in\mathbb{R}^{N\times 2N}$ and $\mathbf{W}^{\text{cat}}_2\in\mathbb{R}^{s\times d_h}$.

\subsection{Denoising Diffusion Probabilistic Models}
Denoising Diffusion Probabilistic Models (DDPMs)~\citep{ho2020denoising} transform data into noise through a gradual forward diffusion process and then learn to reverse this transformation. The forward process incrementally adds Gaussian noise to the data $\mathbf{X}_0$ over $T$ steps:
\begin{equation}
p(\mathbf{X}_t|\mathbf{X}_{t-1}) = \mathcal{N}(\mathbf{X}_t; \sqrt{1-\beta_t}\mathbf{X}_{t-1}, \beta_t\mathbf{I}),
\label{eq:forward_process}
\end{equation}
where $\{\beta_t\}_{t=1}^T$ controls the noise schedule. This defines a Markov chain that progressively corrupts the data. The true reverse process $p(\mathbf{X}_{t-1}|\mathbf{X}_t)$ is generally intractable but becomes tractable when conditioned on $\mathbf{X}_0$:
\begin{equation}
p(\mathbf{X}_{t-1}|\mathbf{X}_t,\mathbf{X}_0) = \mathcal{N}\big(\mathbf{X}_{t-1}; \tilde{\mu}_t(\mathbf{X}_t,\mathbf{X}_0), \tilde{\beta}_t\mathbf{I}\big),
\label{eq:true_reverse}
\end{equation}
where closed-form expressions for $\tilde{\mu}_t$ and $\tilde{\beta}_t$ follow from Bayes’ rule under the forward process.  However, $\mathbf{X}_0$ is unknown at test time, the model instead learns a parameterized reverse process that conditions only on $(\mathbf{X}_t, t)$:
\begin{equation}
q_\theta(\mathbf{X}_{t-1}|\mathbf{X}_t) = \mathcal{N}(\mathbf{X}_{t-1}; \mu_\theta(\mathbf{X}_t,t), \Sigma_\theta(\mathbf{X}_t,t)),
\label{eq:reverse_process}
\end{equation}
Learning proceeds by minimizing the variational bound on the negative log-likelihood of the data, which encourages $q_\theta(\mathbf{X}_{t-1}|\mathbf{X}_t)$ to match the true reverse process $p(\mathbf{X}_{t-1}|\mathbf{X}_t,\mathbf{X}_0)$. This is typically implemented via noise prediction (score matching), where the network predicts the injected noise $\boldsymbol{\epsilon}$ instead of $\mu_\theta$.

\subsection{Derivation of the upper bound of $L(\theta)$}
\label{app:vlb_loss}
The negative log-likelihood $L(\theta)=\mathbb{E}_{p(\mathbf{X}_0 \mid \mathbf{X}_T)}\left[-\log q_\theta(\mathbf{X}_0 \mid \mathbf{X}_T)\right]$ is upper bounded by 
\begin{eqnarray}
    L(\theta) \leq \mathbb{H}\Big(p(\mathbf{X}_0 \mid \mathbf{X}_T)\Big) + \sum_{t=1}^T \mathbb{E}_{p(\mathbf{X}_t \mid \mathbf{X}_T)} \Big[ D_{\mathrm{KL}}\Big(p(\mathbf{X}_{t-1} | \mathbf{X}_t) \ \| \ q_\theta(\mathbf{X}_{t-1} \mid \mathbf{X}_t)\Big) \Big]
\end{eqnarray}
\textit{Proof.} From the definition of $L(\theta)$, we have
\begin{align}
    L(\theta)&=-\mathbb{E}_{p(\mathbf{X}_0 \mid \mathbf{X}_T)}\left[\log q_\theta(\mathbf{X}_0 \mid \mathbf{X}_T)\right]\label{eq:1}\\
    &=-\mathbb{E}_{p(\mathbf{X}_0 \mid \mathbf{X}_T)}\left[\log\left(\int q_\theta(\mathbf{X}_{0:T-1} \mid \mathbf{X}_T)d\mathbf{X}_{1:T-1}\right)\right]\label{eq:2}\\
    &=-\mathbb{E}_{p(\mathbf{X}_0 \mid \mathbf{X}_T)}\left[\log\left(\int p(\mathbf{X}_{1:T-1} \mid \mathbf{X}_0,\mathbf{X}_T)\dfrac{q_\theta(\mathbf{X}_{0:T-1}\mid\mathbf{X}_T)}{p(\mathbf{X}_{1:T-1}\mid\mathbf{X}_0,\mathbf{X}_T)}d\mathbf{X}_{1:T-1}\right)\right]\label{eq:3}\\
    &=-\mathbb{E}_{p(\mathbf{X}_0\mid\mathbf{X}_T)}\left[\log\left(\mathbb{E}_{p(\mathbf{X}_{1:T-1}\mid\mathbf{X}_0,\mathbf{X}_T)}\left[\dfrac{q_\theta(\mathbf{X}_{0:T-1}\mid\mathbf{X}_T)}{p(\mathbf{X}_{1:T-1}\mid\mathbf{X}_0,\mathbf{X}_T)}d\mathbf{X}_{1:T-1}\right]\right)\right]\label{eq:4}\\
    &\leq-\mathbb{E}_{p(\mathbf{X}_{0:T-1}\mid\mathbf{X}_T)}\log\left(\dfrac{q_\theta(\mathbf{X}_{0:T-1}\mid\mathbf{X}_T)}{p(\mathbf{X}_{1:T-1}\mid\mathbf{X}_0,\mathbf{X}_T)}\right)\quad(\text{Jensen's inequality})\label{eq:5}\\
    &=\mathbb{E}_{p(\mathbf{X}_{0:T-1}\mid\mathbf{X}_T)}\log\left(\dfrac{p(\mathbf{X}_{1:T-1}\mid\mathbf{X}_0,\mathbf{X}_T)}{q_\theta(\mathbf{X}_{0:T-1}\mid\mathbf{X}_T)}\right)=L_{VLB}\label{eq:6}
\end{align}
Now, we derive the $L_{VLB}$ as follows:
\begin{align}
    L_{VLB}&=\mathbb{E}_{p(\mathbf{X}_{0:T-1}\mid\mathbf{X}_T)}\left[\log\left(\dfrac{p(\mathbf{X}_{1:T-1}\mid\mathbf{X}_0,\mathbf{X}_T)}{q_\theta(\mathbf{X}_{0:T-1}\mid\mathbf{X}_T)}\right)\right]\label{eq:7}\\
    &=\mathbb{E}_{p(\mathbf{X}_{0:T-1}\mid\mathbf{X}_T)}\left[\log\left(\dfrac{p(\mathbf{X}_{0:T-1}\mid\mathbf{X}_T)}{q_\theta(\mathbf{X}_{0:T-1}\mid\mathbf{X}_T)p(\mathbf{X}_0\mid\mathbf{X}_T)}\right)\right]\label{eq:8}\\
    &=\mathbb{E}_{p(\mathbf{X}_{0:T-1}\mid\mathbf{X}_T)}\left[-\log p(\mathbf{X}_0\mid\mathbf{X}_T)+\log\left(\dfrac{p(\mathbf{X}_{0:T-1}\mid\mathbf{X}_T)}{q_\theta(\mathbf{X}_{0:T-1}\mid\mathbf{X}_T)}\right)\right]\label{eq:9}\\
    &=\mathbb{E}_{p(\mathbf{X}_{0:T-1}\mid\mathbf{X}_T)}\left[-\log p(\mathbf{X}_0\mid\mathbf{X}_T)+\log\left(\prod_{t=1}^T\dfrac{p(\mathbf{X}_{t-1}\mid\mathbf{X}_t)}{q_\theta(\mathbf{X}_{t-1}\mid\mathbf{X}_t)}\right)\right]\label{eq:10}\\
    &=\mathbb{E}_{p(\mathbf{X}_{0:T-1}\mid\mathbf{X}_T)}\left[-\log p(\mathbf{X}_0\mid\mathbf{X}_T)+\sum_{t=1}^T\log\left(\dfrac{p(\mathbf{X}_{t-1}\mid\mathbf{X}_t)}{q_\theta(\mathbf{X}_{t-1}\mid\mathbf{X}_t)}\right)\right]\label{eq:11}\\
    &=\mathbb{H}\Big(p(\mathbf{X}_0 \mid \mathbf{X}_T)\Big) + \sum_{t=1}^T \mathbb{E}_{p(\mathbf{X}_{0:T-1}\mid\mathbf{X}_T)} \left[ \log\left(\dfrac{p(\mathbf{X}_{t-1}\mid\mathbf{X}_t)}{q_\theta(\mathbf{X}_{t-1}\mid\mathbf{X}_t)}\right) \right]
\end{align}
Additionally, we have: 
\begin{align}
    &\qquad\mathbb{E}_{p(\mathbf{X}_{0:T-1}\mid\mathbf{X}_T)}\left[\log\left(\dfrac{p(\mathbf{X}_{t-1}\mid\mathbf{X}_t)}{q_\theta(\mathbf{X}_{t-1}\mid\mathbf{X}_t)}\right)\right]\label{eq:12}\\
    &=\int p(\mathbf{X}_{0:T-1}\mid\mathbf{X}_T)\log\left(\dfrac{p(\mathbf{X}_{t-1}\mid\mathbf{X}_t)}{q_\theta(\mathbf{X}_{t-1}\mid\mathbf{X}_t)}\right)d\mathbf{X}_{0:T-1}\label{eq:13}\\
    &=\int p(\mathbf{X}_{t-1},\mathbf{X}_t\mid\mathbf{X}_T)\log\left(\dfrac{p(\mathbf{X}_{t-1}\mid\mathbf{X}_t)}{q_\theta(\mathbf{X}_{t-1}\mid\mathbf{X}_t)}\right)d\mathbf{X}_{t-1}d\mathbf{X}_t\label{eq:14}\\
    &=\int p(\mathbf{X}_t\mid\mathbf{X}_T)p(\mathbf{X}_{t-1}\mid\mathbf{X}_t)\log\left(\dfrac{p(\mathbf{X}_{t-1}\mid\mathbf{X}_t)}{q_\theta(\mathbf{X}_{t-1}\mid\mathbf{X}_t)}\right)d\mathbf{X}_{t-1}d\mathbf{X}_t\label{eq:15}\\
    &=\mathbb{E}_{p(\mathbf{X}_t\mid\mathbf{X}_T)}\Big[D_{KL}\Big(p(\mathbf{X}_{t-1}\mid\mathbf{X}_t) \ \| \ q_\theta(\mathbf{X}_{t-1}\mid\mathbf{X}_t)\Big)\Big]\label{eq:16}
\end{align}


\subsection{Practical implementation of the proposed algorithm in Eq.\ref{eq:loss-l1+l2}}
\label{app:loss-derivation}
The uncertainty-aware transition parameterization can be obtained via the objective in Eq.\ref{eq:loss-l1+l2}:
\begin{eqnarray}
\theta &=& \argmin_{\theta}\ \ \Big\{ L_1(\theta) + L_2(\theta) \Big\} \
\end{eqnarray}
which combines the matching loss 
\begin{eqnarray}
\label{eq:app-l1}
L_1(\theta) &=&  \underset{(\boldsymbol{X}_t, t)\sim p(\mathbf{X}_t \mid \mathbf{X}_T)}{\mathbb{E}}\Big[\mathrm{D}_{\mathrm{KL}}\Big(p(\boldsymbol{X}_{t-1}\mid \boldsymbol{X}_t) \ \| \ q_\theta(\boldsymbol{X}_{t-1}\mid \boldsymbol{X}_t)\Big)\Big] \ ,
\end{eqnarray}
with the performance-aware loss
\begin{eqnarray}
L_2(\theta) &=& \underset{(\boldsymbol{X}, \boldsymbol{y}) \sim \boldsymbol{D} \ \ }{\mathbb{E}}\underset{\boldsymbol{X}_0 \sim q_\theta(\boldsymbol{X}_0 \mid \boldsymbol{X}_T)}{\mathbb{E}}\Big[\mathrm{loss}\Big(\boldsymbol{X}_0,\boldsymbol{y}\Big)\Big]  \ ,
\end{eqnarray}
where $\boldsymbol{X}$ is sampled from the training dataset $\boldsymbol{D}$ and is embedded with $\boldsymbol{X}_T = \mathrm{embed}(\boldsymbol{X})$. $\boldsymbol{X}_0$ is then sampled via iteratively simulating the current estimate of the probability path $q_\theta(\boldsymbol{X}_{t-1}\mid \boldsymbol{X}_t)$.
In addition, the KL divergence in Eq.~\ref{eq:app-l1} can be reduced to the following loss,
\begin{eqnarray}
D_{\mathrm{KL}}\Big(p \ \| \ q_\theta\Big) &\propto& \frac{1}{2}\left[\mathrm{tr}\left[\sigma_\theta(\boldsymbol{X}_t, t)^{-1}\sigma_t(\boldsymbol{X}_t)\right] + \log\frac{|\sigma_\theta(\boldsymbol{X}_t, t)|}{|\sigma_t(\boldsymbol{X}_t)|}\right]    \\
&+& \frac{1}{2}\Big(m_\theta(\boldsymbol{X}_t, t) -m_t(\boldsymbol{X}_t)\Big)^\top\sigma_\theta(\boldsymbol{X}_t)^{-1}\Big(m_\theta(\boldsymbol{X}_t, t) -m_t(\boldsymbol{X}_t)\Big) \ .
\end{eqnarray}
However, the KL computation still presents a challenge due to the high dimensionality of the covariance matrix $\sigma_t(\boldsymbol{X}_t)$, which complicates the evaluation of the trace and log-determinant terms. To mitigate this, we follow the approach of KEP by approximating $\sigma_t(\boldsymbol{X}_t)$ using a Cholesky-like factor $\mathbf{L}_t$ such that $\sigma_t(\boldsymbol{X}_t) = \mathbf{L}_t \mathbf{L}_t^\top$. For the learned covariance $\sigma_\theta(\boldsymbol{X}_t, t)$, we employ a diagonal parameterization, whose Cholesky-like factor reduces to the element-wise square root of its diagonal entries. As demonstrated in Appendix~\ref{subsubsec:varying-covarience}, this choice is substantially more efficient than a full covariance parameterization without sacrificing performance. Matching these Cholesky-like factors ensures that the corresponding covariance matrices are aligned, effectively nullifying the trace and log-determinant terms in the KL divergence and enabling efficient optimization. Employing Cholesky-like factor and incorporating weighting terms yields the final objective:
\begin{eqnarray}
\theta &=& \argmin_{\theta}\ \ \Big\{ \lambda_{\mathrm{mean}} L_{\mathrm{mean}}(\theta) + \lambda_{\mathrm{Cholesky}} L_{\mathrm{Cholesky}}(\theta) + \lambda_{\mathrm{NLL}}L_2(\theta) \Big\} \ ,
\end{eqnarray}

where $\lambda_{\mathrm{mean}}$, $\lambda_{\mathrm{Cholesky}}$, and $\lambda_{\mathrm{NLL}}$ are weighting coefficients for the mean matching term, the Cholesky-like factor alignment, and the performance-aware loss, respectively. The individual loss components are defined as follows:
\begin{align}
L_{\mathrm{mean}}(\theta) &= \frac{1}{T} \sum_{t=1}^T \mathbb{E}_{p(\mathbf{X}_t | \mathbf{X}_T)} \left[ \left\| m_\theta(\mathbf{X}_t, t) - m_t(\mathbf{X}_t) \right\|_2^2 \right], \label{eq:loss-mean} \\
L_{\mathrm{Cholesky}}(\theta) &= \frac{1}{T} \sum_{t=1}^T \mathbb{E}_{p(\mathbf{X}_t | \mathbf{X}_T)} \left[ \left\| \left( \sigma_\theta(\mathbf{X}_t, t)^{1/2} - \mathrm{Chol}(\sigma_t(\mathbf{X}_t)) \right) \right\|_2^2 \right], \label{eq:loss-cholesky} \\
L_2(\theta) &= \underset{(\boldsymbol{X}, \boldsymbol{y}) \sim \boldsymbol{D} \ \ }{\mathbb{E}}\underset{\boldsymbol{X}_0 \sim q_\theta(\boldsymbol{X}_0 \mid \boldsymbol{X}_T)}{\mathbb{E}}\Big[\mathrm{loss}\Big(\boldsymbol{X}_0,\boldsymbol{y}\Big)\Big] , \label{eq:loss-nll}
\end{align}
where $\mathrm{Chol}(\cdot)$ denotes the Cholesky-like factorization.~\qhung{We provide the pseudocode for one training iteration of~\ourmethod{} in Algorithm~\ref{alg:director-training}.}

\begin{algorithm}[ht]
    \caption{\qhung{One training iteration of~\ourmethod{}.~The pre-trained backbone $f$ is frozen.}}
    \label{alg:director-training}
    \footnotesize
    \textbf{Input}: training pair $(\boldsymbol{X}, \boldsymbol{y})$; pre-trained backbone $f$ with $T$ reconfigured blocks; transition model $q_\theta$ with parameters $\theta$; loss weights $\lambda_{\mathrm{mean}}, \lambda_{\mathrm{Cholesky}}, \lambda_{\mathrm{NLL}}$; learning rate $\eta$.\\
    \textbf{Output}: updated transition-model parameters $\theta$.
    \begin{algorithmic}[1]
        \STATE Embed input: $\boldsymbol{X}_T \leftarrow \mathrm{embed}(\boldsymbol{X})$. \COMMENT{shared with the backbone}
        \STATE With $f$ frozen, run a forward pass to obtain the trajectory $\{\boldsymbol{X}_t\}_{t=0}^{T}$ and, at each block, the ground-truth Gaussian transition statistics $\{(m_t(\boldsymbol{X}_t),\, \sigma_t(\boldsymbol{X}_t))\}_{t=1}^{T}$ (Eq.~\ref{eq:reconfig}).
        \FOR{$t = 1,\dots,T$}
            \STATE Predict $(m_\theta(\boldsymbol{X}_t, t),\, \sigma_\theta(\boldsymbol{X}_t, t)) \leftarrow q_\theta(\boldsymbol{X}_t, t)$. \COMMENT{single-block DiT}
        \ENDFOR
        \STATE Roll out $\boldsymbol{X}_0 \sim q_\theta(\boldsymbol{X}_0 \mid \boldsymbol{X}_T)$ and pass through the classification head to obtain logits $\hat{\boldsymbol{y}}$.
        \STATE $L_{\mathrm{mean}} \leftarrow \frac{1}{T}\sum_{t=1}^{T}\big\|\,m_\theta(\boldsymbol{X}_t, t) - m_t(\boldsymbol{X}_t)\,\big\|_2^2$ \COMMENT{Eq.~\ref{eq:loss-mean}}
        \STATE $L_{\mathrm{Cholesky}} \leftarrow \frac{1}{T}\sum_{t=1}^{T}\big\|\,\sigma_\theta(\boldsymbol{X}_t, t)^{1/2} - \mathrm{Chol}(\sigma_t(\boldsymbol{X}_t))\,\big\|_2^2$ \COMMENT{Eq.~\ref{eq:loss-cholesky}}
        \STATE $L_{\mathrm{NLL}} \leftarrow \mathrm{CE}(\hat{\boldsymbol{y}}, \boldsymbol{y})$ \COMMENT{Eq.~\ref{eq:loss-nll}}
        \STATE $\mathcal{L}(\theta) \leftarrow \lambda_{\mathrm{mean}} L_{\mathrm{mean}} + \lambda_{\mathrm{Cholesky}} L_{\mathrm{Cholesky}} + \lambda_{\mathrm{NLL}} L_{\mathrm{NLL}}$ \COMMENT{Eq.~\ref{eq:loss-l1+l2}}
        \STATE $\theta \leftarrow \theta - \eta\, \nabla_\theta \mathcal{L}(\theta)$
        \STATE \textbf{return} $\theta$
    \end{algorithmic}
\end{algorithm}

\subsection{Additional experiments}
\label{app:experiments}

\subsubsection{Large-scale experiment}

\begin{table*}[h]
    \caption{Comparison of average performance achieved a pre-trained ViT-B-16 model (fine-tuned on CIFAR-10) and its reconfigured variant produced by~\ourmethod.}
    \label{tab:cifar10-finetune}
    \begin{center}
    \begin{tabular}{|l|ccccc|}
        \toprule
        \textbf{Method} & \textbf{\#params} & \textbf{ACC} $\uparrow$ & \textbf{ECE} $\downarrow$ & \textbf{NLL} $\downarrow$ & \textbf{Brier} $\downarrow$ 
        \\ \midrule
        ViT-B-16       
        & 86M & \textbf{97.880} & 18.021 & 2.564 & 6.843 \\
        \ourmethod         
        & 50M & 97.170 & \textbf{1.611} & \textbf{1.122} & \textbf{4.762} \\
        \bottomrule
    \end{tabular}
    \end{center}
\end{table*}

\begin{table*}[h]
    \caption{Comparison of average performance achieved a pre-trained Qwen-2.5 model (fine-tuned on SST-2) and its reconfigured variant produced by~\ourmethod.}
    \label{tab:sst2-finetune}
    \begin{center}
    \begin{tabular}{|l|ccccc|}
        \toprule
        \textbf{Method} & \textbf{\#params} & \textbf{ACC} $\uparrow$ & \textbf{ECE} $\downarrow$ & \textbf{NLL} $\downarrow$ & \textbf{Brier} $\downarrow$ 
        \\ \midrule
        Qwen-2.5
        & 0.494B & 93.005 & 5.864 & 3.300 & 12.184 \\
        \ourmethod
        & 0.479B & \textbf{93.119} & \textbf{5.482} & \textbf{2.953} & \textbf{12.089} \\
        \bottomrule
    \end{tabular}
    \end{center}
\end{table*}
To stress test~\ourmethod, we further evaluate its scalability on a larger-scale experiment which requires reconfiguring a pre-trained ViT-B-16 model (pre-trained on the large-scale ImageNet data and fine-tuned on CIFAR-10 data).~The reported results in Table~\ref{tab:cifar10-finetune} show that~\ourmethod{} 's reconfigured ViT-B-16 achieves substantially better uncertainty calibration with an ECE of 1.611 compared to 18.021 of the original ViT-B-16.~\ourmethod{} also maintains competitive predictive performance (97.17\% vs. 97.88\%) while requiring significantly fewer parameters (50M vs. 86M).~These findings highlight the ability of \ourmethod{} to deliver reliable uncertainty calibration even in large and complex transformer architectures, pre-trained on sophisticated and large-scale data.

\qhung{We further evaluate the scalability of~\ourmethod{} to a large-scale pre-trained language model by reconfiguring the last three transformer blocks of a Qwen-2.5 model fine-tuned on SST-2; due to the compute cost of a full-stack reconfiguration on a billion-parameter language model, the remaining blocks retain their original deterministic computation.~The reported results in Table~\ref{tab:sst2-finetune} show that~\ourmethod{}'s reconfigured Qwen-2.5 achieves better uncertainty calibration with an ECE of 5.482 compared to 5.864 of the original Qwen-2.5, together with consistent improvements across the other calibration metrics, reducing NLL from 3.300 to 2.953 and Brier from 12.184 to 12.089.~\ourmethod{} also slightly improves predictive accuracy (93.119\% vs.\ 93.005\%) while requiring fewer parameters (0.479B vs.\ 0.494B).~These findings highlight the ability of~\ourmethod{} to deliver reliable uncertainty calibration even in large pre-trained language models.}

\subsubsection{Deep Ensembles Results}
\begin{table*}[t]
    \caption{Performance comparison of Deep Ensembles for in-distribution classification on CIFAR-10 and IMDB. KEP-$k$/$n$ denotes a pre-trained transformer using GP-reparameterized architecture (KEP~\citep{chen2024self}) for the last $k$ attention blocks and standard MHSA for the remaining blocks. \textbf{Bold} indicates the better performance in each pairwise comparison between a baseline ensemble (ViT, Transformer, KEP) and diffusion-based reconfigured KEP~\citep{chen2024self} produced by~\ourmethod{} ensemble.}
    \label{tab::deep_ensembles_appendix}
    \begin{center}
    \setlength\tabcolsep{3pt}
    \resizebox{0.85\textwidth}{!}{
    \begin{tabular}{c|l|lllllll}
        \toprule
        \textbf{Dataset} & \textbf{Method} & \textbf{ACC} $\uparrow$ & \textbf{AURC} $\downarrow$ & \textbf{AUROC} $\uparrow$ & \textbf{FPR95} $\downarrow$ & \textbf{ECE} $\downarrow$ & \textbf{NLL} $\downarrow$ & \textbf{Brier} $\downarrow$ 
        \\ \midrule
        \multirow{8}{*}{\rotatebox{90}{\parbox{2cm}{\centering CIFAR-10}}} 
        & ViT       
        & 87.21 & 2.52 & 88.84 & 58.41 & 2.70 & 5.14 & 19.01 \\
        & \ourmethod         
        & \textbf{88.82} & \textbf{1.91} & \textbf{90.09} & \textbf{52.77} & \textbf{1.72} & \textbf{3.69} & \textbf{16.38} \\
        \cmidrule(lr){2-9}
        & KEP-1/7       
        & 87.48 & 2.33 & 89.43 & 56.39 & 2.30 & 4.46 & 18.29 \\
        & \ourmethod              
        & \textbf{89.32} & \textbf{1.78} & \textbf{90.12} & \textbf{54.31} & \textbf{1.67} & \textbf{3.50} & \textbf{15.76} \\
        \cmidrule(lr){2-9}
        & KEP-2/7       
        & 87.09 & 2.41 & 89.48 & 54.07 & 2.71 & 4.48 & 18.69 \\
        & \ourmethod              
        & \textbf{89.20} & \textbf{1.79} & \textbf{90.33} & \textbf{51.85} & \textbf{1.86} & \textbf{3.52} & \textbf{15.73} \\
        \cmidrule(lr){2-9} 
        & KEP-7/7       
        & 83.97 & 3.77 & 86.90 & 62.38 & 3.58 & 5.04 & 23.20 \\
        & \ourmethod              
        & \textbf{89.73} & \textbf{1.70} & \textbf{90.17} & \textbf{54.72} & \textbf{1.62} & \textbf{3.49} & \textbf{15.27} \\
        \midrule
        \multirow{8}{*}{\rotatebox{90}{\parbox{2cm}{\centering IMDB}}} 
        & Transformer     
        & 87.31 & \textbf{3.55} & \textbf{82.92} & 71.45 & 2.36 & 3.04 & 18.44 \\
        & \ourmethod              
        & \textbf{87.45} & 3.56 & 82.83 & \textbf{70.68} & \textbf{1.97} & \textbf{3.00} & \textbf{18.25} \\
        \cmidrule(lr){2-9}
        & KEP-1/5    
        & 87.44 & 3.63 & 82.34 & 72.21 & 1.56 & 3.01 & 18.39 \\
        & \ourmethod              
        & \textbf{87.85} & \textbf{3.44} & \textbf{83.20} & \textbf{71.10} & \textbf{1.25} & \textbf{2.89} & \textbf{17.62} \\
        \cmidrule(lr){2-9}
        & KEP-2/5       
        & 87.77 & 3.51 & 82.44 & 73.14 & 2.85 & 3.04 & 18.18 \\
        & \ourmethod              
        & \textbf{87.88} & \textbf{3.43} & \textbf{82.96} & \textbf{72.29} & \textbf{1.24} & \textbf{2.91} & \textbf{17.70} \\
        \cmidrule(lr){2-9} 
        & KEP-5/5     
        & 86.49 & 4.16 & 81.54 & 74.07 & 2.63 & 3.28 & 19.72 \\
        & \ourmethod              
        & \textbf{87.66} & \textbf{3.43} & \textbf{82.95} & \textbf{72.13} & \textbf{2.07} & \textbf{2.95} & \textbf{18.05} \\
        \bottomrule
    \end{tabular}}
    \end{center}
\end{table*}
We further evaluate the performance of Deep Ensembles \citep{lakshminarayanan2017simple}, a simple yet effective method for uncertainty estimation, when combined with pretrained models and \ourmethod, as reported in Table~\ref{tab::deep_ensembles_appendix}. The results show that \ourmethod{} consistently outperforms pretrained models across both CIFAR-10 and IMDB. On CIFAR-10, \ourmethod{} achieves notable improvements in predictive accuracy (up to +5.8\% over KEP-7/7), while also delivering stronger uncertainty calibration (lower ECE and NLL) and robustness (AURC, AUROC, Brier score). On IMDB, although the baseline Transformer ensemble already achieves strong results, \ourmethod{} further improves calibration (ECE, NLL, Brier) and robustness metrics, with the exception of AURC and AUROC where the Transformer baseline performs slightly better. Overall, these findings demonstrate that \ourmethod{} works effectively to ensemble settings, providing consistent benefits across architectures, datasets, and evaluation metrics.

\qhung{
\subsubsection{Reconfiguring the SGPA Backbone}
\label{subsubsec:reconfig-sgpa}

As discussed in Section~\ref{sec:result1}, SGPA exhibits a pronounced trade-off between calibration and predictive performance, achieving low ECE on CIFAR-10 (1.92) at the cost of a substantially lower accuracy (75.59\%) compared to its ViT and KEP counterparts.~A natural question is whether~\ourmethod{} can mitigate this trade-off by reconfiguring an SGPA backbone in the same way it reconfigures KEP-SVGP and ViT.~To this end, we apply our diffusion-based reconfiguration on top of a pre-trained SGPA backbone~\citep{chen2023calibrating}, following the same two-stage protocol used throughout the paper: we first train an SGPA backbone on each dataset and then train a single-block DiT-based transition model on top of the frozen SGPA features, using the same loss-weighting search procedure described in Appendix~\ref{subsec:varying-loss-weights}.~The resulting performance, alongside the SGPA stage-1 baseline reported in Table~\ref{tab::in_dist_baselines_diff}, is summarized in Table~\ref{tab:reconfig-sgpa}.

\begin{table}[t]
    \caption{Reconfiguring an SGPA backbone with~\ourmethod{} across CIFAR-10, IMDB, and CoLA.~SGPA rows reproduce the values reported in Table~\ref{tab::in_dist_baselines_diff}.~\textbf{Bold} indicates the better value within each pairwise comparison.}
    \label{tab:reconfig-sgpa}
    \begin{center}
    \resizebox{\textwidth}{!}{
    \begin{tabular}{|c|l|lllllll|}
        \toprule
        \textbf{Dataset} & \textbf{Method} & \textbf{ACC/MCC} $\uparrow$ & \textbf{AURC} $\downarrow$ & \textbf{AUROC} $\uparrow$ & \textbf{FPR95} $\downarrow$ & \textbf{ECE} $\downarrow$ & \textbf{NLL} $\downarrow$ & \textbf{Brier} $\downarrow$
        \\ \midrule
        \multirow{2}{*}{CIFAR-10}
        & SGPA
        & 75.59$\pm$3.63 & 8.41$\pm$2.37 & 82.65$\pm$1.71 & 71.78$\pm$2.73 & \textbf{1.92}$\pm$0.55 & 7.11$\pm$0.95 & 33.98$\pm$4.57 \\
        & \ourmethod
        & \textbf{82.65}$\pm$1.16 & \textbf{4.47}$\pm$0.56 & \textbf{86.00}$\pm$0.65 & \textbf{66.39}$\pm$2.34 & 4.53$\pm$0.58 & \textbf{5.30}$\pm$0.29 & \textbf{25.10}$\pm$1.48 \\
        \midrule
        \multirow{2}{*}{IMDB}
        & SGPA
        & 85.39$\pm$0.36 & 4.96$\pm$0.49 & 80.04$\pm$1.14 & 76.44$\pm$0.96 & 6.04$\pm$1.71 & 3.96$\pm$0.50 & 22.19$\pm$1.06 \\
        & \ourmethod
        & \textbf{86.16}$\pm$0.62 & \textbf{4.44}$\pm$0.44 & \textbf{81.04}$\pm$0.76 & \textbf{74.67}$\pm$1.43 & \textbf{3.12}$\pm$1.43 & \textbf{3.36}$\pm$0.16 & \textbf{20.30}$\pm$0.84 \\
        \midrule
        \multirow{2}{*}{CoLA}
        & SGPA
        & 31.53$\pm$2.05 & 20.44$\pm$2.60 & 64.34$\pm$1.95 & 90.79$\pm$0.87 & 26.22$\pm$1.51 & 28.65$\pm$7.23 & 54.08$\pm$2.44 \\
        & \ourmethod
        & \textbf{32.84}$\pm$2.66 & \textbf{19.04}$\pm$1.26 & \textbf{64.67}$\pm$1.18 & \textbf{89.38}$\pm$1.46 & \textbf{17.00}$\pm$7.62 & \textbf{13.43}$\pm$6.22 & \textbf{45.72}$\pm$5.09 \\
        \bottomrule
    \end{tabular}}
    \end{center}
\end{table}

As shown in Table~\ref{tab:reconfig-sgpa},~\ourmethod{} achieves higher predictive performance than the SGPA backbone across all three datasets, with a particularly large gain on CIFAR-10 (82.65\% versus 75.59\% in ACC) and consistent gains on IMDB (86.16\% versus 85.39\% in ACC) and CoLA (32.84 versus 31.53 in MCC).~On CIFAR-10,~\ourmethod{} improves two of the three calibration metrics (NLL from 7.11 to 5.30, Brier from 33.98 to 25.10) and all three failure-prediction metrics (AURC from 8.41 to 4.47, AUROC from 82.65 to 86.00, FPR95 from 71.78 to 66.39), with ECE being the only column where SGPA retains a smaller value (1.92 versus 4.53).~Although SGPA achieves the lowest CIFAR-10 ECE, this result reflects an unfavorable accuracy/calibration trade-off rather than genuinely well-calibrated predictions: ECE alone does not measure predictive utility, and on a balanced 10-class dataset such as CIFAR-10 a random guesser that always outputs the uniform distribution is perfectly calibrated in expectation, with confidence $0.1$ matching accuracy $0.1$.~Reading ECE jointly with NLL and Brier, which jointly penalize under-confident predictions and predictive errors, shows that~\ourmethod{} delivers a substantially better calibration profile overall (NLL $5.30$ versus $7.11$ and Brier $25.10$ versus $33.98$) while raising ACC from $75.59\%$ to $82.65\%$.~On IMDB and CoLA,~\ourmethod{} additionally reduces ECE, NLL, and Brier simultaneously, mirroring the trend observed for the KEP backbone in Table~\ref{tab::in_dist_kep_diff}.~These results demonstrate that the proposed reconfiguration extends naturally to the SGPA backbone, closing the predictive-performance gap that SGPA exhibits in Table~\ref{tab::in_dist_baselines_diff} while retaining the calibration trade-off pattern observed for ViT and KEP in Section~\ref{sec:experiments}.
}

\qhung{
\subsubsection{Effect of Post-hoc Temperature Scaling on Every Baseline}
\label{subsubsec:ts-on-all}

To examine whether the calibration gains reported by~\ourmethod{} can
be explained by a single temperature parameter, we apply post-hoc
TS to \emph{every} baseline considered in the in-distribution setting
on CIFAR-10, including the GP-aware variants (KEP-$k$/$n$ and SGPA)
that are typically left untouched in the literature.~For each
baseline, a single temperature scalar is fitted by minimizing the
validation NLL via L-BFGS and is then applied to the test-time
logits.~The resulting in-distribution metrics, averaged over five
random seeds and computed on the same checkpoints used to produce
Table~\ref{tab::in_dist_kep_diff}, are summarized in
Table~\ref{tab:ts-on-all-cifar10}.

\begin{table*}[t]
    \caption{In-distribution calibration on CIFAR-10 after applying post-hoc Temperature Scaling (TS) to every baseline.~\textbf{Bold} marks the better value within each pairwise comparison; \textcolor{blue}{\textbf{blue}} marks the best value across the whole table.}
    \label{tab:ts-on-all-cifar10}
    \begin{center}
    \resizebox{\textwidth}{!}{
    \begin{tabular}{|l|lllllll|}
        \toprule
        \textbf{Method} & \textbf{ACC} $\uparrow$ & \textbf{AURC} $\downarrow$ & \textbf{AUROC} $\uparrow$ & \textbf{FPR95} $\downarrow$ & \textbf{ECE} $\downarrow$ & \textbf{NLL} $\downarrow$ & \textbf{Brier} $\downarrow$
        \\ \midrule
        ViT~$+$~TS
        & 83.84$\pm$0.09 & 3.88$\pm$0.09 & 86.82$\pm$0.33 & 65.99$\pm$1.73 & 9.20$\pm$0.32 & 6.57$\pm$0.14 & 25.49$\pm$0.12 \\
        \ourmethod~$+$~TS
        & \textbf{85.67}$\pm$0.88 & \textbf{3.04}$\pm$0.40 & \textbf{88.27}$\pm$0.83 & \textbf{60.65}$\pm$1.23 & \textbf{5.68}$\pm$0.51 & \textbf{4.76}$\pm$0.30 & \textbf{21.45}$\pm$1.22 \\
        \cmidrule(lr){1-8}
        KEP-1/7~$+$~TS
        & 84.52$\pm$0.25 & 3.43$\pm$0.10 & 87.85$\pm$0.23 & 63.22$\pm$0.95 & 6.97$\pm$0.19 & 5.36$\pm$0.08 & 23.32$\pm$0.29 \\
        \ourmethod~$+$~TS
        & \textcolor{blue}{\textbf{86.51}}$\pm$0.57 & \textcolor{blue}{\textbf{2.73}}$\pm$0.15 & \textcolor{blue}{\textbf{88.65}}$\pm$0.22 & \textcolor{blue}{\textbf{59.89}}$\pm$0.61 & \textcolor{blue}{\textbf{4.64}}$\pm$0.26 & \textcolor{blue}{\textbf{4.33}}$\pm$0.15 & \textcolor{blue}{\textbf{20.01}}$\pm$0.62 \\
        \bottomrule
    \end{tabular}}
    \end{center}
\end{table*}

As shown in Table~\ref{tab:ts-on-all-cifar10},~\ourmethod{} outperforms each TS-calibrated baseline pairwise on every metric.~Within the ViT pair,~\ourmethod{}~$+$~TS improves ACC from 83.84\% to 85.67\% and ECE from 9.20 to 5.68, while also reducing AURC, NLL, and Brier scores.~The same pattern holds within the KEP-1/7 pair, where every metric improves under~\ourmethod{}, and~\ourmethod{} (from KEP-1/7)~$+$~TS achieves the best value across the whole table on ACC, AURC, AUROC, FPR95, ECE, NLL, and Brier.~These results confirm that the improvements reported in Section~\ref{sec:experiments} originate from the diffusion-based reconfiguration itself, and not from the absence of post-hoc calibration on the baselines, and that the calibration improvements offered by~\ourmethod{} cannot be replicated by simply applying post-hoc TS to existing uncertainty-aware transformers.
}

\subsubsection{Varying covariance structures}
\label{subsubsec:varying-covarience}
The covariance of the pretrained feature distribution, denoted by $\sigma_t(X_t)$, is approximated using a Cholesky-like factorization as employed in KEP-SVGP~\citep{chen2024self}, which we select as the pretrained model for consistency. In addition, we adopt a diagonal parameterization for the learned covariance $\sigma_\theta(X_t, t)$, a design choice that is standard in diffusion models~\citep{ho2020denoising}. 

To examine whether this simplification compromises calibration quality, we conduct a small-scale empirical study comparing the diagonal parameterization against another alternative covariance structure, full parameterization. The results, summarized in Table~\ref{tab:cov_struct}, demonstrate that the diagonal approximation not only yields significantly faster training time but also achieves superior calibration performance relative to the more expressive covariance structure. These observations support the efficiency and effectiveness of the diagonal covariance design in our probabilistic reconfiguration framework.

\qhung{We additionally report the wall-clock training time of the pretrained baselines, which corresponds to the cost of pre-training the underlying ViT or KEP-1/7 backbone for $300$ epochs on CIFAR-10.~As shown in the last column of Table~\ref{tab:cov_struct}, the ViT and KEP-1/7 backbones require roughly $3.85$ and $6.51$ hours respectively, while the unified transition model with the diagonal covariance parameterization is trained on top of either backbone in only $1.75$ hours.~The diffusion-based reconfiguration step in~\ourmethod{} is therefore substantially less expensive than the original pretraining stage, supporting its use as a lightweight post-hoc add-on to existing pretrained transformers.}

\begin{table}[t]
\caption{Comparison of different covariance structures under two pretrained baselines ViT and KEP-1/7.}
\label{tab:cov_struct}
\begin{center}
\setlength\tabcolsep{3pt}
\resizebox{\textwidth}{!}{
\begin{tabular}{llcccccccc}
\toprule
\textbf{Baseline} & \textbf{Covariance Structure} & \textbf{ACC} $\uparrow$ & \textbf{AURC} $\downarrow$ & \textbf{AUROC} $\uparrow$ & \textbf{FPR95} $\downarrow$ & \textbf{ECE} $\downarrow$ & \textbf{NLL} $\downarrow$ & \textbf{Brier} $\downarrow$ & \textbf{Training time (hours)} \\
\midrule
\multirow{3}{*}{ViT}
 & Diagonal              & \textbf{86.38} & \textbf{2.70} & \textbf{88.80} & \textbf{61.16} & \textbf{9.00}  & \textbf{6.34} & \textbf{22.21} & \textbf{1.75} \\
 & Full (dimension-wise) & 81.55 & 4.95 & 85.48 & 68.08 & 10.70 & 7.28 & 28.88 & 2.50 \\
 & Baseline              & 83.90 & 4.02 & 86.71 & 65.16 & 12.57 & 10.94 & 27.80 & 3.85 \\
\midrule
\multirow{3}{*}{KEP-1/7}
 & Diagonal              & \textbf{86.22} & \textbf{2.83} & \textbf{88.35} & \textbf{61.90} & \textbf{8.85}  & \textbf{6.21} & \textbf{22.23} & \textbf{1.75} \\
 & Full (dimension-wise) & 83.80 & 3.81 & 87.26 & 62.10 & 11.16 & 8.12 & 26.34 & 2.50 \\
 & Baseline              & 84.61 & 3.52 & 87.44 & 65.04 & 10.86 & 8.17 & 25.59 & 6.51 \\
\bottomrule
\end{tabular}}
\end{center}
\end{table}

\subsubsection{Computational and memory overhead}
\label{subsubsec:inference-time-mem}
We evaluate the computational and memory overhead of our method against several baselines on the CIFAR-10 dataset. As reported in Table~\ref{tab:infer-time-mem}, our approach achieves substantially lower inference memory consumption than all competing methods, while maintaining an inference time comparable to KEP-1/7.

\begin{table}[t]
\centering
\caption{Inference time and memory consumption on CIFAR-10.}
\label{tab:infer-time-mem}
\begin{tabular}{lcc}
\toprule
\textbf{Method} & \textbf{Inference Time (s)} & \textbf{Inference Memory (GB)} \\
\midrule
KEP-7/7 & 2.1038 & 0.3348 \\
KEP-1/7 & 1.5129 & 0.2916 \\
SGPA    & 4.5169 & 0.5768 \\
ViT     & 1.3173 & 0.1509 \\
\ourmethod & 1.6756 & 0.1040 \\
\bottomrule
\end{tabular}
\end{table}

\subsubsection{Varying unified transition models}
\label{subsubsec:varying-utm}
Regarding kernel design, although score networks in diffusion models are commonly implemented using multilayer perceptrons (MLPs), our empirical results indicate that a DiT-based architecture~\citep{Peebles2022DiT} consistently achieves superior performance. Table~\ref{tab:utm_arch} reports a comparison between MLP and DiT-based score networks on the CIFAR-10 dataset under two different pretrained baselines. While MLP-based score networks exhibit slightly better calibration on certain metrics, they underperform the corresponding baselines on most other evaluation criteria, particularly predictive accuracy. In contrast, the DiT-based architecture improves performance across the majority of metrics without sacrificing stability. Consequently, we adopt DiT as the score network in our framework.

\begin{table}[t]
\centering
\caption{Comparison of unified transition model (UTM) architectures on CIFAR-10 under different pretrained baselines.}
\label{tab:utm_arch}
\begin{tabular}{llccccccc}
\toprule
\textbf{Baseline} & \textbf{UTM} & \textbf{ACC} $\uparrow$ & \textbf{AURC} $\downarrow$ & \textbf{AUROC} $\uparrow$ & \textbf{FPR95} $\downarrow$ & \textbf{ECE} $\downarrow$ & \textbf{NLL} $\downarrow$ & \textbf{Brier} $\downarrow$ \\
\midrule
\multirow{3}{*}{ViT}
 & MLP      & 81.17 & 5.11 & 85.47 & 67.39 & \textbf{1.36}  & \textbf{5.52} & 26.70 \\
 & DiT      & \textbf{86.18} & \textbf{2.95} & \textbf{87.74} & \textbf{63.82} & 8.90  & 6.27 & \textbf{22.62} \\
 & Baseline & 83.90 & 4.02 & 86.71 & 65.16 & 12.57 & 10.94 & 27.80 \\
\midrule
\multirow{3}{*}{KEP-7/7}
 & MLP      & 83.46 & 4.16 & 85.95 & 65.18 & \textbf{4.99}  & \textbf{5.34} & 24.37 \\
 & DiT      & \textbf{87.37} & \textbf{2.43} & \textbf{88.99} & \textbf{60.97} & 8.53  & 6.07 & \textbf{20.84} \\
 & Baseline & 82.34 & 4.42 & 86.45 & 65.80 & 5.81  & 5.59 & 25.68 \\
\bottomrule
\end{tabular}
\end{table}

\subsubsection{Statistical significance testing}
\label{subsubsec:signigicance-testing}
Regarding statistical significance, we report hypothesis testing results for both the CV task (CIFAR-10) and the NLP task (CoLA), corresponding to the results in Table~\ref{tab::in_dist_kep_diff}. For CoLA, we additionally perform an extensive evaluation with 40 runs and compute paired t-tests across metrics.

As summarized in Table~\ref{tab:significance}, the majority of comparisons yield $p$-values below $0.05$, indicating that the observed improvements of our method over the corresponding baselines are statistically significant. This holds consistently across accuracy, uncertainty-aware metrics, and calibration measures, demonstrating the robustness of our performance gains.

\begin{table*}[t]
\centering
\caption{Significance testing results on CIFAR-10 (DF=4) and CoLA (DF=39). We report mean $\pm$ standard error of performance (Perf.) and paired $p$-values.}
\label{tab:significance}
\resizebox{\textwidth}{!}{%
\begin{tabular}{llcccccccccccccccc}
\toprule
\textbf{Dataset} & \textbf{Method} &
\multicolumn{2}{c}{\textbf{ACC / MCC} $\uparrow$}  &
\multicolumn{2}{c}{\textbf{AURC} $\downarrow$}  &
\multicolumn{2}{c}{\textbf{AUROC} $\uparrow$} &
\multicolumn{2}{c}{\textbf{FPR95} $\downarrow$} &
\multicolumn{2}{c}{\textbf{ECE} $\downarrow$} &
\multicolumn{2}{c}{\textbf{NLL} $\downarrow$}&
\multicolumn{2}{c}{\textbf{Brier} $\downarrow$}\\
\cmidrule(lr){3-4} \cmidrule(lr){5-6} \cmidrule(lr){7-8} \cmidrule(lr){9-10} \cmidrule(lr){11-12} \cmidrule(lr){13-14} \cmidrule(lr){15-16}
 & & Perf. & $p$ & Perf. & $p$ & Perf. & $p$ & Perf. & $p$ & Perf. & $p$ & Perf. & $p$ & Perf. & $p$ \\
\midrule
\multirow{6}{*}{\textbf{CIFAR-10}} 
& ViT & 83.84$\pm$0.09 & -- & 4.05$\pm$0.11 & -- & 86.42$\pm$0.37 & -- & 67.13$\pm$1.98 & -- & 12.51$\pm$0.20 & -- & 10.91$\pm$0.39 & -- & 28.03$\pm$0.15 & -- \\
& \textbf{\ourmethod} & \textbf{85.67$\pm$0.88} & 0.0094 & \textbf{3.10$\pm$0.45} & 0.0054 & \textbf{87.90$\pm$1.04} & 0.0255 & \textbf{61.92$\pm$1.85} & 0.0104 & \textbf{9.67$\pm$0.62} & 0.0008 & \textbf{6.94$\pm$0.52} & 0.0001 & \textbf{23.54$\pm$1.46} & 0.0020 \\
\cmidrule(lr){2-16}
& KEP-1/7 & 84.52$\pm$0.25 & -- & 3.52$\pm$0.11 & -- & 87.52$\pm$0.27 & -- & 65.14$\pm$1.27 & -- & 10.93$\pm$0.26 & -- & 8.21$\pm$0.15 & -- & 25.80$\pm$0.40 & -- \\
& \textbf{\ourmethod} & \textbf{86.51$\pm$0.57} & 0.0051 & \textbf{2.78$\pm$0.16} & 0.0034 & \textbf{88.28$\pm$0.16} & 0.0065 & \textbf{61.85$\pm$1.93} & 0.0149 & \textbf{8.61$\pm$0.49} & 0.0020 & \textbf{6.11$\pm$0.27} & 0.0003 & \textbf{21.90$\pm$0.82} & 0.0020 \\
\cmidrule(lr){2-16}
& KEP-7/7 & 82.68$\pm$0.12 & -- & 4.46$\pm$0.05 & -- & 85.71$\pm$0.34 & -- & 66.90$\pm$1.78 & -- & 6.95$\pm$0.36 & -- & 5.89$\pm$0.11 & -- & 25.90$\pm$0.23 & -- \\
& \textbf{\ourmethod} & \textbf{84.74$\pm$0.40} & 0.0000 & \textbf{3.47$\pm$0.18} & 0.0000 & \textbf{87.34$\pm$0.33} & 0.0001 & \textbf{62.88$\pm$1.20} & 0.0023 & \textbf{5.18$\pm$0.33} & 0.0000 & \textbf{4.84$\pm$0.11} & 0.0000 & \textbf{22.48$\pm$0.53} & 0.0000 \\
\midrule
\multirow{8}{*}{\textbf{CoLA}}
& Transformer & 29.92$\pm$1.06 & -- & 20.80$\pm$1.09 & -- & 64.22$\pm$1.33 & -- & 90.01$\pm$2.58 & -- & 26.44$\pm$1.72 & -- & 19.66$\pm$3.79 & -- & 55.09$\pm$2.43 & -- \\
& \textbf{\ourmethod} & \textbf{31.97$\pm$2.26} & 0.0000 & \textbf{19.87$\pm$1.46} & 0.0004 & 64.52$\pm$1.44 & -- & \textbf{89.94$\pm$2.02} & 0.8877 & \textbf{23.33$\pm$4.51} & 0.0001 & \textbf{15.30$\pm$5.62} & 0.0001 & 51.04$\pm$4.80 & 0.0000 \\
\cmidrule(lr){2-16}
& KEP-1/5 & 32.40$\pm$2.32 & -- & 19.94$\pm$1.66 & -- & 64.42$\pm$2.36 & -- & 90.21$\pm$2.12 & -- & 22.45$\pm$4.94 & -- & 15.24$\pm$5.89 & -- & 50.27$\pm$4.65 & -- \\
& \textbf{\ourmethod} & \textbf{31.41$\pm$2.05} & 0.0429 & \textbf{18.91$\pm$1.90} & 0.0098 & \textbf{65.45$\pm$1.72} & 0.0269 & \textbf{89.58$\pm$1.61} & 0.1484 & \textbf{17.47$\pm$8.87} & 0.0027 & \textbf{11.87$\pm$5.96} & 0.0128 & \textbf{46.62$\pm$6.91} & 0.0069 \\
\cmidrule(lr){2-16}
& KEP-5/5 & 31.36$\pm$2.73 & -- & 19.80$\pm$1.92 & -- & 64.66$\pm$0.98 & -- & 90.26$\pm$1.15 & -- & 19.87$\pm$5.69 & -- & 14.74$\pm$8.28 & -- & 48.69$\pm$5.57 & -- \\
& \textbf{\ourmethod} & \textbf{31.53$\pm$1.95} & 0.7545 & \textbf{18.58$\pm$2.49} & 0.0121 & \textbf{65.33$\pm$1.93} & 0.0361 & \textbf{88.70$\pm$1.63} & 0.0000 & \textbf{15.74$\pm$8.57} & 0.0139 & \textbf{10.80$\pm$5.91} & 0.0174 & \textbf{44.76$\pm$6.98} & 0.0076 \\
\bottomrule
\end{tabular}%
}
\end{table*}

\subsubsection{Comparison with Knowledge Distillation}
\label{subsubsec:kd}
\begin{table}[t]
    \caption{Performance comparison between knowledge distillation methods and our method on CIFAR-10 and CIFAR-100. Mean $\pm$ std are reported over five trials.}
    \setlength\tabcolsep{3pt} 
    \label{tab::distillation}
    \begin{center}
    \resizebox{\columnwidth}{!}{
    \begin{tabular}{ll|l|cccc|cccc}
        \toprule
        \multicolumn{2}{c|}{\multirow{2}{*}{\textbf{Method}}} & \multirow{2}{*}{\#\textbf{params}} & \multicolumn{4}{c|}{\textbf{CIFAR-10}} & \multicolumn{4}{c}{\textbf{CIFAR-100}} \\
        \cmidrule(lr){4-7} \cmidrule(lr){8-11}
        & & & \textbf{ACC} $\uparrow$ & \textbf{ECE} $\downarrow$ & \textbf{NLL} $\downarrow$ & \textbf{Brier} $\downarrow$ & \textbf{ACC} $\uparrow$ & \textbf{ECE} $\downarrow$ & \textbf{NLL} $\downarrow$ & \textbf{Brier} $\downarrow$ \\
        \midrule
        Teacher & ViT (7 layers) & 6.27M & 83.84$\pm$0.09 & 12.51$\pm$0.20 & 10.91$\pm$0.39 & 28.03$\pm$0.15 & 52.94$\pm$0.63 & 30.73$\pm$0.61 & 32.71$\pm$0.91 & 74.72$\pm$1.20 \\
        \midrule
        \multirow{2}{*}{Student} & ViT (3 layers) & 2.75M & 82.20$\pm$0.80 & 10.92$\pm$1.53 & 7.90$\pm$0.80 & 28.58$\pm$1.20 & 54.86$\pm$0.65 & 30.12$\pm$0.39 & 31.25$\pm$0.53 & 72.17$\pm$0.78 \\
        & DiT & 2.70M & 84.77$\pm$0.63 & 11.93$\pm$0.51 & 10.26$\pm$0.52 & 26.55$\pm$1.17 & 57.56$\pm$0.37 & 27.37$\pm$0.62 & 27.10$\pm$0.77 & 67.06$\pm$0.83 \\
        \midrule
        \multicolumn{2}{l|}{\ourmethod} & 2.70M & \textbf{85.67}$\pm$0.88 & \textbf{9.67}$\pm$0.62 & \textbf{6.94}$\pm$0.52 & \textbf{23.54}$\pm$1.46 & \textbf{57.79}$\pm$0.57 & \textbf{22.31}$\pm$0.49 & \textbf{22.16}$\pm$0.36 & \textbf{62.57}$\pm$0.85 \\
        \bottomrule
    \end{tabular}}
    \end{center}
\end{table}
We compare \ourmethod~with knowledge distillation (KD)~\cite{hinton2015distillingknowledgeneuralnetwork}. Results are presented in Table \ref{tab::distillation}. We evaluate KD using a pretrained 7-layer ViT as the teacher and a 3-layer ViT or our DiT as the student, trained with KD (temperature scaling = 3), while \ourmethod~use DiT and trained with our loss function. Although KD improves accuracy, its calibration metrics remain higher compared to \ourmethod, which achieves superior accuracy and calibration. Additionally, our propagation model is highly parameter-efficient, using only 2.7M parameters, less than half of the original model's 6.27M parameters, resulting in improved memory efficiency.

\subsubsection{In-Distribution Classification}
We consolidate the experimental results from Table~\ref{tab::in_dist_kep_diff} and Table~\ref{tab::in_dist_baselines_diff}, and further include additional findings based on our alignment with KEP-2/7 (vision) and KEP-2/5 (language) configurations, presented in Table~\ref{tab::in_dist_appendix}. \ourmethod{} demonstrates state-of-the-art performance across nearly all scenarios, achieving the \textcolor{blue}{\textbf{highest}} score in 23 out of 28 settings and ranking \textcolor{brown}{\textbf{second}} in another 23 out of 28. These results highlight the substantial advantage of \ourmethod{} over existing baselines, in terms of both predictive accuracy and uncertainty calibration.

\begin{table*}
    \caption{Performance comparison for in-distribution classification across four tasks. KEP-$k$/$n$ denotes a pre-trained transformer using GP-reparameterized architecture (KEP~\citep{chen2024self}) for the last $k$ attention blocks and standard MHSA for the remaining blocks. \textbf{Bold} indicates the better performance in each pairwise comparison between a pre-trained model (ViT, Transformer, KEP) and diffusion-based reconfigured KEP~\citep{chen2024self} produced by~\ourmethod{} . \textcolor{blue}{\textbf{Blue}} marks the best result across all baselines for a dataset, and \textcolor{brown}{\textbf{brown}} denotes the second-best.}
    \label{tab::in_dist_appendix}
    \begin{center}
    \setlength\tabcolsep{3pt} 
    \resizebox{0.78\textwidth}{!}{
    \begin{tabular}{c|l|lllllll}
        \toprule
        \textbf{Dataset} & \textbf{Method} & \textbf{ACC/MCC} $\uparrow$ & \textbf{AURC} $\downarrow$ & \textbf{AUROC} $\uparrow$ & \textbf{FPR95} $\downarrow$ & \textbf{ECE} $\downarrow$ & \textbf{NLL} $\downarrow$ & \textbf{Brier} $\downarrow$ 
        \\ \midrule
        \multirow{13}{*}{\rotatebox{90}{\parbox{2cm}{\centering CIFAR-10}}} 
        & TS 
        & 83.84$\pm$0.09 & 3.88$\pm$0.10 & 86.82$\pm$0.37 & 65.99$\pm$1.94 & 9.22$\pm$0.36 & 6.58$\pm$0.16 & 25.50$\pm$0.13 \\
        & MCD 
        & 84.06$\pm$0.23 & 8.65$\pm$0.03 & 86.51$\pm$0.32 & 66.15$\pm$0.60 & 9.47$\pm$0.16 & 8.36$\pm$0.32 & 25.45$\pm$0.29 \\
        & KFLLA 
        & 83.84$\pm$0.10 & 3.91$\pm$0.11 & 86.71$\pm$0.45 & 65.44$\pm$1.58 & 8.18$\pm$0.80 & 6.09$\pm$0.40 & 24.98$\pm$0.59 \\
        & SV-DKL 
        & 83.23$\pm$0.17 & 4.39$\pm$0.18 & 85.94$\pm$0.36 & 66.96$\pm$1.30 & 11.64$\pm$0.81 & 9.85$\pm$1.09 & 27.97$\pm$0.77 \\
        & SGPA 
        & 75.59$\pm$3.63 & 8.41$\pm$2.37 & 82.65$\pm$1.71 & 71.78$\pm$2.73 & \textcolor{blue}{\textbf{1.92}}$\pm$0.55 & 7.11$\pm$0.95 & 33.98$\pm$4.57 \\
        \cmidrule(lr){2-9}
        & ViT       
        & 83.84$\pm$0.09 & 4.05$\pm$0.11 & 86.42$\pm$0.37 & 67.13$\pm$1.98 & 12.51$\pm$0.20 & 10.91$\pm$0.39 & 28.03$\pm$0.15 \\
        & \ourmethod         
        & \textbf{85.67}$\pm$0.88 & \textbf{3.10}$\pm$0.45 & \textbf{87.90}$\pm$1.04 & \textbf{61.92}$\pm$1.85 & \textbf{9.67}$\pm$0.62 & \textbf{6.94}$\pm$0.52 & \textbf{23.54}$\pm$1.46 \\
        \cmidrule(lr){2-9}
        & KEP-1/7
        & 84.52$\pm$0.25 & 3.52$\pm$0.11 & 87.52$\pm$0.27 & 65.14$\pm$1.27 & 10.93$\pm$0.26 & 8.21$\pm$0.15 & 25.80$\pm$0.40 \\
        & \ourmethod
        & \textbf{\textcolor{brown}{86.51}}$\pm$0.57 & \textbf{\textcolor{brown}{2.78}}$\pm$0.16 & \textbf{\textcolor{brown}{88.28}}$\pm$0.16 & \textbf{\textcolor{brown}{61.85}}$\pm$1.93 & \textbf{8.61}$\pm$0.49 & \textbf{6.11}$\pm$0.27 & \textbf{\textcolor{brown}{21.90}}$\pm$0.82 \\
        \cmidrule(lr){2-9}
        & KEP-2/7
        & 84.32$\pm$0.75 & 3.65$\pm$0.22 & 87.17$\pm$0.35 & 65.30$\pm$1.35 & 11.11$\pm$0.46 & 8.43$\pm$0.26 & 26.20$\pm$1.08 \\
        & \ourmethod
        & \textbf{\textcolor{blue}{86.62}}$\pm$0.18 & \textbf{\textcolor{blue}{2.70}}$\pm$0.07 & \textbf{\textcolor{blue}{88.52}}$\pm$0.36 & \textbf{\textcolor{blue}{60.70}}$\pm$2.31 & \textbf{8.54}$\pm$0.14 & \textbf{5.96}$\pm$0.14 & \textbf{\textcolor{blue}{21.59}}$\pm$0.25 \\
        \cmidrule(lr){2-9}
        & KEP-7/7
        & 82.68$\pm$0.12 & 4.46$\pm$0.05 & 85.71$\pm$0.34 & 66.90$\pm$1.78 & 6.95$\pm$0.36 & \textcolor{brown}{\textbf{5.89}}$\pm$0.11 & 25.90$\pm$0.23 \\
        & \ourmethod
        & \textbf{84.74}$\pm$0.40 & \textbf{3.47}$\pm$0.18 & \textbf{87.34}$\pm$0.33 & \textbf{62.88}$\pm$1.20 & \textbf{\textcolor{brown}{5.18}}$\pm$0.33 & \textbf{\textcolor{blue}{4.84}}$\pm$0.11 & \textbf{22.48}$\pm$0.53 \\
        \midrule
        \multirow{13}{*}{\rotatebox{90}{\parbox{2cm}{\centering CIFAR-100}}}
        & TS 
        & 52.94$\pm$0.63 & 22.34$\pm$0.61 & 82.29$\pm$0.48 & 71.65$\pm$1.98 & \textcolor{brown}{\textbf{17.06}}$\pm$0.42 & 21.57$\pm$0.52 & 64.77$\pm$0.95 \\
        & MCD 
        & 53.49$\pm$0.62 & 22.24$\pm$0.56 & 81.60$\pm$0.19 & 73.02$\pm$0.51 & 25.93$\pm$0.37 & 29.24$\pm$0.73 & 70.02$\pm$0.92 \\
        & KFLLA 
        & 52.27$\pm$0.86 & 23.96$\pm$0.78 & 81.30$\pm$0.48 & 71.42$\pm$1.92 & 18.52$\pm$5.40 & 20.89$\pm$0.57 & 66.51$\pm$1.66 \\
        & SV-DKL
        & 51.03$\pm$0.60 & 24.38$\pm$0.43 & 81.32$\pm$0.50 & 73.99$\pm$1.40 & 25.46$\pm$0.72 & 28.93$\pm$0.66 & 71.90$\pm$0.74 \\
        & SGPA 
        & 52.77$\pm$0.52 & 22.84$\pm$0.52 & 81.65$\pm$0.36 & 72.02$\pm$1.74 & \textcolor{blue}{\textbf{10.33}}$\pm$2.25 & \textcolor{blue}{\textbf{19.10}}$\pm$0.57 & 62.08$\pm$1.04 \\
        \cmidrule(lr){2-9}
        & ViT       
        & 52.94$\pm$0.63 & 22.88$\pm$0.64 & 81.07$\pm$0.56 & 75.96$\pm$2.39 & 30.73$\pm$0.61 & 32.71$\pm$0.91 & 74.72$\pm$1.20 \\
        & \ourmethod              
        & \textbf{57.79}$\pm$0.57 & \textbf{18.58}$\pm$0.37 & \textbf{82.60}$\pm$0.11 & \textbf{71.50}$\pm$1.42 & \textbf{22.31}$\pm$0.49 & \textbf{22.16}$\pm$0.36 & \textbf{62.57}$\pm$0.85 \\
        \cmidrule(lr){2-9}
        & KEP-1/7      
        & 55.74$\pm$0.77 & 20.20$\pm$0.64 & 82.10$\pm$0.20 & 73.82$\pm$0.92 & 27.07$\pm$0.71 & 27.45$\pm$0.62 & 68.54$\pm$1.18 \\
        & \ourmethod              
        & \textbf{58.78}$\pm$1.52 & \textbf{17.63}$\pm$1.08 & \textbf{\textcolor{brown}{82.99}}$\pm$0.41 & \textbf{\textcolor{brown}{71.40}}$\pm$1.24 & \textbf{21.17}$\pm$1.03 & \textbf{21.15}$\pm$0.93 & \textbf{60.96}$\pm$1.73 \\
        \cmidrule(lr){2-9}
        & KEP-2/7      
        & 55.82$\pm$1.12 & 20.14$\pm$0.94 & 82.16$\pm$0.32 & 73.34$\pm$1.05 & 27.37$\pm$0.55 & 27.74$\pm$0.51 & 68.65$\pm$1.42 \\
        & \ourmethod              
        & \textbf{\textcolor{brown}{59.45}}$\pm$1.06 & \textbf{\textcolor{brown}{17.10}}$\pm$0.80 & \textbf{\textcolor{blue}{83.27}}$\pm$0.33 & \textbf{\textcolor{blue}{69.94}}$\pm$1.33 & \textbf{21.44}$\pm$0.86 & \textbf{20.85}$\pm$0.49 & \textbf{\textcolor{brown}{60.15}}$\pm$1.16 \\
        \cmidrule(lr){2-9} 
        & KEP-7/7      
        & 57.06$\pm$0.56 & 19.38$\pm$0.60 & 82.02$\pm$0.39 & 72.78$\pm$0.69 & \textbf{21.31}$\pm$3.85 & 22.41$\pm$2.29 & 63.07$\pm$2.11 \\
        & \ourmethod              
        & \textbf{\textcolor{blue}{60.85}}$\pm$2.98 & \textbf{\textcolor{blue}{16.35}}$\pm$2.34 & \textbf{82.91}$\pm$0.67 & \textbf{71.46}$\pm$1.43 & 21.43$\pm$2.09 & \textbf{\textcolor{brown}{20.55}}$\pm$2.42 & \textbf{\textcolor{blue}{58.91}}$\pm$4.63 \\
        \midrule
        \multirow{13}{*}{\rotatebox{90}{\parbox{2cm}{\centering IMDB}}} 
        & TS
        & 85.59$\pm$0.50 & 4.73$\pm$0.32 & 80.75$\pm$0.55 & 75.45$\pm$1.10 & \textcolor{brown}{\textbf{2.91}}$\pm$1.51 & 3.41$\pm$0.13 & 21.04$\pm$0.74 \\
        & MCD
        & 85.96$\pm$0.42 & 4.40$\pm$0.24 & 81.40$\pm$0.55 & 74.79$\pm$0.88 & 4.18$\pm$2.03 & 3.47$\pm$0.23 & 20.72$\pm$0.82 \\
        & KFLLA
        & 85.59$\pm$0.50 & 4.71$\pm$0.30 & 80.82$\pm$0.48 & 75.45$\pm$1.11 & 5.84$\pm$2.21 & 6.93$\pm$0.00 & 21.86$\pm$1.19 \\
        & SV-DKL
        & 85.69$\pm$0.66 & 5.58$\pm$0.79 & 78.54$\pm$2.20 & 75.32$\pm$0.84 & 8.52$\pm$1.57 & 4.49$\pm$0.65 & 23.10$\pm$1.66 \\
        & SGPA
        & 85.39$\pm$0.36 & 4.96$\pm$0.49 & 80.04$\pm$1.14 & 76.44$\pm$0.96 & 6.04$\pm$1.71 & 3.96$\pm$0.50 & 22.19$\pm$1.06 \\
        \cmidrule(lr){2-9}
        & Transformer     
        & 85.59$\pm$0.50 & 4.73$\pm$0.32 & 80.75$\pm$0.55 & 75.45$\pm$1.10 & 6.96$\pm$2.05 & 3.95$\pm$0.44 & 22.28$\pm$1.26 \\
        & \ourmethod              
        & \textbf{86.07}$\pm$0.61 & \textbf{4.57}$\pm$0.39 & \textbf{80.84}$\pm$0.67 & \textbf{74.24}$\pm$0.87 & \textbf{5.40}$\pm$1.90 & \textbf{3.60}$\pm$0.34 & \textbf{21.08}$\pm$1.32 \\
        \cmidrule(lr){2-9}
        & KEP-1/5    
        & 85.76$\pm$0.71 & 4.54$\pm$0.42 & 81.02$\pm$0.70 & 74.87$\pm$0.87 & 5.51$\pm$2.94 & 3.79$\pm$0.51 & 21.62$\pm$1.42 \\
        & \ourmethod              
        & \textbf{\textcolor{blue}{87.13}}$\pm$0.19 & \textbf{\textcolor{brown}{4.07}}$\pm$0.30 & \textbf{\textcolor{brown}{81.55}}$\pm$0.62 & \textbf{\textcolor{blue}{73.35}}$\pm$0.14 & \textbf{3.16}$\pm$2.54 & \textbf{\textcolor{brown}{3.24}}$\pm$0.31 & \textbf{\textcolor{blue}{19.31}}$\pm$0.97 \\
        \cmidrule(lr){2-9}
        & KEP-2/5       
        & 86.52$\pm$0.72 & 4.18$\pm$0.37 & 81.54$\pm$0.54 & 73.82$\pm$1.68 & 5.72$\pm$1.20 & 3.58$\pm$0.29 & 20.51$\pm$1.16 \\
        & \ourmethod              
        & \textbf{\textcolor{brown}{86.95}}$\pm$0.19 & \textbf{\textcolor{blue}{4.05}}$\pm$0.30 & \textbf{\textcolor{blue}{81.64}}$\pm$0.85 & \textbf{\textcolor{brown}{73.46}}$\pm$1.34 & \textbf{3.46}$\pm$1.40 & \textbf{\textcolor{blue}{3.23}}$\pm$0.13 & \textbf{\textcolor{brown}{19.42}}$\pm$0.49 \\
        \cmidrule(lr){2-9} 
        & KEP-5/5     
        & 84.57$\pm$0.81 & 5.48$\pm$0.60 & 79.32$\pm$1.25 & 77.03$\pm$1.48 & 7.83$\pm$3.28 & 5.17$\pm$2.13 & 24.23$\pm$2.56 \\
        & \ourmethod              
        & \textbf{85.74}$\pm$0.34 & \textbf{4.58}$\pm$0.23 & \textbf{80.95}$\pm$0.42 & \textbf{75.08}$\pm$1.15 & \textbf{\textcolor{blue}{2.33}}$\pm$1.54 & \textbf{3.36}$\pm$0.12 & \textbf{20.70}$\pm$0.65 \\
        \midrule
        \multirow{13}{*}{\rotatebox{90}{\parbox{2cm}{\centering CoLA}}}
        & TS 
        & 29.92$\pm$1.17 & 20.84$\pm$1.23 & 64.31$\pm$1.44 & 89.93$\pm$2.95 & 23.22$\pm$2.99 & 11.04$\pm$1.91 & 51.70$\pm$3.42 \\
        & MCD 
        & 30.04$\pm$1.02 & 20.66$\pm$1.11 & 64.53$\pm$1.00 & 89.51$\pm$1.35 & 24.96$\pm$1.79 & 17.83$\pm$3.61 & 53.52$\pm$2.59 \\
        & KFLLA 
        & 29.89$\pm$1.14 & 20.82$\pm$1.26 & 64.22$\pm$1.46 & 89.87$\pm$3.24 & 24.36$\pm$2.25 & 12.16$\pm$1.49 & 52.80$\pm$2.85 \\
        & SV-DKL 
        & 30.07$\pm$1.41 & 22.76$\pm$2.28 & 61.98$\pm$3.09 & \textcolor{blue}{\textbf{89.00}}$\pm$2.55 & 25.71$\pm$1.60 & 17.96$\pm$3.26 & 54.40$\pm$2.13 \\
        & SGPA 
        & 31.53$\pm$2.05 & 20.44$\pm$2.60 & 64.34$\pm$1.95 & 90.79$\pm$0.87 & 26.22$\pm$1.51 & 28.65$\pm$7.23 & 54.08$\pm$2.44 \\
        \cmidrule(lr){2-9}
        & Transformer       
        & 29.92$\pm$1.17 & 20.80$\pm$1.21 & 64.22$\pm$1.46 & 90.01$\pm$2.84 & 26.44$\pm$1.90 & 19.66$\pm$4.18 & 55.09$\pm$2.68 \\
        & \ourmethod              
        & \textbf{31.85}$\pm$2.46 & \textbf{19.74}$\pm$1.62 & \textbf{64.52}$\pm$1.71 & \textbf{89.52}$\pm$3.61 & \textbf{23.94}$\pm$0.49 & \textbf{14.29}$\pm$3.01 & \textbf{50.82}$\pm$1.19 \\
        \cmidrule(lr){2-9}
        & KEP-1/5      
        & 30.86$\pm$2.03 & 19.86$\pm$2.04 & \textcolor{brown}{\textbf{65.18}}$\pm$1.83 & \textbf{\textcolor{brown}{89.10}}$\pm$2.60 & 24.98$\pm$2.08 & 16.28$\pm$4.35 & 52.86$\pm$2.88 \\
        & \ourmethod              
        & \textbf{31.84}$\pm$1.88 & \textbf{19.42}$\pm$1.83 & \textbf{\textcolor{blue}{65.54}}$\pm$1.78 & 90.37$\pm$0.89 & \textbf{13.77}$\pm$6.58 & \textbf{8.26}$\pm$3.75 & \textbf{43.54}$\pm$3.65 \\
        \cmidrule(lr){2-9}
        & KEP-2/5     
        & 30.73$\pm$2.29 & 20.36$\pm$1.40 & 64.03$\pm$1.21 & 90.23$\pm$1.86 & 21.78$\pm$6.62 & 15.52$\pm$7.74 & 50.14$\pm$5.92 \\
        & \ourmethod              
        & \textbf{\textcolor{blue}{33.03}}$\pm$1.36 & \textbf{\textcolor{brown}{19.06}}$\pm$3.05 & \textbf{64.78}$\pm$2.31 & \textbf{89.79}$\pm$2.65 & \textbf{\textcolor{blue}{11.45}}$\pm$4.78 & \textbf{\textcolor{blue}{6.72}}$\pm$0.98 & \textbf{\textcolor{blue}{41.64}}$\pm$3.58 \\
        \cmidrule(lr){2-9} 
        & KEP-5/5     
        & 29.28$\pm$1.21 & 20.82$\pm$1.98 & 64.47$\pm$0.90 & 89.65$\pm$1.04 & 18.95$\pm$5.66 & 11.83$\pm$7.96 & 48.65$\pm$5.84 \\
        & \ourmethod              
        & \textbf{\textcolor{brown}{32.05}}$\pm$1.56 & \textbf{\textcolor{blue}{18.69}}$\pm$1.39 & \textbf{64.71}$\pm$1.37 & \textbf{89.53}$\pm$1.67 & \textbf{\textcolor{brown}{12.88}}$\pm$5.74 & \textbf{\textcolor{brown}{7.68}}$\pm$1.70 & \textbf{\textcolor{brown}{42.20}}$\pm$2.50 \\
        \bottomrule
    \end{tabular}}
    \end{center}
\end{table*}

\subsubsection{Distribution shift robustness}

\begin{figure}[t]
    \centering
    \includegraphics[width=\linewidth]{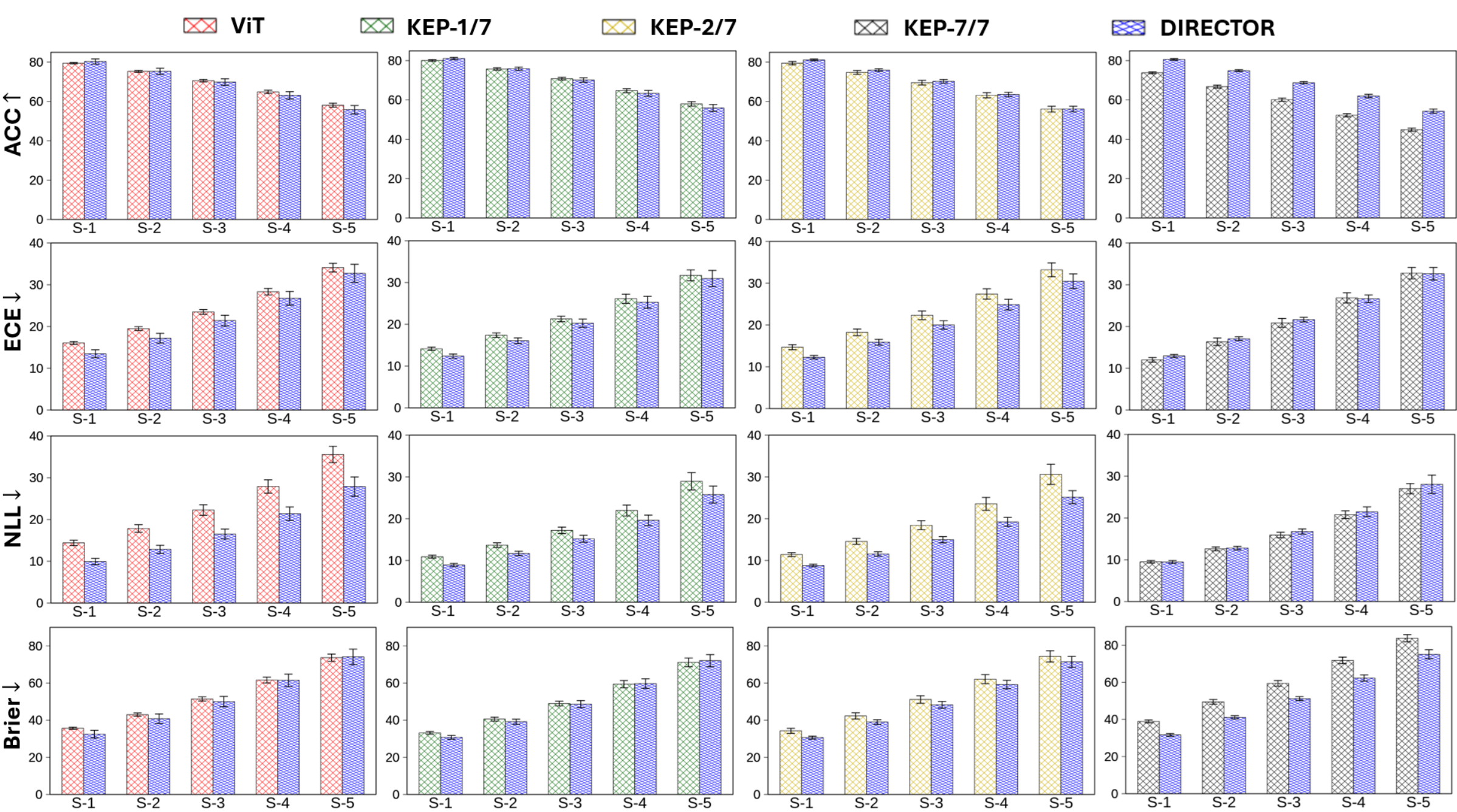}
    \caption{Calibration comparison of pre-trained models with their corresponding diffusion-based reconfigured produced by~\ourmethod{} on CIFAR-10-C over 5 severity levels of corruption. The notation $\texttt{S-k}$ represents the severity level $k$. \ourmethod{} achieves competitive accuracy and outperforms pre-trained models in most calibration metrics.}
    \label{fig:cifar-10-c-five-severity}
\end{figure}

\begin{table}[t]
    \caption{Performance comparison on CIFAR-10-C, with 15 corruptions across five severity levels over five trials. }
    \vspace{-3mm}
    \label{tab::dis_shift_appendix}
    \begin{center}
    \resizebox{\columnwidth}{!}{
    \begin{tabular}{l|lllllll}
        \toprule
        \textbf{Method} 
        & \textbf{ACC} $\uparrow$ 
        & \textbf{AURC} $\downarrow$ 
        & \textbf{AUROC} $\uparrow$ & \textbf{FPR95} $\downarrow$ & \textbf{ECE} $\downarrow$ & \textbf{NLL} $\downarrow$ & \textbf{Brier} $\downarrow$ 
        \\ \midrule
        MCD
        & 69.73$\pm$0.36 & 14.46$\pm$0.22 & 79.11$\pm$0.11 & 76.96$\pm$0.20 & 19.58$\pm$0.23 & 18.48$\pm$0.67 & 48.70$\pm$0.47 \\
        KFLLA & 69.64$\pm$0.43 & 14.46$\pm$0.41 & 79.27$\pm$0.23 & 76.34$\pm$0.33 & 17.46$\pm$1.18 & 12.75$\pm$0.80 & 47.18$\pm$1.06 \\
        SVDKL & 68.88$\pm$0.36 & 15.43$\pm$0.56 & 78.56$\pm$0.50 & 77.22$\pm$0.55 & 22.67$\pm$1.18 & 21.02$\pm$2.57 & 52.29$\pm$1.35 \\
        SGPA & 57.73$\pm$1.25 & 24.82$\pm$1.23 & 74.81$\pm$0.65 & 80.92$\pm$0.54 & 12.47$\pm$2.11 & 13.60$\pm$0.48 & 57.91$\pm$1.39 
        \\ \midrule
        ViT             
        & \textbf{69.67}$\pm$0.34 & \textbf{14.66}$\pm$0.27 & 78.92$\pm$0.16 & 77.74$\pm$0.28 & 24.30$\pm$0.31 & 23.59$\pm$1.00 & 53.07$\pm$0.59 
        \\
        \ourmethod   
        & 68.89$\pm$1.44 & 15.31$\pm$1.30 & \textbf{79.17}$\pm$0.86 & \textbf{77.10}$\pm$1.00 & \textbf{22.32}$\pm$1.09 & \textbf{17.71}$\pm$1.02 & \textbf{51.77}$\pm$2.39
        \\ \midrule
        KEP-1/7
        & \textbf{69.87}$\pm$0.45 & \textbf{14.30}$\pm$0.50 & 79.49$\pm$0.38 & 76.95$\pm$0.43 & 22.12$\pm$0.47 & 18.54$\pm$0.63 & 50.65$\pm$0.90
        \\
        \ourmethod 
        & 69.29$\pm$0.66 & 15.00$\pm$0.69 & \textbf{79.69}$\pm$0.27 & \textbf{76.38}$\pm$0.40 & \textbf{20.98}$\pm$0.65 & \textbf{16.23}$\pm$0.60 & \textbf{50.07}$\pm$1.20
        \\ \midrule
        KEP-2/7      
        & 68.63$\pm$0.75 & 15.46$\pm$0.66 & 78.90$\pm$0.37 & 77.57$\pm$0.42 & 23.18$\pm$0.63 & 19.72$\pm$0.76 & 52.82$\pm$1.30 
        \\ 
        \ourmethod   
        & \textbf{69.41}$\pm$0.55 & \textbf{14.95}$\pm$0.60 & \textbf{79.73}$\pm$0.42 & \textbf{76.29}$\pm$0.73 & \textbf{20.71}$\pm$0.63 & \textbf{15.94}$\pm$0.44 & \textbf{49.71}$\pm$1.10
        \\ \midrule
        KEP-7/7       
        & 59.57$\pm$0.30 & 23.77$\pm$0.40 & 75.56$\pm$0.30 & 80.47$\pm$0.27 & \textbf{21.78}$\pm$0.59 & \textbf{17.17}$\pm$0.39 & 60.67$\pm$0.72
        \\
        \ourmethod  
        & \textbf{68.12}$\pm$0.26 & \textbf{16.45}$\pm$0.28 & \textbf{79.08}$\pm$0.22 & \textbf{76.73}$\pm$0.29 & 22.19$\pm$0.19 & 17.70$\pm$0.17 & \textbf{52.27}$\pm$0.35
        \\ \bottomrule
    \end{tabular}}
    \end{center}
    \vspace{-3mm}
\end{table}

Additional experiments evaluating the distributional robustness of \ourmethod{} are presented in Tables~\ref{tab::dis_shift_appendix},\ref{tab:dis_shift_cola_appendix}, and Figure\ref{fig:cifar-10-c-five-severity}. These results include evaluations of \ourmethod{} aligned with KEP-2/5 and KEP-2/7 configurations. Notably, the TS baseline is excluded from these comparisons, as it is specifically tailored for in-distribution tasks.

Figure~\ref{fig:cifar-10-c-five-severity} illustrates a calibration comparison between pre-trained models (ViT, KEP) and \ourmethod{} on the CIFAR-10-C dataset, across five corruption severity levels averaged over 15 corruption types. \ourmethod{} demonstrates competitive, and in many cases superior, predictive accuracy, particularly when aligned with KEP-7/7. In terms of uncertainty calibration, \ourmethod{} exhibits significantly improved calibration, achieving lower values in ECE, NLL, and Brier score compared to pre-trained baselines.

Table~\ref{tab::dis_shift_appendix} summarizes performance averaged over the 15 corruptions and five severity levels on CIFAR-10-C. While maintaining competitive predictive accuracy with ViT and KEP, \ourmethod{} substantially outperforms them in calibration metrics. When aligned with KEP-2/7, our best-performing configuration of \ourmethod{} achieves both competitive accuracy and slightly improved calibration compared to other baselines. Additionally, \ourmethod{} can be enhanced by integrating post-training baselines such as MCD or aligning with attention-modified methods like SGPA to further improve calibration. However, due to computational constraints, we leave these extensions for future investigation.

Finally, Table~\ref{tab:dis_shift_cola_appendix} compares \ourmethod{} against pre-trained models (Transformer and KEP) and other baselines on the CoLA OOD dataset. \ourmethod{} achieves the highest MCC score, outperforming all baselines, and shows significant gains in both calibration and failure prediction metrics.

\begin{table}[t]
    \caption{Performance comparison on CoLA OOD over five trials.}
    \vspace{-3mm}
    \label{tab:dis_shift_cola_appendix}
    \begin{center}
    \resizebox{\columnwidth}{!}
    {
    \begin{tabular}{l|lllllll}
        \toprule
        \textbf{Method} 
        & \textbf{MCC} $\uparrow$ 
        & \textbf{AURC} $\downarrow$ 
        & \textbf{AUROC} $\uparrow$ & \textbf{FPR95} $\downarrow$ & \textbf{ECE} $\downarrow$ & \textbf{NLL} $\downarrow$ & \textbf{Brier} $\downarrow$ 
        \\ \midrule
        MC Dropout 
        & 18.54$\pm$4.14 & 25.85$\pm$1.03 & 63.31$\pm$2.18 & 90.17$\pm$2.95 & 32.02$\pm$2.56 & 23.84$\pm$5.19 & 65.50$\pm$4.46 \\
        KFLLA 
        & 18.43$\pm$3.55 & 25.89$\pm$0.99 & 63.31$\pm$1.65 & 90.50$\pm$3.25 & 29.94$\pm$2.86 & 14.64$\pm$1.87 & 62.72$\pm$4.37 \\
        SVDKL 
        & 19.32$\pm$3.57 & 27.58$\pm$2.51 & 60.65$\pm$1.41 & 89.97$\pm$2.67 & 30.97$\pm$3.10 & 21.23$\pm$3.67 & 64.07$\pm$4.62 \\
        SGPA 
        & 19.34$\pm$6.23 & 27.48$\pm$2.55 & 61.68$\pm$2.29 & 90.14$\pm$2.30 & 31.15$\pm$1.66 & 35.18$\pm$8.59 & 63.62$\pm$3.65
        \\ \midrule
        Transformer             
        & 18.43$\pm$3.55 & 25.85$\pm$1.00 & \textbf{63.38}$\pm$1.69 & 90.39$\pm$3.27 & 31.99$\pm$2.70 & 23.82$\pm$5.16 & 65.50$\pm$4.32 
        \\
        \ourmethod   
        & \textbf{23.06}$\pm$4.69 & \textbf{25.61}$\pm$2.30 & 61.32$\pm$3.21 & \textbf{88.23}$\pm$3.89 & \textbf{28.72}$\pm$1.87 & \textbf{17.18}$\pm$3.75 & \textbf{59.91}$\pm$2.20
        \\ \midrule
        KEP-1/5
        & 19.44$\pm$1.94 & 25.21$\pm$1.53 & \textbf{63.65}$\pm$3.19 & \textbf{87.35}$\pm$2.43 & 30.33$\pm$1.48 & 19.67$\pm$4.82 & 61.97$\pm$2.25
        \\
        \ourmethod 
        & \textbf{22.10}$\pm$5.49 & \textbf{24.91}$\pm$1.80 & 62.23$\pm$2.98 & 91.55$\pm$3.19 & \textbf{17.50}$\pm$7.52 & \textbf{9.34}$\pm$4.50 & \textbf{49.92}$\pm$5.80
        \\ \midrule
        KEP-2/5      
        & \textbf{20.38}$\pm$5.22 & 24.39$\pm$2.05 & 63.60$\pm$1.71 & 90.75$\pm$2.62 & 26.04$\pm$7.51 & 17.94$\pm$9.12 & 57.77$\pm$8.04 
        \\ 
        \ourmethod   
        & 20.11$\pm$2.48 & \textbf{23.23}$\pm$2.36 & \textbf{63.62}$\pm$0.91 & \textbf{89.81}$\pm$1.37 & \textbf{15.67}$\pm$5.90 & \textbf{7.64}$\pm$1.45 & \textbf{48.23}$\pm$5.01
        \\ \midrule
        KEP-5/5       
        & \textbf{21.14}$\pm$3.48 & 24.06$\pm$2.27 & 63.33$\pm$1.36 & 90.46$\pm$2.20 & 23.27$\pm$6.61 & 14.25$\pm$10.66 & 55.12$\pm$7.78
        \\
        \ourmethod  
        & 20.20$\pm$5.46 & \textbf{21.97}$\pm$1.72 & \textbf{65.12}$\pm$1.63 & \textbf{90.17}$\pm$3.89 & \textbf{17.49}$\pm$5.93 & \textbf{8.61}$\pm$1.99 & \textbf{48.55}$\pm$3.67
        \\ \bottomrule
    \end{tabular}}
    \end{center}
    \vspace{-3mm}
\end{table}

\subsubsection{Effect of Varying Loss Weights}
\label{subsec:varying-loss-weights}
\begin{table}[t]
    \caption{Ablation on loss weighting on CoLA when diffusion-based reconfiguring KEP-5/5 with \ourmethod. We report mean $\pm$ std across 5 seeds. The configuration $(\lambda_{\text{mean}}, \lambda_{\text{Chol}}, \lambda_{\text{NLL}}) = (0.5, 0.2, 0.3)$ (\textcolor{gray}{gray}) is chosen as it balances predictive performance (MCC) with uncertainty calibration. This setting is applied consistently for all diffusion-based reconfigured KEP~\citep{chen2024self} produced by~\ourmethod{} .}
    \label{tab:loss_weights_cola}
    \begin{center}
    \resizebox{0.95\columnwidth}{!}
    {
    \begin{tabular}{l|lll|lll}
        \toprule
        \textbf{Method} & $\lambda_{\text{mean}}$ & $\lambda_{\text{Chol}}$ & $\lambda_{\text{NLL}}$ & \textbf{MCC} $\uparrow$ & \textbf{ECE} $\downarrow$ & \textbf{NLL} $\downarrow$ \\
        \midrule
        Transformers & -- & -- & -- & 29.92$\pm$1.17 & 26.44$\pm$1.90 & 19.66 $\pm$ 4.18 \\
        \midrule
        \multirow{7}{*}{\ourmethod{}} & 0.9  & 0.05 & 0.05 & 29.98$\pm$3.45 & 5.96$\pm$4.60 & 5.86$\pm$0.49 \\
        & 0.6  & 0.1  & 0.3  & 33.18$\pm$1.53 & 18.36$\pm$2.29 & 10.09$\pm$2.38 \\
        & 0.5  & 0.3  & 0.2  & 32.26$\pm$2.87 & 15.27$\pm$9.06 & 9.51$\pm$2.70 \\
        & 0.5  & 0.25 & 0.25 & 32.96$\pm$1.85 & 12.27$\pm$8.57 & 8.23$\pm$2.71 \\
        \rowcolor{gray!15}\cellcolor{white}
        & 0.5  & 0.2  & 0.3  & 31.81$\pm$2.30 & 12.65$\pm$6.04 & 7.69$\pm$1.70 \\
        & 0.3  & 0.2  & 0.5  & 31.86$\pm$2.79 & 18.64$\pm$8.43 & 13.77$\pm$7.32 \\
        & 0.3  & 0.1  & 0.6  & 30.21$\pm$2.32 & 16.50$\pm$8.60 & 10.74$\pm$5.80 \\
        & 0.25 & 0.25 & 0.5  & 33.38$\pm$2.08 & 22.26$\pm$3.72 & 15.10$\pm$4.85 \\
        & 0.05 & 0.05 & 0.9  & 30.74$\pm$1.44 & 15.50$\pm$6.34 & 9.52$\pm$4.12 \\
        \bottomrule
    \end{tabular}}
    \end{center}
    \vspace{-1mm}
\end{table}

We perform extensive ablation studies on the loss weight configurations to investigate their impact on \ourmethod’s performance on both the CoLA and CIFAR-10 datasets. Specifically, we examine how varying the weights assigned to the mean matching term ($\lambda_{\text{mean}}$), the Cholesky-like factor alignment ($\lambda_{\text{Chol}}$), and the performance-aware loss ($\lambda_{\text{NLL}}$) affects both predictive accuracy and uncertainty calibration. When applying diffusion-based reconfiguration with \ourmethod{} to KEP-5/5 on CoLA (see Table~\ref{tab:loss_weights_cola}), we observe distinct trends: excessively high $\lambda_{\text{mean}}$, as in the configuration $(0.9, 0.05, 0.05)$, produces very strong calibration while maintaining decent predictive performance. Conversely, assigning too much weight to $\lambda_{\text{NLL}}$, as in $(0.05, 0.05, 0.9)$, prioritizes accuracy at the expense of calibration, leading to more confident yet less well-calibrated predictions. Intermediate configurations, such as $(0.5, 0.2, 0.3)$, provide a balanced trade-off, achieving high predictive performance while significantly improving calibration. Based on these insights, we adopt $(0.5, 0.2, 0.3)$ consistently for all diffusion-based reconfigured KEP produced by \ourmethod, as it offers the most reliable combination of accuracy and calibrated uncertainty.

We also study the effect of varying the weights for the mean matching term ($\lambda_{\text{mean}}$) and the performance-aware loss ($\lambda_{\text{NLL}}$) while setting $\lambda_{\text{Chol}}=0$ during diffusion-based reconfiguring ViT with \ourmethod{} on CIFAR-10 (Table~\ref{tab:loss_weights_cifar}). When $\lambda_{\text{mean}}$ is excluded (e.g., $(0.0, 1.0)$), the NLL term dominates, which may slightly improve calibration but leads to reduced predictive performance. In contrast, including both mean matching ($\lambda_{\text{mean}}\neq 0$) and performance-aware loss consistently enhances accuracy and uncertainty calibration. Notably, the configuration $(0.8, 0.2)$ achieves the best overall balance, with the highest accuracy (87.16) and lowest ECE (8.54). Consequently, we adopt $(0.8, 0.2)$ for all diffusion-based reconfigured ViT produced by \ourmethod.

\begin{table}[!t]
    \caption{Ablation on loss weighting on CIFAR-10 (single run) when reconfiguring ViT with \ourmethod. We report test accuracy (ACC), Expected Calibration Error (ECE), and Negative Log-Likelihood (NLL). The configuration $(\lambda_{\text{mean}}, \lambda_{\text{NLL}}) = (0.8, 0.2)$ (\textcolor{gray}{gray}) is selected as it achieves the best balance between accuracy and calibration, and is used for all diffusion-based reconfigured ViT produced by~\ourmethod{} .}
    \label{tab:loss_weights_cifar}
    \begin{center}
    \resizebox{0.95\columnwidth}{!}
    {
   \begin{tabular}{l|ll|lllllll}
        \toprule
        \textbf{Method} & $\lambda_{\text{mean}}$ & $\lambda_{\text{NLL}}$
        & \textbf{ACC} $\uparrow$ 
        & \textbf{AURC} $\downarrow$ 
        & \textbf{AUROC} $\uparrow$ 
        & \textbf{FPR95} $\downarrow$ 
        & \textbf{ECE} $\downarrow$ 
        & \textbf{NLL} $\downarrow$ 
        & \textbf{Brier} $\downarrow$ \\
        \midrule
        ViT & -- & -- & 84.22 & 3.90 & 86.62 & 65.34 & 12.39 & 10.67 & 27.52 \\
        \midrule
        \multirow{10}{*}{\ourmethod{}} & 0.0 & 1.0 & 84.03 & 4.01 & 85.97 & 67.81 & 12.12 & 9.69 & 27.49 \\
        & 0.1 & 0.9 & 85.24 & 3.26 & 87.54 & 62.87 & 10.97 &  8.51 & 24.99 \\
        & 0.2 & 0.8 & 85.67 & 3.03 & 88.16 & 61.83 & 10.52 &  7.75 & 24.11 \\
        & 0.3 & 0.7 & 86.08 & 3.06 & 87.57 & 59.91 & 10.24 &  7.86 & 23.44 \\
        & 0.4 & 0.6 & 86.05 & 2.91 & 88.51 & 62.22 & 10.25 &  7.61 & 23.45 \\
        & 0.5 & 0.5 & 86.18 & 2.86 & 88.21 & 63.89 &  9.97 &  7.41 & 23.45 \\
        & 0.6 & 0.4 & 86.60 & 2.83 & 87.81 & 64.93 &  9.50 &  6.98 & 22.58 \\
        & 0.7 & 0.3 & 86.66 & 2.73 & 88.39 & 61.02 &  9.18 &  6.74 & 22.05 \\
        \rowcolor{gray!15}\cellcolor{white}
        & 0.8 & 0.2 & \textbf{87.16} & \textbf{2.52} & \textbf{88.92} & \textbf{57.94} & \textbf{8.54} & \textbf{6.06} & \textbf{20.99} \\
        & 0.9 & 0.1 & 85.09 & 3.54 & 86.73 & 64.19 & 10.15 &  7.37 & 24.79 \\
        \bottomrule
    \end{tabular}}
    \end{center}
    \vspace{-3mm}
\end{table}

\subsubsection{Training Configuration and Learning Dynamics}
\label{subsubsec:utm-training}

\begin{figure}[t]
    \centering
    \begin{subfigure}[b]{0.495\textwidth}
        \centering
        \includegraphics[width=\textwidth]{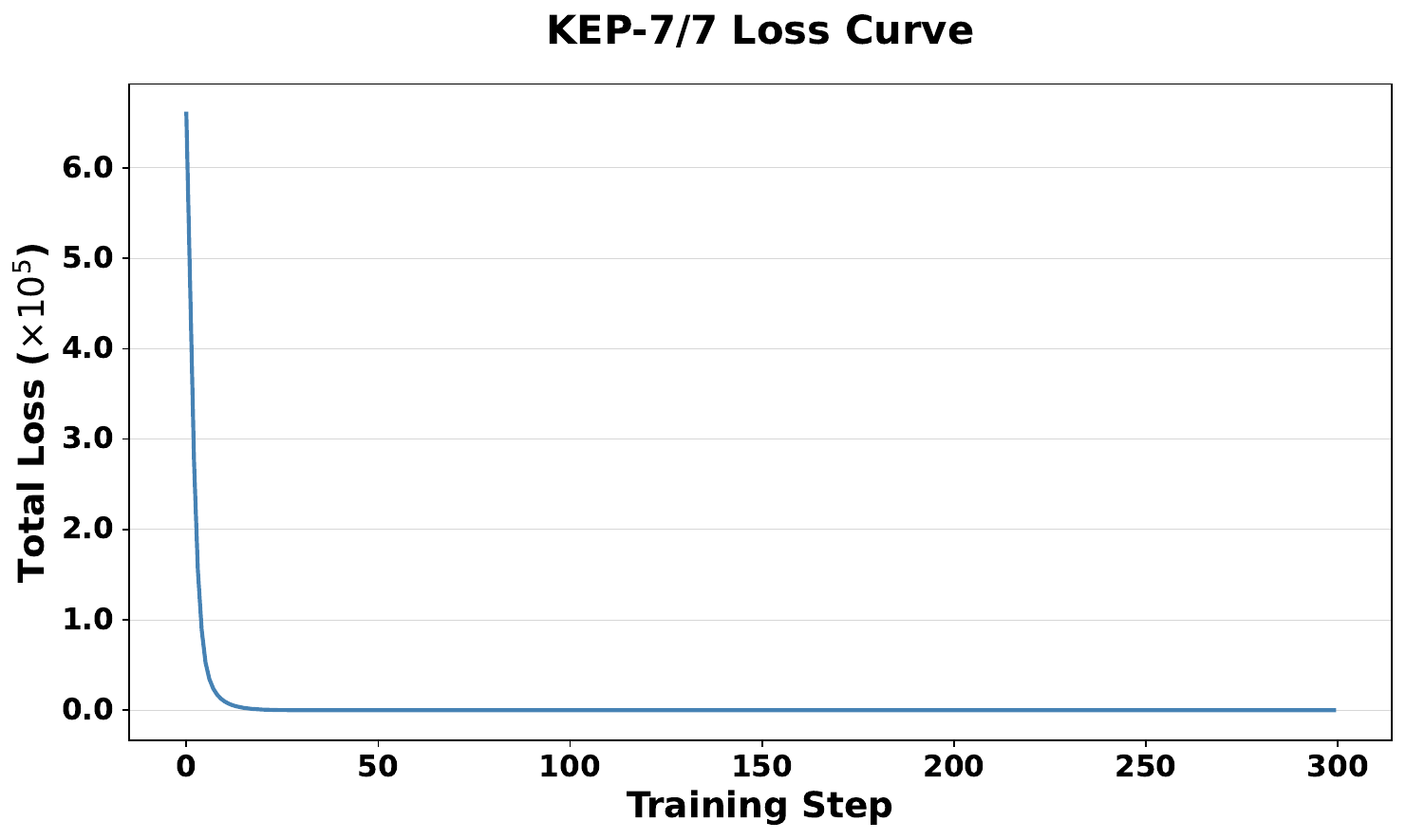}
        \caption{KEP-7/7 Loss Curve}
        \label{fig:loss_kep}
    \end{subfigure}
    \hfill
    \begin{subfigure}[b]{0.495\textwidth}
        \centering
        \includegraphics[width=\textwidth]{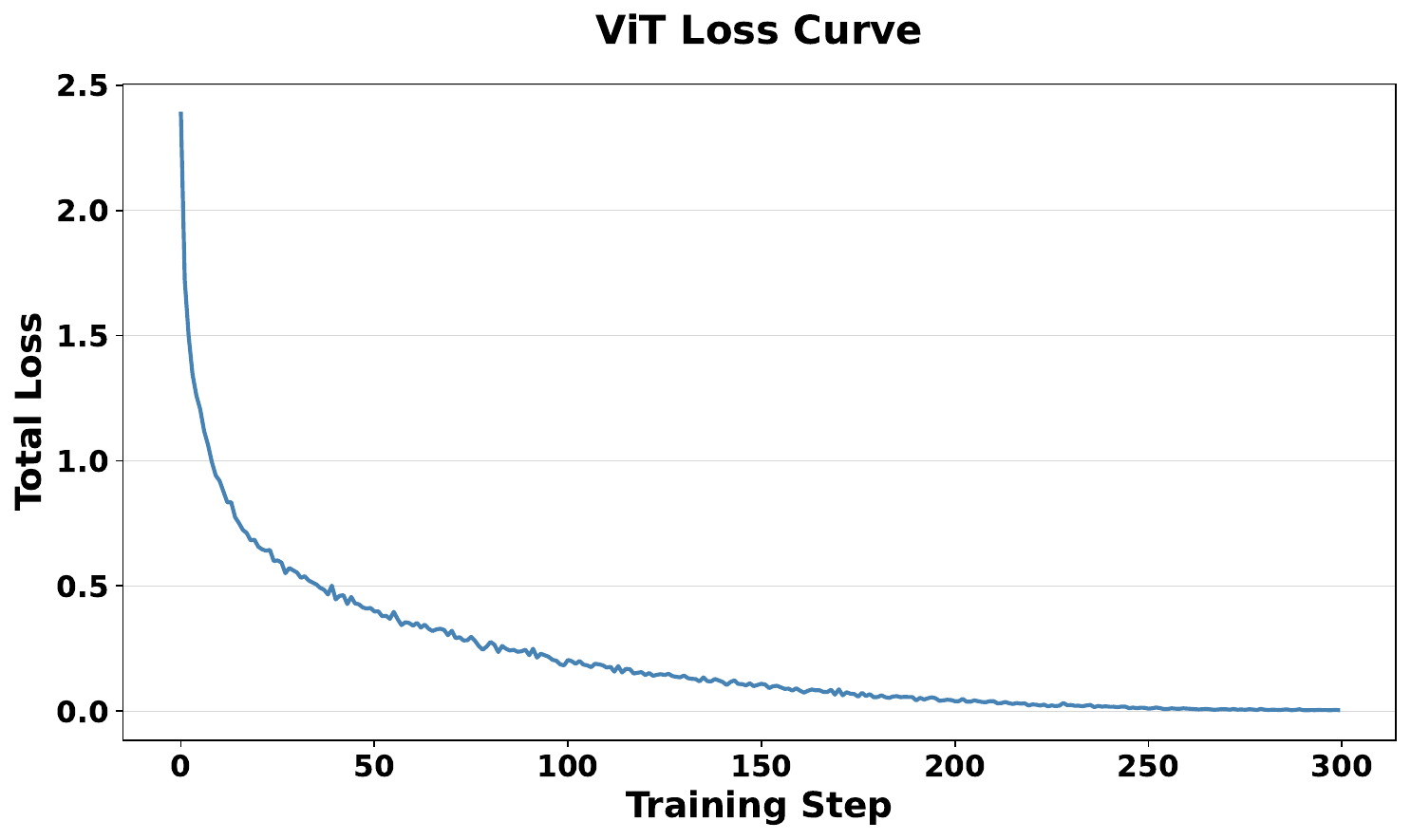}
        \caption{ViT Loss Curve}
        \label{fig:loss_vit}
    \end{subfigure}
    
    \vspace{0.5cm}
    
    \begin{subfigure}[b]{0.495\textwidth}
        \centering
        \includegraphics[width=\textwidth]{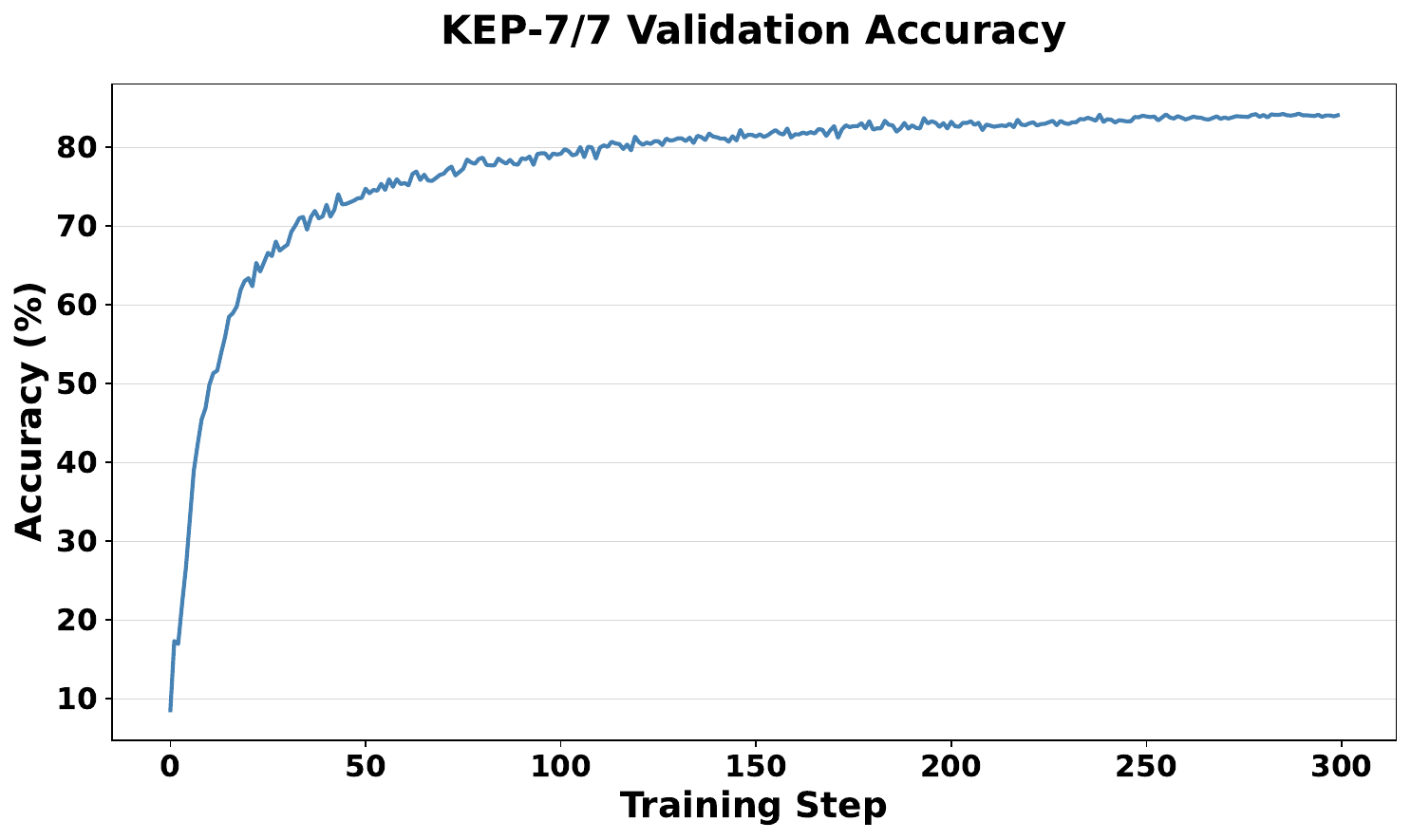}
        \caption{KEP-7/7 Validation Accuracy}
        \label{fig:val_acc_kep}
    \end{subfigure}
    \hfill
    \begin{subfigure}[b]{0.495\textwidth}
        \centering
        \includegraphics[width=\textwidth]{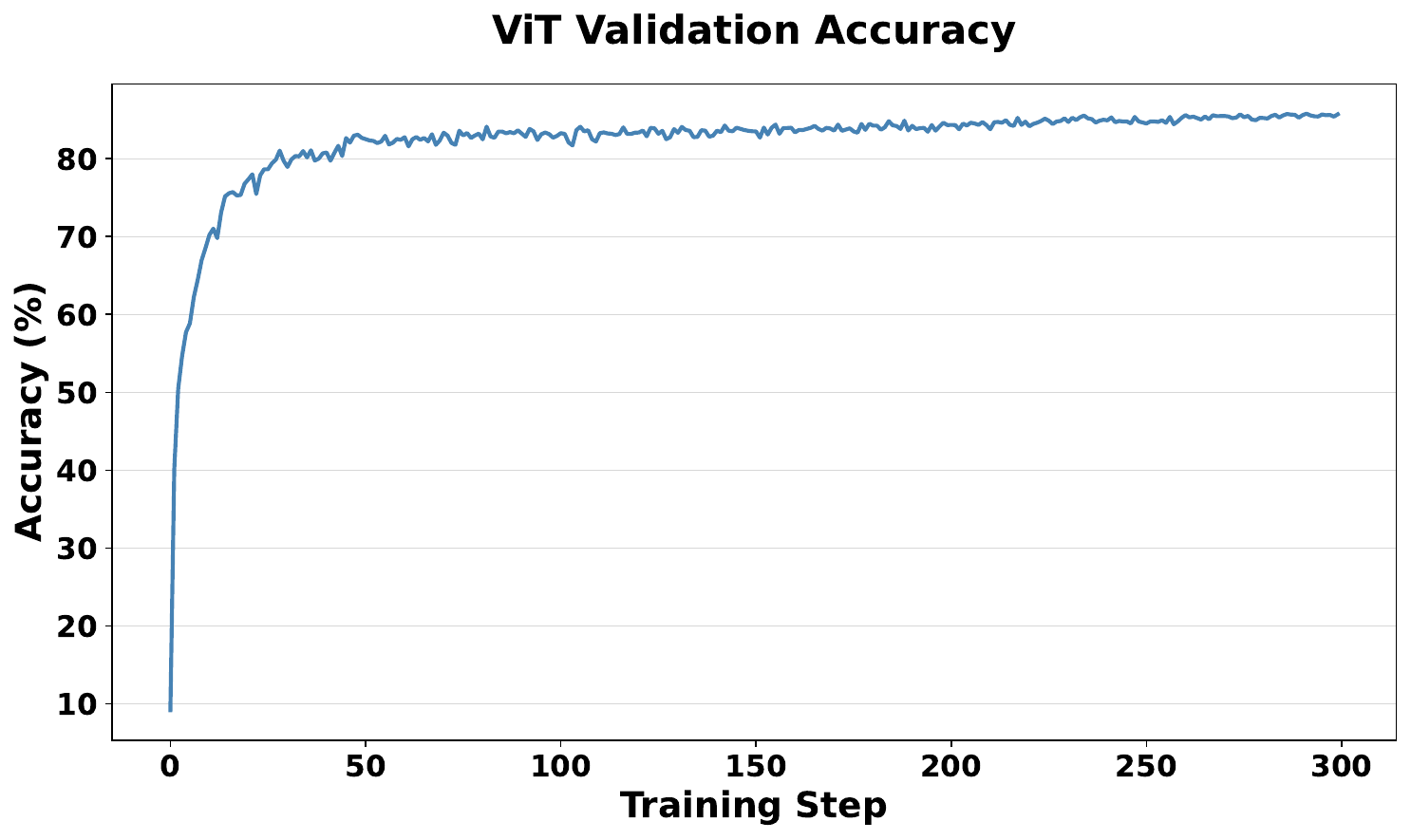}
        \caption{ViT Validation Accuracy}
        \label{fig:val_acc_vit}
    \end{subfigure}
    
    \caption{Training curves comparison between KEP-7/7 and ViT models. (a) and (b) show the loss curves, while (c) and (d) show the validation accuracy over training steps.}
    \label{fig:training_curves}
\end{figure}

\textbf{Pre-trained transformer-based models.}~We reproduce the pre-trained ViT and KEP-SVGP models following the experimental settings in~\cite{chen2024self}, as their weights are not publicly available. To ensure that the reproduced models converge properly, we report training loss and validation accuracy over epochs for ViT and KEP-7/7 on CIFAR-10 in Figure~\ref{fig:training_curves}. The curves indicate stable convergence for both architectures, providing a reliable starting point for subsequent reconfiguration.

\textbf{\ourmethod.}~Building on these pre-trained models, we reconfigure each using a DiT-based unified transition model that takes $(X_t, t)$ as input and outputs the feature representation at diffusion step $t$. This unified spatiotemporal transition model is implemented using a single-block DiT, which conditions on the diffusion timestep via adaptive LayerNorm-zero (AdaLN-Zero). We adopt the default DiT configuration while adjusting the model dimension $d_{\text{model}}$ to match the corresponding pre-trained backbone, ensuring compatibility. For reproducibility, all training hyperparameters for the unified transition model are summarized in Table~\ref{tab:training_setting}.

\begin{table}[t]
    \centering
    \small
    \caption{Hyperparameters for unified transition model training.}
    \begin{tabular}{|l|c|} 
        \toprule
        Hyperparameters&Settings\\
        \midrule
        $d_{\text{model}}$ & 384 (CV), 256 (CoLA), 128 (IMDB) \\
        Number of attention heads & 12 (CV), 8 (CoLA) , 4 (IMDB)\\
        Depth of DiT & 1 \\
        Epochs&100 (CV), 50 (NLP)\\
        Batch size&128 (CV), 32 (NLP)\\
        Dropout&0.1\\
        Optimizer&Adam\\
        Learning rate&1e-3 (CV), 5e-3 (NLP)\\
        Adam $\beta$&(0.9, 0.999)\\
        Weight decay&1e-5\\
        Scheduler&CosineAnnealingLR\\
        Cosine cycle epochs&50\\
        Minimal learning rate&1e-5\\
        Warmup Epochs&5\\
        \bottomrule
    \end{tabular}
    \vspace{-10pt}
    \label{tab:training_setting}
\end{table}

\subsection{Broader Impact}
\label{susec:broader-impact}
This work advances the reliability and interpretability of large pre-trained Transformer models by enabling principled uncertainty calibration, a critical requirement for deploying AI in high stakes domains such as healthcare, autonomous systems, and scientific discovery. By introducing a scalable framework for embedding probabilistic reasoning within existing architectures, our approach promotes safer and more accountable AI systems. Additionally, it opens new opportunities for research in uncertainty-aware foundation models and supports the responsible use of AI in socially impactful applications.

\subsection{Limitations}
\label{subsec:limitations}
While our approach enables principled uncertainty propagation in pre-trained Transformers, it introduces additional computational overhead by absorbing the pre-trained probability path into the propagation model. As the propagation model must be configured with a size compatible with that of the pre-trained model, the computational cost grows proportionally with model scale. Consequently, large-scale or very deep architectures in real-world settings require further investigation to validate the efficacy of the proposed method, which we leave for future work due to computational constraints. In addition, although we demonstrate strong empirical performance across vision and language tasks, further evaluation is needed in domains with sparse supervision or non-stationary data distributions.


\newpage
\section*{NeurIPS Paper Checklist}

\begin{enumerate}

\item {\bf Claims}
    \item[] Question: Do the main claims made in the abstract and introduction accurately reflect the paper's contributions and scope?
    \item[] Answer: \answerYes{}
    \item[] Justification: The contributions are summarised in the abstract and Section~\ref{sec:intro}, and substantiated by the experiments in Section~\ref{sec:experiments} and Appendix~\ref{app:experiments}.

\item {\bf Limitations}
    \item[] Question: Does the paper discuss the limitations of the work performed by the authors?
    \item[] Answer: \answerYes{}
    \item[] Justification: We discuss this in Appendix~\ref{subsec:limitations}, including the scale of the evaluated backbones and the iterative-sampling cost of diffusion-based propagation.

\item {\bf Theory assumptions and proofs}
    \item[] Question: For each theoretical result, does the paper provide the full set of assumptions and a complete (and correct) proof?
    \item[] Answer: \answerNA{}
    \item[] Justification: The paper does not state formal theorems; the variational training objective is derived in Appendix~\ref{app:loss-derivation}, with the underlying GP and Cholesky-like covariance assumptions stated in Sections~\ref{subsec:cut}--\ref{subsec:diffomer} and Appendix~\ref{subsubsec:varying-covarience}.

    \item {\bf Experimental result reproducibility}
    \item[] Question: Does the paper fully disclose all the information needed to reproduce the main experimental results of the paper to the extent that it affects the main claims and/or conclusions of the paper (regardless of whether the code and data are provided or not)?
    \item[] Answer: \answerYes{}
    \item[] Justification: All information needed to reproduce our experiments is provided in Section~\ref{sec:experiments} and Appendix~\ref{app:experiments}; Algorithm~\ref{alg:director-training} gives the exact training iteration of~\ourmethod{}.

\item {\bf Open access to data and code}
    \item[] Question: Does the paper provide open access to the data and code, with sufficient instructions to faithfully reproduce the main experimental results, as described in supplemental material?
    \item[] Answer: \answerYes{}
    \item[] Justification: All datasets used and pre-trained baselines are publicly available; our experimental code is submitted as supplementary material with this manuscript.

\item {\bf Experimental setting/details}
    \item[] Question: Does the paper specify all the training and test details (e.g., data splits, hyperparameters, how they were chosen, type of optimizer) necessary to understand the results?
    \item[] Answer: \answerYes{}
    \item[] Justification: Such details can be found in Section~\ref{sec:experiments} and Appendix~\ref{subsubsec:utm-training} (Table~\ref{tab:training_setting}); the loss-weight selection procedure is described in Appendix~\ref{subsec:varying-loss-weights}.

\item {\bf Experiment statistical significance}
    \item[] Question: Does the paper report error bars suitably and correctly defined or other appropriate information about the statistical significance of the experiments?
    \item[] Answer: \answerYes{}
    \item[] Justification: All metrics in Tables~\ref{tab::in_dist_kep_diff} and~\ref{tab::in_dist_baselines_diff} are reported as mean $\pm$ standard error over five seeds, and Appendix~\ref{subsubsec:signigicance-testing} reports paired $t$-tests for the main comparisons.

\item {\bf Experiments compute resources}
    \item[] Question: For each experiment, does the paper provide sufficient information on the computer resources (type of compute workers, memory, time of execution) needed to reproduce the experiments?
    \item[] Answer: \answerYes{}
    \item[] Justification: Section~\ref{sec:experiments} reports the computing resource; Appendix~\ref{subsubsec:inference-time-mem} reports per-method inference time and memory, and Table~\ref{tab:cov_struct} reports stage-1 and stage-2 wall-clock training times on CIFAR-10.

\item {\bf Code of ethics}
    \item[] Question: Does the research conducted in the paper conform, in every respect, with the NeurIPS Code of Ethics \url{https://neurips.cc/public/EthicsGuidelines}?
    \item[] Answer: \answerYes{}
    \item[] Justification: We have read the NeurIPS Code of Ethics and do not find our work in violation of any aspects.

\item {\bf Broader impacts}
    \item[] Question: Does the paper discuss both potential positive societal impacts and negative societal impacts of the work performed?
    \item[] Answer: \answerYes{}
    \item[] Justification: We discuss broader impacts in Appendix~\ref{susec:broader-impact}; improved uncertainty calibration benefits the trustworthy deployment of pre-trained transformers in risk-sensitive domains, while we do not introduce content-generating models that pose specific misuse risks.

\item {\bf Safeguards}
    \item[] Question: Does the paper describe safeguards that have been put in place for responsible release of data or models that have a high risk for misuse (e.g., pre-trained language models, image generators, or scraped datasets)?
    \item[] Answer: \answerNA{}
    \item[] Justification: Our work does not release any new datasets or pre-trained generative models with a high risk of misuse; the proposed method is a calibration mechanism applied on top of already-public backbones.

\item {\bf Licenses for existing assets}
    \item[] Question: Are the creators or original owners of assets (e.g., code, data, models), used in the paper, properly credited and are the license and terms of use explicitly mentioned and properly respected?
    \item[] Answer: \answerYes{}
    \item[] Justification: We cite the source of all datasets and pre-trained models used in our experiments (CIFAR-10/100, IMDB, CoLA, the HuggingFace ViT-B-16 and Qwen-2.5 checkpoints) and use them under their original licences.

\item {\bf New assets}
    \item[] Question: Are new assets introduced in the paper well documented and is the documentation provided alongside the assets?
    \item[] Answer: \answerNA{}
    \item[] Justification: Our work does not release any new datasets or pre-trained models; only training and evaluation code is released as supplementary material.

\item {\bf Crowdsourcing and research with human subjects}
    \item[] Question: For crowdsourcing experiments and research with human subjects, does the paper include the full text of instructions given to participants and screenshots, if applicable, as well as details about compensation (if any)?
    \item[] Answer: \answerNA{}
    \item[] Justification: Our work does not involve crowdsourcing nor research with human subjects.

\item {\bf Institutional review board (IRB) approvals or equivalent for research with human subjects}
    \item[] Question: Does the paper describe potential risks incurred by study participants, whether such risks were disclosed to the subjects, and whether Institutional Review Board (IRB) approvals (or an equivalent approval/review based on the requirements of your country or institution) were obtained?
    \item[] Answer: \answerNA{}
    \item[] Justification: Our work does not involve crowdsourcing nor research with human subjects.

\item {\bf Declaration of LLM usage}
    \item[] Question: Does the paper describe the usage of LLMs if it is an important, original, or non-standard component of the core methods in this research? Note that if the LLM is used only for writing, editing, or formatting purposes and does \emph{not} impact the core methodology, scientific rigor, or originality of the research, declaration is not required.
    \item[] Answer: \answerNA{}
    \item[] Justification: This research employs LLMs exclusively for editing purposes, with no impact on the core methodology. The Qwen-2.5 model in Appendix~\ref{app:experiments} is used as a publicly available pre-trained backbone on which we apply our reconfiguration method, not as a methodological component.

\end{enumerate}

\end{document}